\documentclass{article}
\PassOptionsToPackage{compatV3}{fancyhdr} %
\usepackage{iclr2026_conference,times}

\usepackage{natbib}
\usepackage[utf8]{inputenc} %
\usepackage[T1]{fontenc}    %
\usepackage{hyperref}       %
\hypersetup{
  pdftitle={AstaBench: Rigorous Benchmarking of AI Agents with a Scientific Research Suite},
  pdfauthor={Jonathan Bragg and Mike D'Arcy and Nishant Balepur and Dan Bareket and Bhavana Dalvi and Sergey Feldman and Dany Haddad and Jena D. Hwang and Peter Jansen and Varsha Kishore and Bodhisattwa Prasad Majumder and Aakanksha Naik and Sigal Rahamimov and Kyle Richardson and Amanpreet Singh and Harshit Surana and Aryeh Tiktinsky and Rosni Vasu and Guy Wiener and Chloe Anastasiades and Stefan Candra and Jason Dunkelberger and Dan Emery and Rob Evans and Malachi Hamada and Regan Huff and Rodney Kinney and Matt Latzke and Jaron Lochner and Ruben Lozano-Aguilera and Cecile Nguyen and Smita Rao and Amber Tanaka and Brooke Vlahos and Peter Clark and Doug Downey and Yoav Goldberg and Ashish Sabharwal and Daniel S. Weld},
}
\usepackage{url}            %
\usepackage{xurl}           %
\usepackage{booktabs}       %
\usepackage{enumitem} 
\usepackage{makecell}
\usepackage[htt]{hyphenat}  %
\usepackage{multirow}       %
\usepackage{amsfonts}       %
\usepackage{nicefrac}       %
\usepackage{microtype}      %
\usepackage[table]{xcolor}  %
\definecolor{backgroundcolor}{HTML}{FAF2E9} %
\definecolor{primarycolor}{RGB}{77,116,120}
\definecolor{linkcolor}{RGB}{77,116,120}
\usepackage{dirtree}
\usepackage{graphicx}
\usepackage[capitalise]{cleveref}  %
\usepackage{amssymb}
\usepackage{subcaption}
\usepackage{rotating}
\usepackage{tablefootnote}
\usepackage{multifootnote}  %
\usepackage{listings}
\usepackage[tight-spacing=true]{siunitx}
\usepackage{placeins}   %
\usepackage{mdframed}     %
\setlist[description]{style=standard,leftmargin=0pt,labelsep=0.5em,font=\normalfont\bfseries}

\usepackage{fancyvrb}
\usepackage{fvextra}
\newcommand{\inputblockfile}[1]{%
  \VerbatimInput[fontsize=\small,breaklines=true]{#1}%
}

\usepackage{xspace}
\usepackage{xstring}

\newcommand\finalcopy

\ifdefined\finalcopy
    \iclrfinalcopy
    \newcommand{\multiredactforblind}[1]{\footnote{#1}}
\else
    \newcommand{\multiredactforblind}[1]{\footnotetag{urlredacted}}
\fi

\ifdefined\enablecomments
    \newcommand{\mycomment}[1]{#1}
\else
    \newcommand{\mycomment}[1]{}
\fi

\newcommand{\point}[1]{{\em #1}}

\newcommand{\comment}[1]{}

\newcommand{\numagentclasses}{22\xspace}
\newcommand{\numagentclassesinagentbaselines}{16\xspace}
\newcommand{\numagentinstances}{57\xspace}
\newcommand{\numastaagentclasses}{nine\xspace}

\newcommand{\numproblems}{2400\xspace}
\newcommand{\numproblemsplus}{2400+\xspace}

\newcommand{\figformat}{pdf}

\newrobustcmd\B{\DeclareFontSeriesDefault[rm]{bf}{b}\bfseries}

\newcommand{\yes}{\checkmark\xspace}
\DeclareRobustCommand{\somewhat}{\raisebox{-0.2ex}{$\sim$}\xspace}
\newcommand{\no}{$\times$\xspace}
\newcommand{\missing}{?\xspace}
\newcommand{\cicheck}[4]{%
  \ifx\relax#2\relax%
    \xspace%
  \else%
    \allowbreak{\scriptsize$\pm$}\xspace%
  \fi%
}

\newcommand{\opennessTextClosed}{Closed \& UI only}
\newcommand{\opennessTextClosedWithApi}{Closed source \& API available}
\newcommand{\opennessTextOpenClosedWeight}{Open-source, closed-weight}
\newcommand{\opennessTextOpenOpenWeight}{Open-source, open-weight}
\newcommand{\opennessSymbolClosed}{\no}
\newcommand{\opennessSymbolClosedWithApi}{$\mathcal{A}$\xspace}
\newcommand{\opennessSymbolOpenClosedWeight}{\somewhat}
\newcommand{\opennessSymbolOpenOpenWeight}{\yes}
\newcommand{\toolingTextStandard}{Standard}
\newcommand{\toolingTextEquivalent}{Custom interface\xspace}
\newcommand{\toolingTextCustom}{Fully custom}
\newcommand{\toolingSymbolCustom}{\no}
\newcommand{\toolingSymbolEquivalent}{\somewhat}
\newcommand{\toolingSymbolStandard}{\yes}

\makeatletter
\def\DefineAbbr#1#2{%
  \DeclareRobustCommand#1{#2%
    \futurelet\nextchar\DefineAbbr@i
  }%
}
\def\DefineAbbr@i{%
  \ifx\nextchar,%
    \let\nextcmd\@gobble
  \else
    \let\nextcmd\@empty
  \fi
  \nextcmd ,\xspace
}
\DefineAbbr{\eg}{e.g.}
\DefineAbbr{\ie}{i.e.}
\DefineAbbr{\Eg}{E.g.}
\DefineAbbr{\Ie}{I.e.}
\makeatother

\let\origurl\url
\let\orighref\href

\ifdefined\finalcopy
\else
  \renewcommand{\url}[1]{%
    \IfSubStr{#1}{allen}{%
      {URL redacted for blind review}%
    }{%
        \IfSubStr{#1}{ai2}{%
          {URL redacted for blind review}%
        }{%
          \origurl{#1}%
        }%
    }%
  }

  \renewcommand{\href}[2]{%
    \IfSubStr{#1}{allen}{%
      {#2}%
    }{%
        \IfSubStr{#1}{ai2}{%
          {#2}%
        }{%
          \orighref{#1}{#2}%
        }%
    }%
  }
\fi

\makeatletter
\newcommand{\textttsmall}[1]{%
  \texttt{%
    \ifdim\f@size pt>\dimexpr9pt\relax%
      \small%
    \fi%
    \hyphenchar\font=`\-\relax #1%
  }%
}
\makeatother

\newcommand{\agentcodeurl}[2]{\url{https://github.com/allenai/agent-baselines/tree/1ce836604c37da38de2a69614800c20ca6616349/#1}\texttt{@\detokenize{#2}}}
\newcommand{\leaderboardurlraw}{https://allenai.org/asta/leaderboard}
\newcommand{\leaderboardurl}{\url{\leaderboardurlraw}}
\newcommand{\astaurlraw}{https://asta.allen.ai}
\newcommand{\astaurl}{\url{\astaurlraw}}

\newcommand{\astamarketingurl}{\url{\astamarketingurl}}
\newcommand{\astabenchurlraw}{https://github.com/allenai/asta-bench}
\newcommand{\astabenchurl}{\url{\astabenchurlraw}}
\newcommand{\agentevalurlraw}{https://github.com/allenai/agent-eval}
\newcommand{\agentevalurl}{\url{\agentevalurlraw}}

\newcommand{\agentbaselinesurl}{\url{https://github.com/allenai/agent-baselines}}
\newcommand{\agentpaperfinderurl}{\url{https://github.com/allenai/asta-paper-finder}}

\newcommand{\astabench}{AstaBench\xspace}
\newcommand{\astaenvironment}{Asta Environment\xspace}
\newcommand{\agentssuite}{Agents Suite\xspace}
\newcommand{\agentevaltoolkit}{Agents Evaluation Toolkit\xspace}
\newcommand{\agentbaselines}{\textttsmall{agent-baselines}\xspace}
\newcommand{\agenteval}{\textttsmall{agent-eval}\xspace}

\newcommand{\catlit}{Literature Understanding\xspace}
\newcommand{\catlitshort}{Lit. Und.\xspace}
\newcommand{\catdata}{Data Analysis\xspace}
\newcommand{\catcoding}{Code \& Execution\xspace}
\newcommand{\catcodingshort}{Code \& Exec.\xspace}
\newcommand{\catendtoend}{End-to-End Discovery\xspace}
\newcommand{\catendtoendshort}{End-to-End Disc.\xspace}

\newcommand{\toollitapi}{\textttsmall{Asta Scientific Corpus}\xspace}

\newcommand{\toolnotebook}{\textttsmall{Computational Notebook}\xspace}

\newcommand{\evalpaperfinder}{\textttsmall{PaperFindingBench}\xspace}
\newcommand{\evalpaperfinderNoSmall}{\texttt{PaperFindingBench}\xspace}
\newcommand{\evalpaperfinderPlain}{{PaperFindingBench}\xspace}

\newcommand{\evallitqasearchft}{\textttsmall{LitQA2-FullText-Search}\xspace}
\newcommand{\evallitqasearchftNoSmall}{\texttt{LitQA2-FullText-Search}\xspace}
\newcommand{\evallitqasearchftPlain}{{LitQA2-FullText-Search}\xspace}
\newcommand{\evallitqaft}{\textttsmall{LitQA2-FullText}\xspace}
\newcommand{\evallitqaftNoSmall}{\texttt{LitQA2-FullText}\xspace}
\newcommand{\evallitqaftPlain}{{LitQA2-FullText}\xspace}
\newcommand{\evalsqa}{\textttsmall{ScholarQA-CS2}\xspace}
\newcommand{\evalsqaNoSmall}{\texttt{ScholarQA-CS2}\xspace}
\newcommand{\evalsqaPlain}{{ScholarQA-CS2}\xspace}
\newcommand{\evaltables}{\textttsmall{ArxivDIGESTables-Clean}\xspace}
\newcommand{\evaltablesNoSmall}{\texttt{ArxivDIGESTables-Clean}\xspace}
\newcommand{\evaltablesPlain}{{ArxivDIGESTables-Clean}\xspace}

\newcommand{\evaldiscoverybench}{\textttsmall{DiscoveryBench}\xspace}
\newcommand{\evaldiscoverybenchNoSmall}{\texttt{DiscoveryBench}\xspace}
\newcommand{\evaldiscoverybenchPlain}{{DiscoveryBench}\xspace}

\newcommand{\evalcorebench}{\textttsmall{CORE-Bench-Hard$^-$}\xspace}
\newcommand{\evalcorebenchNoSmall}{\texttt{CORE-Bench-Hard$^-$}\xspace}
\newcommand{\evalcorebenchPlain}{{CORE-Bench-Hard$^-$}\xspace}
\newcommand{\evalsuper}{\textttsmall{SUPER-Expert}\xspace}
\newcommand{\evalsuperNoSmall}{\texttt{SUPER-Expert}\xspace}
\newcommand{\evalsuperPlain}{{SUPER-Expert}\xspace}
\newcommand{\evaldatasci}{\textttsmall{DS-1000}\xspace}
\newcommand{\evaldatasciNoSmall}{\texttt{DS-1000}\xspace}
\newcommand{\evaldatasciPlain}{{DS-1000}\xspace}

\newcommand{\evalendtoend}{\textttsmall{E2E-Bench}\xspace}
\newcommand{\evalendtoendNoSmall}{\texttt{E2E-Bench}\xspace}
\newcommand{\evalendtoendPlain}{{E2E-Bench}\xspace}
\newcommand{\evalendtoendhard}{\textttsmall{E2E-Bench-Hard}\xspace}
\newcommand{\evalendtoendhardNoSmall}{\texttt{E2E-Bench-Hard}\xspace}
\newcommand{\evalendtoendhardPlain}{{E2E-Bench-Hard}\xspace}

\newcommand{\agentReAct}{\textttsmall{ReAct}\xspace}
\newcommand{\agentReActNoSmall}{\texttt{ReAct}\xspace}
\newcommand{\agentReActPlain}{{ReAct}\xspace}
\newcommand{\agentSmolagents}{\textttsmall{Smolagents Coder}\xspace}
\newcommand{\agentSmolagentsNoSmall}{\texttt{Smolagents Coder}\xspace}
\newcommand{\agentSmolagentsPlain}{{Smolagents Coder}\xspace}
\newcommand{\agentAsta}{\textttsmall{Asta v0}\xspace}
\newcommand{\agentScholarQA}{\textttsmall{Asta Scholar QA (w/ Tables)}\xspace}
\newcommand{\agentScholarQANoTables}{\textttsmall{Asta Scholar QA}\xspace}

\newcommand{\agentPaperFinder}{\textttsmall{Asta Paper Finder}\xspace}
\newcommand{\agentPaperFinderNoSmall}{\texttt{Asta Paper Finder}\xspace}
\newcommand{\agentPaperFinderPlain}{{Asta Paper Finder}\xspace}
\newcommand{\agentElicit}{\textttsmall{Elicit}\xspace}
\newcommand{\agentPerplexitySQA}{\textttsmall{Perplexity Sonar Deep Research}\xspace}
\newcommand{\agentYouComResearch}{\textttsmall{You.com Research API}\xspace}

\newcommand{\agentYouComSearch}{\textttsmall{You.com Search API}\xspace}
\newcommand{\agentSciSpace}{\textttsmall{SciSpace Deep Review}\xspace}
\newcommand{\agentOpenScholar}{\textttsmall{OpenSciLM}\xspace}
\newcommand{\agentOpenAIDeepResearch}{\textttsmall{OpenAI Deep Research}\xspace}

\newcommand{\agentFutureHouseCrow}{\textttsmall{FutureHouse Crow}\xspace}
\newcommand{\agentFutureHouseFalcon}{\textttsmall{FutureHouse Falcon}\xspace}
\newcommand{\agentSTORM}{\textttsmall{STORM}\xspace}
\newcommand{\agentAstaTableAgent}{\textttsmall{Asta Table Synthesis}\xspace}
\newcommand{\agentDataVoyager}{\textttsmall{Asta DataVoyager}\xspace}

\newcommand{\agentAstaCode}{\textttsmall{Asta Code}\xspace}
\newcommand{\agentAutoAsta}{\textttsmall{Asta Panda}\xspace}
\newcommand{\agentCodeScientist}{\textttsmall{Asta CodeScientist}\xspace}
\newcommand{\agentFaker}{\textttsmall{Faker}\xspace}
\newcommand{\captionDaggerNotice}{ $\dagger$ denotes models not pinned to a date-stamped version.\xspace}
\newcommand{\captionParetoNotice}{Bold denotes the agent is on Pareto-optimal frontier for that column pair.\xspace}
\newcommand{\modelDRAmbiguous}{\textttsmall{o3-/o4-mini- deep-research}\xspace}

\newcommand{\modelOThreeDRShort}{\textttsmall{o3-deep-research}\xspace}
\newcommand{\modelOThree}{\textttsmall{o3-2025-04-16}\xspace}
\newcommand{\modelOThreeShort}{\textttsmall{o3}\xspace}
\newcommand{\modelOThreeUnpinned}{\textttsmall{o3}$^\dagger$\xspace}
\newcommand{\modelOThreeUnpinnedID}{\textttsmall{o3}\xspace}

\newcommand{\modelOThreeMiniShort}{\textttsmall{o3-mini}\xspace}
\newcommand{\modelGPTFourPointOne}{\textttsmall{gpt-4.1-2025-04-14}\xspace}
\newcommand{\modelGPTFourPointOneShort}{\textttsmall{gpt-4.1}\xspace}
\newcommand{\modelGPTFourPointOneUnpinned}{\textttsmall{gpt-4.1}$^\dagger$\xspace}
\newcommand{\modelGPTFourPointOneUnpinnedID}{\textttsmall{gpt-4.1}\xspace}

\newcommand{\modelGPTFourPointOneMiniShort}{\textttsmall{gpt-4.1-mini}\xspace}
\newcommand{\modelGPTFourO}{\textttsmall{gpt-4o-2024-08-06}\xspace}
\newcommand{\modelGPTFourOMiniShort}{\textttsmall{gpt-4o-mini}\xspace}
\newcommand{\modelGPTThreeFiveTurbo}{\textttsmall{gpt-3.5-turbo-0125}\xspace}
\newcommand{\modelGPTThreeFiveTurboShort}{\textttsmall{gpt-3.5-turbo}\xspace}
\newcommand{\modelGPTFourTurboShort}{\textttsmall{gpt-4-turbo}\xspace}
\newcommand{\modelGPTFourOShort}{\textttsmall{gpt-4o}\xspace}
\newcommand{\modelGPTFourOUnpinned}{\textttsmall{gpt-4o}$^\dagger$\xspace}
\newcommand{\modelGPTFourOUnpinnedID}{\textttsmall{gpt-4o}\xspace}
\newcommand{\modelGPTFive}{\textttsmall{gpt-5-2025-08-07}\xspace}
\newcommand{\modelGPTFiveShort}{\textttsmall{gpt-5}\xspace}
\newcommand{\modelGPTFiveUnpinned}{\textttsmall{gpt-5}$^\dagger$\xspace}
\newcommand{\modelGPTFiveUnpinnedID}{\textttsmall{gpt-5}\xspace}
\newcommand{\modelGPTFiveUnpinnedMinimal}{\textttsmall{gpt-5}$^\dagger$\textttsmall{:effort=minimal}\xspace}
\newcommand{\modelGPTFiveMini}{\textttsmall{gpt-5-mini-2025-08-07}\xspace}
\newcommand{\modelGPTFiveMiniShort}{\textttsmall{gpt-5-mini}\xspace}
\newcommand{\modelGPTFiveMiniUnpinned}{\textttsmall{gpt-5-mini}$^\dagger$\xspace}
\newcommand{\modelGPTFiveMiniUnpinnedID}{\textttsmall{gpt-5-mini}\xspace}

\newcommand{\modelClaudeSonnetFour}{\textttsmall{claude-sonnet-4-20250514}\xspace}
\newcommand{\modelClaudeSonnetFourShort}{\textttsmall{claude-sonnet-4}\xspace}
\newcommand{\modelClaudeThreeFiveHaiku}{\textttsmall{claude-3-5-haiku-20241022}\xspace}
\newcommand{\modelClaudeThreeFiveHaikuShort}{\textttsmall{claude-3-5-haiku}\xspace}
\newcommand{\modelClaudeSonnetThreeSeven}{\textttsmall{claude-3-7-sonnet-20250219}\xspace}
\newcommand{\modelClaudeSonnetThreeSevenShort}{\textttsmall{claude-3-7-sonnet}\xspace}

\newcommand{\modelGeminiTwoPointFivePro}{\textttsmall{gemini-2.5-pro}\xspace}
\newcommand{\modelGeminiTwoPointFiveProShort}{\textttsmall{gemini-2.5-pro}\xspace}
\newcommand{\modelGeminiTwoPointFiveFlashUnpinned}{\textttsmall{gemini-2.5-flash}$^\dagger$\xspace}
\newcommand{\modelGeminiTwoPointFiveFlashUnpinnedID}{\textttsmall{gemini-2.5-flash}\xspace}
\newcommand{\modelGeminiTwoPointFiveFlash}{\textttsmall{gemini-2.5-flash-preview-05-20}\xspace}
\newcommand{\modelGeminiTwoPointFiveFlashShort}{\textttsmall{gemini-2.5-flash}\xspace}
\newcommand{\modelGeminiTwoFlash}{\textttsmall{gemini-2.0-flash}\xspace}
\newcommand{\modelGeminiTwoFlashShort}{\textttsmall{gemini-2-flash}\xspace}

\newcommand{\modelLlamaFourScout}{\textttsmall{Llama-4-Scout-17B-16E-Instruct}\xspace}
\newcommand{\modelLlamaFourScoutShort}{\textttsmall{llama-4-scout}\xspace}

\newcommand{\modelPerplexitySonarDeepResearch}{\textttsmall{sonar-deep-research}\xspace}
\newcommand{\modelPerplexitySonarDeepResearchShort}{\textttsmall{sonar-deep-research}\xspace}
\newcommand{\modelOpenScholar}{\textttsmall{llama-3.1-openscholar-8b}\xspace}

\makeatletter
\@ifpackageloaded{array}{}{\RequirePackage{array}}
\@ifpackageloaded{collcell}{}{\RequirePackage{collcell}}
\makeatother

\def\umeanops{}
\def\uuncops{}
\newcommand{\uncertaintysize}{\footnotesize}

\ExplSyntaxOn
\NewDocumentCommand{\processcell}{m}
{
  \tl_if_in:nnTF {#1} {+-}
  {
    \seq_set_split:Nnn \l_tmpa_seq {~+-~} {#1}
    \seq_get_left:NN \l_tmpa_seq \l_tmpa_tl
    \seq_get_right:NN \l_tmpa_seq \l_tmpb_tl
    
    \expandafter\tablenum\expandafter[\umeanops]{\l_tmpa_tl}
    {\uncertaintysize\,$\pm$\,}
    {\uncertaintysize\expandafter\tablenum\expandafter[\uuncops]{\l_tmpb_tl}}
  }
  {
    #1
  }
}
\ExplSyntaxOff

\newcolumntype{U}[2]{%
  >{\def\umeanops{#1}\def\uuncops{#2}%
    \collectcell\processcell}%
  c%
  <{\endcollectcell}%
}

\newcolumntype{L}[1]{>{\raggedright\arraybackslash}p{#1}}
\newcolumntype{R}[1]{>{\raggedleft\arraybackslash}p{#1}}
\newcolumntype{A}{U{table-format=2.1,round-mode=places,round-precision=1}{table-format=2.1{*},round-mode=places,round-precision=1}}
\newcolumntype{B}{U{table-format=1.3,round-mode=places,round-precision=3}{table-format=1.3{*},round-mode=places,round-precision=3}}
\newcolumntype{C}{S[round-mode=places,round-precision=1,table-format=2.1{*}]}
\newcolumntype{D}{S[round-mode=places,drop-uncertainty,round-precision=2,table-format=1.2{*}]}

\author{%
Jonathan~Bragg$^1$ \quad Mike~D'Arcy$^1$ \\
\AND
Nishant~Balepur$^{2,*}$ \quad Dan~Bareket$^1$ \hspace{-1px}\quad Bhavana~Dalvi$^1$ \hspace{-1px}\quad Sergey~Feldman$^1$ \quad Dany~Haddad$^1$ \quad\\\bfseries Jena~D.~Hwang$^1$ \quad Peter~Jansen$^{1,3}$ \quad Varsha~Kishore$^{1,6}$ \quad Bodhisattwa~Prasad~Majumder$^1$ \quad\\\bfseries Aakanksha~Naik$^1$ \quad Sigal~Rahamimov$^1$ \quad Kyle~Richardson$^1$ \quad Amanpreet~Singh$^1$ \quad\\\bfseries Harshit~Surana$^{1}$ \quad Aryeh~Tiktinsky$^1$ \quad Rosni~Vasu$^{4,*}$ \quad Guy~Wiener$^1$ \\
\AND
Chloe~Anastasiades$^1$ \quad Stefan~Candra$^1$ \quad Jason~Dunkelberger$^1$ \quad Dan~Emery$^1$ \quad\\\bfseries Rob~Evans$^1$ \quad Malachi~Hamada$^1$ \quad
\quad Regan~Huff$^1$ \quad Rodney~Kinney$^1$ \quad Matt~Latzke$^1$ \quad\\\bfseries Jaron~Lochner$^1$ \quad Ruben~Lozano-Aguilera$^1$ \quad Cecile~Nguyen$^1$ \quad Smita~Rao$^1$ \quad\\\bfseries Amber~Tanaka$^1$ \quad Brooke~Vlahos$^1$ \\
\AND\vspace{-24pt}
Peter~Clark$^1$ \quad Doug~Downey$^1$ \quad Yoav~Goldberg$^{1,5}$ \quad Ashish~Sabharwal$^1$ \quad Daniel~S.~Weld$^1$
\\[0.8em]
\parbox{\textwidth}{\rule{0pt}{28pt}{\normalfont\noindent%
$^1$Asta Team, Allen Institute for AI, $^2$University of Maryland, $^3$University of Arizona, \\[0em]
$^4$University of Zurich, $^5$Bar-Ilan  University, $^6$University of Washington,  \\[0em]
$^*$Work performed while at Ai2 Asta Team. \,\\ %
}}
}%

\title{AstaBench: Rigorous Benchmarking of AI Agents with a %
Scientific Research Suite}
\date{}

\begin{document}
\maketitle
\begin{abstract}
AI agents hold the potential to revolutionize scientific productivity by automating literature reviews, replicating experiments, analyzing data, and even proposing new directions of inquiry;
indeed, there are now many such agents, ranging from general-purpose ``deep research'' systems to specialized science-specific agents, such as AI Scientist and AIGS.
Rigorous evaluation of these agents is critical for  progress. Yet existing benchmarks fall short on several fronts: they often
(1) lack reproducible agent tools necessary for a controlled comparison of core agentic capabilities; (2) do not account for confounding variables such as model cost and tool access;
(3) do not provide standardized interfaces for quick agent prototyping and evaluation;
(4) fail to provide holistic, product-informed measures of real-world use cases such as science research;
and (5) lack comprehensive baseline agents necessary to identify true advances.
In response, we define principles and tooling for more rigorously benchmarking agents. Using these, we present \href{\astabenchurlraw}{\astabench}, a suite that provides a holistic measure of agentic ability to perform scientific research, comprising \numproblemsplus problems spanning the entire scientific discovery process and multiple scientific domains, and including many problems inspired by actual user requests to deployed \href{\astaurlraw}{Asta} agents.
Our suite comes with the first scientific research environment
with production-grade search tools that enable controlled, reproducible evaluation, better accounting for confounders.
Alongside, we provide a comprehensive suite of \numastaagentclasses science-optimized classes of Asta agents and numerous baselines.
Our extensive evaluation of \numagentinstances agents across \numagentclasses agent classes reveals several interesting findings, most importantly that despite meaningful progress on certain individual aspects, AI remains far from solving the challenge of science research assistance.
\end{abstract}

\section{Introduction}

AI agents are increasingly being applied to complex real-world use cases. In particular, they hold the promise to revolutionize scientific productivity by automating reviews of the literature, replicating complex experiments, analyzing high volumes of data, and even proposing new avenues to explore.
Large organizations such as OpenAI and Google are investing in general-purpose ``deep research'' systems to help everyone, including scientists, comb through literature much more effectively. We even have specialized science-specific agents, such as AI Scientist~\citep{Lu2024TheAS,Yamada2025TheAS} and AIGS~\citep{Liu2024AIGSGS}, targeting scientific research.
With so many different agents---many behind paywalls and all evaluated in  bespoke ways---how are end users and AI developers to know which perform best?

Unfortunately, existing agent benchmark suites have several deficiencies, when considered as a general measure of AI skill, including for their ability to do scientific research (\cref{tab:comparison}).
First, suites often \point{lack the standard task environments and tools} necessary for realistic, controlled comparison of agents on a level playing field; %
for example, no large-scale, controlled document retrieval tools exist,
making it unclear whether a `winning' agent has superior AI capabilities or merely access to a more relevant information source.
Second, they \point{fail to properly account for confounding variables}; we are unaware of agent benchmarks that account for variations in tool usage, and only a few like HAL~\citep{hal} measure cost, which is critical since even simplistic strategies (\eg, taking a majority vote over repeated invocation) can boost accuracy by spending more \citep{kapoor2024aiagentsthatmatter}.
Third, \point{benchmark suite interfaces are rarely standardized for use by general agents}, since suite developers typically assume either that users will evaluate only agents that come with the suite (and so it is fine for evals to be coupled to agents, as in the case of OpenHands~\citep{wang2025openhands} or AutoGen~\citep{fourney2024magentic}) or that users will build only specialized agents for specific benchmarks (as is the case with general suites like HAL~\citep{hal}). Measuring new agents on a full suite typically requires time-consuming interventions ranging from extensive decoupling to manually clarifying task instructions that were not written with general agents in mind; this  harms reproducibility and controlled comparison.
Fourth, they often \point{lack tasks that are informed by authentic product usage data} (typically guarded by technology companies),  raising concerns that higher scores may not lead to meaningful real-world benefit.
Finally, benchmark suites \point{lack comprehensive agent baselines} for proper comparison. As a result,  most published evaluations only compare to a small number of other agents or ablations, making it difficult to assess whether claimed improvements represent genuine advances.

\begin{table}[t]
\setlength{\tabcolsep}{3pt}
\centering

\caption{AstaBench improves over existing agent benchmark suites in several ways. It tests holistic scientific reasoning (\ie, a broad spectrum of task types and across more than one scientific domain). Many of its problems are inspired by actual user requests to our deployed Asta agents. Its standard tool environment isolates core agentic abilities (e.g., planning, tool-calling, etc.) from information access. \astabench's scoring controls for confounders, such as computational cost, and its tasks are defined using a uniform format that supports general-purpose agents.
The table's final column, titled `Cls.', indicates the number of agent classes (e.g., ReAct) that are used to instantiate (e.g., with specific LLMs) the total number of agents listed in preceding column; \astabench includes more classes of agents than prior benchmarking efforts.
}
\label{tab:comparison}

\small
\rowcolors{5}{}{gray!10}
\begin{tabular}{@{}L{1.6cm}L{1.8cm}L{1.7cm}L{1.8cm}L{2.2cm}L{2.1cm}R{0.55cm}r}
\toprule
& & \multicolumn{5}{c}{Relevant for all agent benchmarks} \\
\cmidrule{3-8}
& \multicolumn{1}{c}{\makecell{Holistic sci. \\ reasoning}} & \multicolumn{1}{c}{\makecell{Product\\usage-based}} & \multicolumn{1}{c}{\makecell{Controlled, \\realistic tools}} & \multicolumn{1}{c}{\makecell{Scoring accounts\\for confounders}} & \multicolumn{1}{c}{\makecell{Tasks ready for\\ general agents}} & \multicolumn{2}{c}{\makecell{\# Agents \\ \hline Total Cls.}} \\
\midrule
\rowcolor{backgroundcolor}
\textbf{\textcolor{primarycolor}{\astabench}} & \yes Broad (weighted towards CS) & \somewhat Lit.\ tasks & \yes Prod.-grade lit.\ corpus & \yes Costs, tools, openness & \yes Decoupled, with standard formats & \numagentinstances & \numagentclasses \\
\ifdefined\finalcopy
    AutoGen\-Bench & \no No science & \no & \no & \no & \no Coupled to agent framework & 7 & 11 \\
    BixBench & \somewhat Bio data science & \no & \no & \no & \somewhat Non-standard notebook tools & 2 & 2 \\
    BrowserGym & \no No science & \no & \no & \no & \yes Ready for web agents & 10 & 2 \\
\fi
HAL & \somewhat Coding & \no & \no & \somewhat Costs & \no Non-standard formats & 113 & 10 \\
Inspect Evals & \somewhat Coding, knowledge & \no  & \no & \no & \no Non-standard formats & 18 & 1 \\
LAB-Bench & \somewhat Bio & \no & \no & \no & \no Non-standard formats & 12 & 3 \\
OpenHands Evals & \somewhat Coding, data analysis & \no & \no & \somewhat Costs & \no Coupled to agent framework & 53 & 6 \\
\ifdefined\finalcopy
    ScienceAgent\-Bench & \somewhat Data analysis & \no & \no & \somewhat Costs & \no Coupled to agents & 17 & 3  \\
    Terminal-Bench & \somewhat Coding & \no & \no & \no & \yes Ready for terminal agents & 33 & 12 \\
    Vector Inst. Leaderboard & \no No science & \no & \no & \no & \no Non-standard formats & 5 & 1 \\
\fi
\bottomrule
\end{tabular}
\end{table}

In response, we present a set of benchmarking principles
and a benchmark suite, built upon these principles, that overcomes the aforementioned limitations, along with open-source resources that enable more rigorous, comprehensive measurement. Specifically:%

\begin{itemize}[leftmargin=*]

\item We formalize principles for rigorously benchmarking agents (\cref{sec:principles}), which address key limitations of current agent benchmark suites.

\item Guided by our principles, we present
\astabench\multiredactforblind{\astabenchurl}
(\cref{sec:benchmarks}), a more rigorous agent benchmark suite
that is \point{a holistic measure of scientific research}, which exercises a broad spectrum of skills---including literature understanding, data understanding, planning, tool use, coding, and search---and comprises over \numproblems problems spanning the full scientific discovery process and multiple scientific domains, including many problems based on real user requests from
Asta,\multiredactforblind{\astaurl}
where we have deployed several of our agents for public use.
It is easy to integrate new general agents with \astabench, which provides a standardized task interface.

\ifdefined\finalcopy
\else
\multifootnote[urlredacted]{\,URL redacted for blind review}
\fi

\item \astabench includes the powerful \astaenvironment (\cref{sec:env}), the \point{first agent environment that enables controlled, reproducible evaluation with production-grade search tools} for retrieving information from a large corpus of scientific literature.

\item We also introduce the
\agenteval\agentevaltoolkit\multiredactforblind{\agentevalurl}
(\cref{sec:lb-toolkit}), which enables defining a benchmark suite and leaderboard with time-invariant cost accounting using model usages logged by Inspect~\citep{inspect_ai_framework}, a standard agent evaluation framework that provides broad model and evaluation compatibility.

\item We introduce \astabench
Leaderboard\multiredactforblind{\leaderboardurl}
built using this Toolkit. It's
the \point{first agent leaderboard to properly account for confounding variables} such as
the tools used by the agent and inference cost.

\item Finally, we present the
\agentbaselines\agentssuite\multiredactforblind{\agentbaselinesurl}
(\cref{sec:agents}), the \point{most comprehensive standardized agents suite}, comprised of \numastaagentclasses Asta agent classes that have been optimized for scientific research tasks, as well as numerous baselines.

\end{itemize}

\begin{figure}[t]
    \centering
        \includegraphics[width=0.75\linewidth]{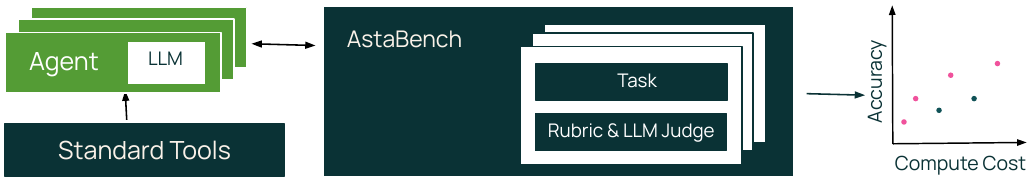}
    \caption{Using \astabench we evaluated \numagentclasses agent classes on a diverse set of science tasks while controlling the set of available tools, \eg, to ensure  each agent has access to the same set of scientific papers. \astabench leaderboards record not just agents accuracy but also how much computation is required to achieve that performance.}
    \label{fig:asta-evaluation}
    \vspace*{-0.2in}
\end{figure}

Together, the \astabench benchmark suite, agent environment, agents suite, and leaderboard enable a \point{holistic measurement of the current state of LLM agents for scientific research assistance}, as well as a path for continuous improvement (\cref{fig:asta-evaluation}). We report on an extensive set of experiments on \astabench using our agents suite with \numagentinstances agents spanning \numagentclasses classes of agent architectures, ranging from task-specific agents such as Asta Scholar QA and Asta CodeScientist to generic, ReAct-style architectures applicable to the broad range of benchmarks within \astabench.  We find that while meaningful progress has been made on many fronts, \point{science research assistance remains far from solved}. \cref{sec:experiments} summarizes our findings, with more details in the appendices.

These findings provide a current snapshot of the state of scientific research assistance agents. But this is only a starting point. \astabench offers the ability to help the community continually and systematically assess progress (or lack thereof) as new agents are designed, something that has been difficult to do holistically. We hope \astabench will continue to serve as a valuable guide for the development of future agents through its clear targets, cost-aware performance reporting, and transparent evaluation regimen.

\section{Related Work}
\label{sec:relate}

Our efforts relate to two recent threads of research: the development of \emph{holistic agent evaluations} that test a wide range of LLM-driven automation (for a general review, see \cite{yehudai2025surveyagenteval}) and the development of new benchmarks for measuring the \emph{scientific reasoning} of LLMs and their use as \emph{scientific assistants and agents} \citep{wang2023scientific}. We consider each in turn.

{\bf Holistic Agent Evaluations}
The last few years have seen a surge in benchmarks and evaluation frameworks that attempt to holistically measure the reasoning abilities of LLMs \citep[\eg][]{gu2024olmes, eleuther-eval-harness, lighteval, evalchemy2024}. Given the rise of LLM-driven automation, recent efforts have centered around new benchmarks and frameworks for evaluating LLM {\em agents}.
\cref{tab:comparison} highlights recent efforts that are most closely related to \astabench in terms of their scope as holistic or science agent benchmarks: AutoGenBench~\citep{fourney2024magentic}, BixBench~\citep{mitchener2025bixbench}, BrowserGym \citep{lesellier2025browsergym}, the Holistic Agent Leaderboard (HAL)~\citep{hal}, Inspect Evals~\citep{inspect_evals_2025}, Lab-Bench~\citep{laurent2024labbench}, OpenHands Evals~\citep{wang2025openhands}, ScienceAgentBench~\citep{scienceagentbench}, Terminal-Bench~\citep{terminalbench2025}, and the Vector Institute Leaderboard~\citep{vector2025eval}.%
\footnote{Agent counts for \cref{tab:comparison} were derived from live leaderboards and repositories accessed August 2025, in addition to the canonical benchmark references~\citep{autogen_agents_repo2025,browsergym_leaderboard2025,hal_leaderboard2025,inspect_evals_dashboard2025,openhands_agents_repo2025,openhands_leaderboard2025,terminalbench_leaderboard2025}.}
We compare these efforts to \astabench across the following dimensions: \textbf{holistic scientific reasoning} (\ie, focuses on a broad spectrum of task types and across more than one scientific domain), \textbf{product usage-based} (\ie, involves tasks based on product use cases), \textbf{controlled, realistic tools} (\ie, distributes standard, realistic tools that allow for controlled comparison of agents), \textbf{scoring accounts for confounders} (\ie, scores systematically account for cost, controlled tool use, and other confounders), \textbf{general agents} (\ie, tasks have uniform formats that support general-purpose agents), and \textbf{number of agents} (\ie, total number and number of different classes of agent).

\astabench stands out on these dimensions, which are key to advancing scientific AI and increasing benchmarking rigor generally (\cref{sec:principles}).
In terms of science, the other agent benchmark suites are all less holistic, either more limited in terms of task category (\eg, HAL's only science tasks are coding tasks) or the domain (\eg, LAB-Bench is limited to biology);
AstaBench is also the only benchmark to leverage data from a companion product (Asta) in its tasks.
Despite its importance, few suites have seriously focused on cost (HAL is an exception), and none have distributed standard tools that are decoupled from agents or agent frameworks.
While some leaderboards are scaling up the number of agents they test (again, notably HAL), all test far fewer agent classes (architectures) compared to \astabench, which also {\em distributes} open-source code for these agent classes through \agentbaselines\agentssuite.

{\bf Science Benchmarks and Agents for Science}
Naturally, the rise of powerful large language models (LLMs) has led to much recent interest in LLM-driven approaches to scientific research-related tasks. Many new benchmarks have been developed, often focusing on particular sub-problems in the full research pipeline, including scientific coding and execution~\citep{tian2024scicode, lai2023ds, chen2025mlrbench, chan2025mlebench, mlagentbench}, data analysis~\citep{majumder-etal-2025-discoverybench,xu2025researcherbench},  research reproduction~\citep{bogin-etal-2024-super-emnlp,siegel-etal-2025-corebench, tang2025airesearcher, kon2025expbench, xiang2025scireplicate, starace2025paperbench, zhao2025autoreproduce, yan2025lmr}, ideation and hypothesis generation~\citep{ruan2024liveideabench,si2024can, vasu2025hyperliteraturegroundedhypothesisgeneration}, and literature retrieval and understanding~\citep{shi2025spar,he-etal-2025-pasa}, among others~\citep{zhu2025safescientist}. \astabench spans many of these task categories, and provides the most comprehensive evaluation of scientific agent performance to date (\cref{tab:comparison}).

Increased LLM capabilities have led to emergence of a host of agents for end-to-end, open-ended scientific discovery, including AI Scientist~\citep{Lu2024TheAS, Yamada2025TheAS}, Agent Lab~\citep{agentlab}, AIGS~\citep{Liu2024AIGSGS}, and CodeScientist~\citep{codescientist}, among others~\citep{cheng2025language}. To bring clarity to this area (and accelerate its progress), \astabench introduces a new end-to-end task that evaluates an agent's ability to complete a research project, starting from an idea and ending with a written report and code. We believe this task is a useful complement to the many existing benchmarks that focus on more narrow problems in the research pipeline.%

\section{\astabench: A Holistic Scientific Research Benchmark Suite}
\label{sec:benchmarks}

We present \astabench, the first benchmark suite for holistic evaluation of agents' ability to perform scientific research.
Crucially, our suite is reproducible even as science progresses, since it comes with the first realistic, reproducible search tools (\cref{sec:env}).
Our suite implements a new standard interface for agent benchmark suites and provides time-invariant cost reporting through the \agenteval\agentevaltoolkit (\cref{sec:lb-toolkit})). %
As such, \astabench is ready for use by new general agents such as those in our agent baselines suite (\cref{sec:agents}).

\vspace{1em}
\noindent
\astabench comprises the following 11 benchmarks (summarized in \cref{tab:evals}, with full details in \cref{sec:appendix-evals} and example inputs in \cref{sec:appendix:evals-samples}; note that \astabench uses slightly modified versions of some of the cited datasets):
{\evalpaperfinder} tests an agent's ability to handle challenging scientific search queries.
{\evallitqaft/\evallitqasearchft}~\citep{Skarlinski2024LanguageAgents} measure an agent's ability to answer questions and retrieve papers within the biomedical domain.
{\evalsqa} tests an agent's ability to answer long-form scientific questions.
{\evaltables}~\citep{newman-etal-2024-arxivdigestables} tests an agent's ability to create a literature review table.
{\evalsuper}~\citep{bogin-etal-2024-super-emnlp} tests the ability of code agents to set up and execute Python machine learning experiments reported in ML and NLP papers.
{\evalcorebench}~\citep{siegel-etal-2025-corebench} tests an agent's ability to reproduce experiments and analyses from papers.\footnote{\evalcorebench omits GPU-requiring tasks from the original CORE-Bench-Hard; see \cref{sec:appendix-evals}.}
{\evaldatasci}~\citep{lai2023ds} tests the ability of agents on data science tasks encountered in research.
{\evaldiscoverybench}~\citep{majumder-etal-2025-discoverybench} tests whether the agent can automatically find and verify hypotheses from given dataset(s).
{\evalendtoend/\evalendtoendhard} test whether agents can perform the full research pipeline of ideation, planning, (software) experiment design, implementation, execution, analysis, and producing a final report.

Full details of how these tasks are scored can be found in \cref{sec:appendix-evals}.  Some use LLMs as judges to evaluate outputs against rubrics (\evalpaperfinder, \evalsqa, \evaltables, \evaldiscoverybench, \evalendtoend, \evalendtoendhard) while others use programmatic evaluation (\evallitqaft, \evallitqasearchft, \evalsuper, \evalcorebench, \evaldatasci).

\begin{table}[t]
  \centering

  \caption{%
    \astabench benchmarks, spanning four task categories: \catlit, \catcoding, \catdata, and \catendtoend.
    Benchmarks are fully reproducible when paired with the \astaenvironment tools listed in the `Tools' column, which come standard with each benchmark:
    \toolnotebook (Code) or \toollitapi (Corpus) tools that restrict to papers before the specified `Date Cutoff' (exclusive).
    (Original datasets were filtered to ensure questions are answerable with the environment.)
    $^\ddagger$For \evaltables, corpus tools are restricted to snippet search with specific paper IDs for each problem.
    $^*$~indicates created by us, %
    and $\dagger$~indicates previously unreleased.
  }
  \label{tab:evals}

  \small
  
  \begin{tabular}{@{}l l l r r l c@{}}
    \toprule
    Name & Task category & Domains & Test & Val & Tools & Date Cutoff \\
    \midrule
    \evalpaperfinder$^*$$\dagger$        & \catlitshort (search) & CS    & 267 & 66  & Corpus & 2025-06-01 \\
    \evallitqasearchft        & \catlitshort (search) & Biology    & 75 & 10  & Corpus & 2024-10-17 \\
    \evalsqa$^*$$\dagger$                & \catlitshort (report) & CS    & 100 & 100 & Corpus & 2025-05-01 \\
    \evallitqaft              & \catlitshort (MC) & Biology             & 75 & 10  & Corpus & 2024-10-17 \\
    \evaltables$^*$             & \catlitshort (table) & Mixed & 100 & 70  & Snippet$^\ddagger$ & Paper IDs \\
    \addlinespace[0.5em]

    \evalsuper$^*$              & \catcodingshort & CS & 45 & 50    & Code & --- \\
    \evalcorebench          & \catcodingshort & Mixed & 37 & 35         & Code & --- \\
    \evaldatasci            & \catcodingshort & CS & 900 & 100          & Code & --- \\
    \addlinespace[0.5em]

    \evaldiscoverybench$^*$     & \catdata & Mixed & 239 & 25     & Code & --- \\
    \addlinespace[0.5em]

    \evalendtoend$^*$$\dagger$           & \catendtoendshort & CS    & 40 & 10   & Code & --- \\
    \evalendtoendhard$^*$$\dagger$           & \catendtoendshort & CS    & 40 & 10   & Code & --- \\
    \bottomrule
  \end{tabular}
\end{table}

\section{\astaenvironment}
\label{sec:env}

\astaenvironment is, to our knowledge, the first realistic, reproducible scientific research environment for agents. It provides standardized tools, an evaluation toolkit, a leaderboard, and numerous agents.

\subsection{Standard Tools for Agents}
\label{sec:standard-tools}
\label{sec:jupyter-tool}

\astaenvironment provides a comprehensive set of standard tools for science research assistance, from which each \astabench task includes a specific subset based on its requirements
(\cref{tab:evals}).

\noindent\textbf{\toollitapi:}
A toolset for accessing the scientific literature, which represents the first production-grade, reproducible search tools for agents. 
These tools can restrict outputs to papers preceeding a date; \astabench uses this feature to limit results to the date of benchmark creation so that new papers do not contaminate results (see cutoffs for specific tasks in \cref{tab:evals}).
The \textttsmall{snippet\_search} tool can be further restricted to papers with specific IDs so that it can be used as a text retrieval mechanism over those papers (useful for detailed literature analysis, \eg in \evaltables).
It provides the following specific tools via the MCP (Model Context Protocol) standard:
    \textttsmall{snippet\_search}, 
    \textttsmall{search\_papers\_by\_relevance}, 
    \textttsmall{get\_paper}, 
    \textttsmall{get\_paper\_batch}, 
    \textttsmall{get\_citations}, 
    \textttsmall{search\_authors\_by\_name}, 
    \textttsmall{get\_author\_papers}, 
    \textttsmall{search\_paper\_by\_title}

\noindent\textbf{\toolnotebook:} A stateful computational (Jupyter) notebook. The tool can execute Python code as well as standard IPython magic commands like \textttsmall{\%\%writefile}, \textttsmall{\%matplotlib inline}, and \textttsmall{!shell\_command}. Python variables and environment are maintained between calls so that the tool can be used to solve problems incrementally.  By default, the tool returns a timeout message to the agent if a single cell takes more than 5 minutes to execute.
Since the tool needs to execute code, it lives in a new sandbox image that's created by the framework.

Our tools feature improved agent compatibility compared to other suites.
They are cleanly decoupled from agents and provide easy integration via MCP.
Code executed in our sandbox can call tools provided by the main (host) execution environment (\eg \toollitapi), enabling testing of code execution agents, \eg agents that implement the CodeAct~\citep{codeact_wang2024} pattern.

\subsection{\texttt{agent-eval} Evaluation Toolkit \& \astabench Leaderboard}
\label{sec:lb-toolkit}

We use Inspect~\citep{inspect_ai_framework} as the framework for implementing our individual agentic benchmarks, as it provides broad model provider and tool compatibility, useful logging and debugging affordances, and a growing set of compatible evals~\citep{inspect_evals_2025}. %
However, Inspect logs only model usages (not normalized dollar amounts) and it lacks tooling for defining benchmark suites with unified scoring or leaderboards.
To fill this gap, we present the
\agenteval\multiredactforblind{\agentevalurl}
agent leaderboard toolkit, which provides a benchmark suite, reporting, and leaderboard layer on top of a suite of Inspect-formatted benchmarks; it features:

\noindent\textbf{Time-invariant cost calculation:}
The \agenteval toolkit computes normalized dollar costs based on model usages logged through Inspect.
For mapping model usages to prices, we use a frozen snapshot of the \texttt{litellm} cost map, which is community-sourced for broad model coverage.\footnote{We supplement the cost map with prices for custom models based on Together AI (\url{https://www.together.ai/}) generic model size-based pricing.}
It factors in cache discounts for agents that take advantage of caching, as this is an increasingly adopted optimization technique (and providers like OpenAI provide these discounts automatically); however, it does not factor in any latency-related discounts (\eg service tier or batching).  Using a frozen snapshot allows a fair comparison of evaluation costs even if API prices change between evaluations.\footnote{The cost map snapshot used for the leaderboard may be periodically updated, but we will always re-calculate all costs based on the current snapshot to ensure fair comparison.}

\noindent\textbf{Reporting that accounts for confounders:}
In addition to cost, the \agenteval toolkit and leaderboards categorize agent evaluation submissions according to their reproducibility and degree of control based on the following dimensions (full definitions in \cref{sec:appendix-reporting}):

\begin{itemize}[itemsep=0pt,topsep=-2pt,leftmargin=5ex]
    \item \textbf{Agent openness} \emph{(is the agent implementation open?):} \opennessTextOpenOpenWeight~(\opennessSymbolOpenOpenWeight), \opennessTextOpenClosedWeight~(\opennessSymbolOpenClosedWeight), \opennessTextClosedWithApi~(\opennessSymbolClosedWithApi), or \opennessTextClosed~(\opennessSymbolClosed)
    
    \item \textbf{Agent tooling} \emph{(does the agent use the provided standard tools for the tasks?):} \toolingTextStandard~(\toolingSymbolStandard), \toolingTextEquivalent~(\toolingSymbolEquivalent), or \toolingTextCustom~(\toolingSymbolCustom)
\end{itemize}

\noindent\textbf{Leaderboard web interface:} In addition to the \agenteval CLI-based leaderboard interface (which requires authentication currently unavailable to the public for \astabench), we also include a web application interface for the \astabench
Leaderboard\multiredactforblind{\leaderboardurl},
which supports external submissions (with Hugging Face user-based authentication) and provides interactive plots and tables.

\subsection{\texttt{agent-baselines} \agentssuite}
\label{sec:agents}

To enable comprehensive measurement on \astabench and other benchmarks---and advance the state of the art---we provide the 
\agentbaselines\agentssuite,\multiredactforblind{\agentbaselinesurl}
which consists of a large set of agents from \numagentclassesinagentbaselines agent classes\footnote{Slightly less than the \numagentclasses we evaluate because some are closed source and thus not usable on new inputs; however, we provide ways to reproducing those results based on cached answers obtained for our experiments.} with a standard Inspect-compatible interface.  \cref{tab:agents} lists these agents, grouped into (1) the Asta agents that we optimized for scientific research tasks and (2) numerous baseline agents that we evaluate. Detailed descriptions are deferred to \cref{sec:appendix-agents}.

\begin{table}[tp]
  \centering

  \caption{
  Agent classes in the \agentbaselines\agentssuite, with Asta agents in the top section and baseline agents in the bottom section.
  ``\toolingTextStandard'' tooling means that the only tools used are the ones distributed with the \astabench tasks; ``\toolingTextEquivalent'' means that standard date-restricted search is used but additional custom tooling may be used; ``\toolingTextCustom'' means that tooling is custom and standard search tools are not used.
  }

  \small

  \rowcolors{2}{gray!10}{}
  \begin{tabular}{@{}l l L{1cm} l@{}}
    \toprule
    Name & Task optimization & Open-source & Tooling \\
    \midrule

    \agentPaperFinder & \catlitshort (search) & \yes Yes & \toolingSymbolEquivalent \toolingTextEquivalent \\
    \agentScholarQANoTables      & \catlitshort (report) & \yes Yes & \toolingSymbolEquivalent \toolingTextEquivalent \\
    \agentScholarQA      & \catlitshort (report) & \yes Yes & \toolingSymbolEquivalent \toolingTextEquivalent \\
    \agentAstaTableAgent & \catlitshort (table) & \yes Yes & \toolingSymbolEquivalent \toolingTextEquivalent \\
    \agentAstaCode      & \catcoding & \yes Yes & \toolingSymbolEquivalent \toolingTextEquivalent \\
    \agentDataVoyager    & \catdata & \yes Yes & \toolingSymbolEquivalent \toolingTextEquivalent \\
    \agentAutoAsta       & \catendtoendshort & \yes Yes & \toolingSymbolCustom \toolingTextCustom \\
    \agentCodeScientist  & \catendtoendshort & \yes Yes & \toolingSymbolCustom \toolingTextCustom \\
    \agentAsta  & Multi & \yes Yes & \toolingSymbolCustom \toolingTextCustom \\

    \midrule

    \agentReAct          & None (general) & \yes Yes & \toolingSymbolStandard \toolingTextStandard  \\
    \agentSmolagents & None (general) & \yes Yes & \toolingSymbolEquivalent \toolingTextEquivalent \\
    \agentYouComSearch & \catlitshort (search) & \no & \toolingSymbolCustom \toolingTextCustom \\
     \agentElicit         & \catlitshort (report) & \no & \toolingSymbolCustom \toolingTextCustom \\
    \agentFutureHouseCrow    & \catlitshort (report) & \no & \toolingSymbolCustom \toolingTextCustom \\
    \agentFutureHouseFalcon    & \catlitshort (report) & \no & \toolingSymbolCustom \toolingTextCustom \\
    \agentOpenAIDeepResearch & \catlitshort (report) & \no & \toolingSymbolCustom \toolingTextCustom \\
    \agentOpenScholar    & \catlitshort (report) & \yes Yes & \toolingSymbolEquivalent \toolingTextEquivalent \\
    \agentPerplexitySQA  & \catlitshort (report) & \no & \toolingSymbolCustom \toolingTextCustom \\
    \agentSciSpace       & \catlitshort (report) & \no & \toolingSymbolCustom \toolingTextCustom \\
    \agentSTORM          & \catlitshort (report) & \yes Yes & \toolingSymbolCustom \toolingTextCustom \\
    \agentYouComResearch & \catlitshort (report) & \no & \toolingSymbolCustom \toolingTextCustom \\
    \agentFaker & \catendtoendshort & \yes Yes & \toolingSymbolStandard \toolingTextStandard \\
    \bottomrule
  \end{tabular}
  \label{tab:agents}
\end{table}

\section{Experiments}
\label{sec:experiments}

We now present experimental results, which we have also used to seed the interactive \astabench
leaderboard.\multiredactforblind{\leaderboardurl}
Our experiments were conducted over a period of several months. %
Since one may boost scores  by using more compute (eg using repetition and majority vote)~\citep{Dodge2019ShowYW}, we report cost as well as accuracy. We also report the standard deviation of our measurements. For brevity, when an agent was tested with multiple different models, we report the top result(s) plus any other significant data points. The entire set of results, plus plots of scores vs.\ costs including the Pareto frontier (showing the best agent for a given cost), are in \cref{sec:appendix:full-results}.

Some agents (e.g., ReAct) can attempt {\it all} 11 benchmarks; others are category-specific or even benchmark-specific. \cref{tab:results-overall} shows the overall results for those agents attempting {\it all} benchmarks, as well as agents that can solve all the benchmarks in at least one category. 
Category- and benchmark-specific results are presented in \cref{sec:headline-results} for space reasons.

\begin{figure}[htbp]
  \centering
  \includegraphics[width=\textwidth]{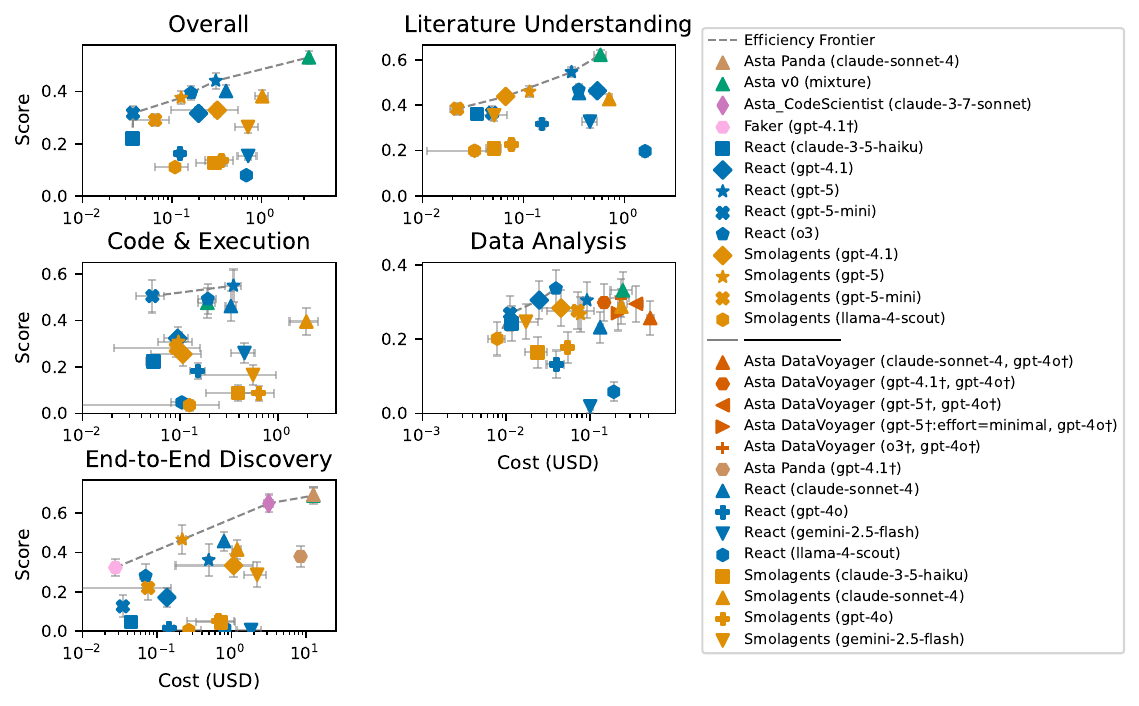}
  \vspace*{-0.3in}
  \caption{Score vs.\ cost analysis for overall and category results (from \cref{tab:results-overall,apx:tab:results-overall,apx:tab:results-data,apx:tab:results-endtoend}). Points indicate means. Points on the Pareto frontier are connected with dotted lines, representing optimal quality-cost trade-offs for each category (\catlit, \catcoding, \catdata, \catendtoend).
    \captionDaggerNotice
    Note: the x-axis (cost per answer in dollars) uses a log scale. For more detailed plots for individual categories and benchmarks, see \cref{sec:appendix:full-results}.
    }
  \label{fig:results-overall}
\end{figure}

\noindent
As noted above, agents powered by closed weight LLMs currently far exceed the reach of those powered by open weight LLMs. On the other hand, simply switching the underlying LLM with the latest and greatest one isn't necessarily a reliable recipe for success on \astabench. As a case in point, one of the newest LLMs, \modelGPTFiveShort, provides only a modest boost over an earlier ``reasoning LLM'', \modelOThreeShort, except on three benchmarks. In fact, \modelGPTFiveShort hurts the performance of several specialized agents.

\point{Tools designed specifically for science research assistance can significantly help AI agents}. This is most noticable with \agentAsta, which scores $\sim$9\% higher than the next best agent, \agentReAct with \modelGPTFiveShort (53.0\% vs. 44.0\%). However, this comes with the trade-off of significantly higher development (engineering) cost, and (for some tasks, specifically in end-to-end-discovery) higher inference cost.

\point{None of the commercial scientific research agents were able to perform the full range of research tasks in \astabench.}  The best such API-based agent (\agentFutureHouseFalcon) and the best closed one (\agentOpenAIDeepResearch) score well on literature understanding, but are unable to perform the full spectrum of science research assistance.

\point{Science research assistance is still far from solved}, as evidenced by the generally low overall scores for the full gamut of agents, from fully open to fully closed. For example:
The best open source agent with open weights LLMs scores a terrible 11.1\% (\agentSmolagents with \modelLlamaFourScout) (\cref{tab:results-overall}).
The best open source agent with closed LLM(s) is much better: 53.0\% (\agentAsta) (\cref{tab:results-overall}).
While the best API-based agent (\agentFutureHouseFalcon) and closed agent (\agentOpenAIDeepResearch) score well on a single benchmark (\cref{tab:results-lit-qa}), they are stymied by the full range of tasks. %

The cost-performance tradeoff across agents, highlighted by the Asta leaderboard's Pareto curve provides several interesting insights. \point{The best economical model is \agentReAct with \modelGPTFiveMiniShort}, scoring 32\%---within 21\% (absolute) of the best performing models---while costing over an order of magnitude less at \$0.04 per problem.

\point{Powering a general agent with an expensive model can {\em lower} the overall cost.}  
Though the per-token cost is 3 to 25 times lower for gemini-flash  and llama-scout compared to o3 or sonnet, the weaker models often take more steps or get stuck in loops, causing a ReAct agent to end up being twice as expensive in addition to lower-performing.

Surprisingly, most of our specialized agents (\agentScholarQANoTables (\cref{tab:results-lit-qa}), \agentDataVoyager (\cref{tab:results-overall}), \agentAstaCode (\cref{tab:results-coding})) \point{perform worse with \modelGPTFiveShort} than with previous models, while \agentReAct performs much better. One possible explanation for this is that \modelGPTFiveShort has been tuned to do well with now-common ReAct-style workflows, and conversely may be relatively less adaptive to alternate workflows. If this is indeed true, and trends continue, there may be diminishing value in application-specific workflows. %

As the LLM underlying \agentReAct, \point{\modelGPTFiveShort's boost over \modelOThreeShort is generally light}, with only a gain of 0\%-5\% across most benchmarks. However, \modelGPTFiveShort provides a huge boost in 4 benchmarks: +13.4\% absolute %
on \evalsqa (\cref{tab:results-lit-qa}), + 24.8\% %
on \evalsuper (\cref{tab:results-coding}),
+25.3\% on \evallitqasearchft (\cref{tab:results-lit-search}), %
and +21.1\% %
on \evalendtoendhard (\cref{tab:results-endtoend}).

In general, today's agents are reasonably good at literature understanding. However, despite some recent progress, coding, experiment execution, data analysis, and data-driven discovery still remain major, unsolved problems for science assistance agents.

{\bf Literature Understanding:}
For literature search agents, \point{\agentPaperFinder stands out as an impressive system}, scoring much higher than its closest rival (\agentReAct) on \evalpaperfinder and \evallitqasearchft (\cref{tab:results-lit-search}). Despite this, it is clear that the paper-finding task is far from `solved,' requiring further work to achieve truly comprehensive results.

For literature  question-answering agents, our results (\cref{tab:results-lit-qa}) suggest that (among other things):
\point{The best models have relatively good performance in this category}, scoring around 80\%. This is likely because literature understanding has been a strong focus of many task-optimized agents in the community (or conversely, the community has targeted literature understanding because this category is particularly well suited for language models).
\point{\agentScholarQANoTables, \agentElicit, and \agentSciSpace %
are the best tools on these tests} (all score about 85\% or higher on \evalsqa, \cref{tab:results-lit-qa}).  For all three tools, the higher performance is driven by the citation subscores of the evaluation.
The \point{other external/commercial agents are not far behind, but also do not do significantly better than the best  \agentReAct baseline}. This is indeed surprising given \agentReAct's simplicity, but is also an indicator of the challenging nature of the task that requires system responses to be precise and cover the relevant points as well as cite the correct supporting sources for claims as necessary.

For literature  review table generation agents, our results (\cref{tab:results-lit-table}) suggest that:
\point{even the best models do not yet achieve strong performance in this category}, with recall scores around 43\%, likely due to limited efforts to build task-optimized agents in this space.
\point{\agentAstaTableAgent, backed by \modelGPTFiveShort, wins on this task, beating the best general agents}. However, \agentAstaTableAgent backed by \modelGPTFiveMiniShort also shows competitive performance, at just $13\%$ of the cost.

{\bf Code and Execution:}
 \point{Coding and execution is far from solved}---all agents score low on these tasks, e.g., all but two scored below 25\% on \evalsuper (\agentReAct with \modelGPTFiveShort scored 41\% and \modelGPTFiveMiniShort scored 37\%; \cref{tab:results-coding,apx:tab:results-coding}). Coding and execution thus remain major bottlenecks for assisting with and automating science.

\point{The impact of using \modelGPTFiveShort is highly unpredictable.} Surprisingly, running the general \agentReAct agent with \modelGPTFiveShort significantly improves its performance (compared to running with other LLMs), while running the more custom-built \agentSmolagents with \modelGPTFiveShort notably {\it decreases} performance. One possible explanation is that \modelGPTFiveShort has been tuned for the common ReAct-style workflow, making \modelGPTFiveShort less adaptive to alternate workflows. %

{\bf Data Analysis:}
Similarly, \point{automated data analysis and data-driven discovery is a major, unsolved challenge for science assistance agents}. We see agents struggle with this benchmark, with the maximum score being only 34\% (\cref{tab:results-overall}) despite increased attention in the community.

{\bf End-to-End Discovery:}
\point{End-to-end discovery remains far from being meaningfully solved.} Although the {\it average} research step completion scores appear reasonable (scores up to $\sim$70\%, \cref{tab:results-endtoend}), the likelihood of completing {\it all} experiment steps remains near zero. For example, given $\sim$10 steps per experiment, and a success rate of 70\% per step, the success rate to complete {\it all} steps in the experiment will be $\approx 0.7^{10} \approx 3\%$ (see
 \cref{tab:e2e-overall-completion-scores} for actual numbers, reaching a maximum of 5\%).
A lot more work is needed, and we hope these benchmarks will help push research forward in this direction.

\section{Conclusion and Future Work}
In summary, we identify limitations of current approaches to benchmarking agents, and present methodology and tooling for doing so more rigorously.
Using this methodology and tooling, we introduce \astabench, a holistic benchmark suite for scientific research that addresses key limitations.
\astabench is the first major agent benchmark suite to come with standard environment and tools that enable controlled comparison of agents: the \astaenvironment, the first scientific research environment for agents with realistic, controlled search tools.
Alongside, we present the \agentbaselines\agentssuite, a large suite of standardized agents, which we used to conduct experiments on \astabench with \numagentinstances agents across  \numagentclasses architectural classes. This revealed several interesting findings, most importantly that despite meaningful progress on certain individual aspects, agentic AI remains far from solving the challenge of scientific research assistance.
We invite the community to make submissions to the \astabench Leaderboard, which is powered by our \agenteval\agentevaltoolkit.

This work opens up many exciting possibilities for the agentic AI, scientific research assistance, and automated scientific discovery communities.
We are actively pushing the performance-cost frontiers in AstaBench and closing the gap for truly open agents by developing new agent techniques, tools, and open models specialized for scientific research.
We are also enhancing agent abilities to manage complex context, from improving on Asta v0 simple orchestration techniques to handling long-duration tasks in complex research projects.
We are continuing to research how to refine our LLM-as-a-judge grading procedures, especially for challenging scientific discovery tasks.
We plan to develop fresh benchmark problems that use the latest scientific knowledge, which is contamination-resistant and past the training cut-off date of models.
We also plan to build benchmarks that test more aspects of collaboration with humans, and deepen coverage of problems in impactful fields such as biomedicine.
Finally, we are committed to continuing to measure the latest advances---both by testing the latest LLMs and by adding more agent architectures to \agentbaselines.

\section*{Ethics Statement}

We took care to adhere to a high ethics bar.
We obtained legal review for all material presented in this work.
The new real-world user queries used in the Literature Understanding tasks were collected with user consent.
We also credit any benchmarks that we adapted for use in our suite, as well as agents that we leverage, citing those works.
When measuring existing agents, we worked with the agent creators where possible to ensure they are measured fairly, including Elicit, Future House, and SciSpace.

\section*{Reproducibility Statement}
We took special care to make this work reproducible; indeed, reproducibility is a core value proposition of our benchmark suite.
\astabench comes with open source code for all included benchmarks, agents, and core infrastructure---as well as logs of all reported experiment.
The framework logs and reports specific repository commits, including for data.
The agent tools in \astabench improve reproducibility by providing date-restricted access to the supporting document corpus.

\newcommand{\llmusestmt}{We used AI-based tools (Claude Code, Github Copilot, ChatGPT) for analyzing results data, generating code to populate plots and tables, identifying errors and missing references, and (minor) writing assistance.}

\ifdefined\finalcopy
    \section*{Author Contributions}
\label{sec:contrib}

Authors listed in alphabetical order within each section:
\begin{itemize}
    \item \textbf{Project leadership, framework, and general agent development:} Jonathan Bragg, Mike D'Arcy

    \item \textbf{Research by task category (benchmarks and agents):}
    \begin{itemize}
        \item \textbf{\catlit (paper finding):} Dan Bareket, Yoav Goldberg, Sigal Rahamimov, Aryeh Tiktinsky, Guy Wiener
        \item \textbf{\catlit (summarization and QA):} Nishant Balepur, Doug Downey, Sergey Feldman, Dany Haddad, Jena D. Hwang, Varsha Kishore, Aakanksha Naik, Amanpreet Singh, Daniel S. Weld
        \item \textbf{\catlit (table generation):}
            \begin{itemize}
            \item {\em Benchmark:} Aakanksha Naik
            \item {\em Agent:} Mike D'Arcy, Dany Haddad, Aakanksha Naik
            \end{itemize}
        \item \textbf{\catcoding:} Mike D'Arcy, Kyle Richardson
        \item \textbf{\catdata:} Bodhisattwa Prasad Majumder, Harshit Surana
        \item \textbf{\catendtoend:} Peter Clark, Bhavana Dalvi, Peter Jansen, Rosni Vasu
    \end{itemize}

    \item \textbf{Engineering:}
        \begin{itemize}
        \item \textbf{Frameworks and leaderboard data:} Chloe Anastasiades, Stefan Candra, Regan Huff, Rodney Kinney
        \item \textbf{Leaderboard web application:} Jason Dunkelberger, Dan Emery, Cecile Nguyen, Smita Rao, Amber Tanaka, Brooke Vlahos
        \item \textbf{Management:} Jaron Lochner, Smita Rao, Rob Evans
        \end{itemize}

    \item \textbf{Design:} Matt Latzke

     \item \textbf{Support and data annotation:} Malachi Hamada

    \item \textbf{Product management:} Ruben Lozano-Aguilera

    \item \textbf{Management, mentorship, and advice:} Peter Clark, Doug Downey, Yoav Goldberg, Ashish Sabharwal, Daniel S. Weld
\end{itemize}

    \paragraph{The Use of Large Language Models (LLMs)}
    \llmusestmt
    \section*{Acknowledgments}

This work would not have been possible without a broad and supportive community.
In particular, we thank:
David Albright and Kyle Wiggers for communications support and useful feedback;
Crystal Nam for legal support;
Ali Farhadi and Sophie Lebrecht for insightful feedback and encouragement;
Stephen Kelman for design support;
the creators and maintainers of the Inspect evaluation framework;
the creators of the external datasets that we have integrated;
and
the data workers who contributed to the creation of those datasets and the datasets that we created.

\else
    \section*{The Use of Large Language Models (LLMs)}
    \llmusestmt
\fi

\bibliographystyle{iclr2026_conference}
\bibliography{refs}

\appendix

\crefname{appendix}{appendix}{appendices}
\Crefname{appendix}{Appendix}{Appendices}
\crefalias{section}{appendix}
\crefalias{subsection}{appendix}
\crefalias{subsubsection}{appendix}

\section{Principles for Benchmarking Agents}
\label{sec:principles}

We propose the following principles for more rigorously benchmarking agents:

\begin{enumerate}
    \item \textbf{The task suite must represent the complexity of real-world usage.}
    In order to determine whether agents can serve as effective assistants for a use case, it is necessary to test a broad range of relevant tasks.
    Real-world product usage provides an informative basis for determining appropriate tasks, but unfortunately such data is typically guarded by product companies (who use it to create private evaluations) and unavailable to academic benchmark creators.
    Moreover, in order to measure progress towards broadly capable agents, the task suite should require exercising a range of advanced, general skills such as reasoning, planning, tool use, search, coding, and data analysis.

    \item \textbf{A standard, realistic, and reproducible environment and tools must accompany the suite for controlled comparison of AI capabilities.}
    The environment should be realistic to measure agents' ability to act in the real world. At the same time, the environment and tools must be standard and reproducible to facilitate controlled comparison across different agents.
    Most existing benchmark suites lack standard tools,
    leading agent developers to use disparate environments and tools that obscure whether performance differences are due to superior AI capabilities or other enhancements.
    It is particularly important that benchmark suites provide {\em standard search tools} with reproducible test-time access to the same document corpus, yet large-scale, optimized search indexes are costly to create and public search tools are not reproducible; we are unaware of any such public, reproducible, large-scale search tools.

    \item \textbf{Reporting must account for confounding variables—especially computational cost and tool usage.}
    It's essential to account for cost, since even simplistic strategies, such as repeating a task many times and taking majority votes, can boost accuracy by burning cash~\cite{kapoor2024aiagentsthatmatter,hal}. Controlling for tool usage is also essential to separate gains due to model or agent architecture advancements from benefits due to privileged access to specialized information sources.

    \item \textbf{Task interfaces must be standardized to facilitate integration of general agents.}
    General agents that can perform many different tasks are likely to better meet diverse real-world needs. Unfortunately, most previous benchmark suites require general agent developers to adapt agents for individual tasks, introducing developer bias and hindering development.
    To support the development of general agents, task interfaces should provide
    `reasonable' accommodation for an intelligent agent that has not been developed specifically for the test tasks: complete task instructions, task-required tools, and submission affordances---all in a standard format.

    \item \textbf{Comprehensive agent baselines with standard interfaces are needed to measure state-of-the-art.}
    A large integrated suite of agent baselines must be available to identify which agents are truly state-of-the-art agents and to provide high-quality starting points for future development, yet is lacking from current agent suites resulting in most evaluations comparing only to a small number of other agents or ablations on the evaluator's own agent.

\end{enumerate}

\section{Evaluation Toolkit: Openness and Tooling}
\label{sec:appendix-reporting}

Definitions for the \textbf{Agent openness} and \textbf{Agent tooling} classifications for baseline:

\begin{itemize}
\item \textbf{Agent openness} describes the transparency and reproducibility of an agent's implementation:
    \begin{itemize}
    \item \textbf{\opennessTextOpenOpenWeight\ (\opennessSymbolOpenOpenWeight):} Both agent code and ML model weights are publicly available, enabling full end-to-end reproducibility.
    \item \textbf{\opennessTextOpenClosedWeight\ (\opennessSymbolOpenClosedWeight):} Agent code is available but relies on proprietary ML models, allowing partial reproducibility of the approach.
    \item \textbf{\opennessTextClosedWithApi\ (\opennessSymbolClosedWithApi)}: Implementation details are proprietary, but the system is accessible via API, enabling result verification but not method reproduction.
    \item \textbf{\opennessTextClosed\ (\opennessSymbolClosed):} Neither code nor programmatic API access is available.
    \end{itemize}

\item \textbf{Agent tooling} describes the tool usage and execution environment of an agent during evaluation:

    \begin{itemize}
    \item \textbf{\toolingTextStandard\ (\toolingSymbolStandard):} Uses only predefined tools from the evaluation environment (as defined in Inspect's {\texttt state.tools}).
    \item \textbf{\toolingTextEquivalent\ (\toolingSymbolEquivalent):} Uses custom tools for accessing an equivalent underlying environment, which for \astabench we define as task-relevant portions of the \astaenvironment:
    \begin{itemize}
    \item \textbf{Literature tasks:} Information access is limited to date-restricted usage of the \toollitapi.
    \item \textbf{Code tasks:} Code execution is limited to an IPython shell in a machine environment initialized with the standard \astaenvironment sandbox Dockerfile (or equivalent).
    \end{itemize}
    \item \textbf{\toolingTextCustom\ (\toolingSymbolCustom):} Uses tools beyond constraints of Standard or Custom interface.
    \end{itemize}

\end{itemize}

\section{Supporting Experimental Results}
\label{sec:headline-results}

This section contains supplemental tables and figures for the narrative in \cref{sec:experiments}.
\cref{tab:results-overall} shows the overall results for those agents attempting {\it all} benchmarks, as well as agents that can solve all the benchmarks in at least one category. 
We then show category-specific results, for Literature Understanding (\cref{tab:results-lit-search,tab:results-lit-qa,tab:results-lit-table}), Code and Execution (\cref{tab:results-coding}), Data Analysis (\cref{tab:results-data}), and End-to-End Discovery (\cref{tab:results-endtoend}). For details about referenced agents and models, refer to \cref{tab:agents,tab:models}, respectively.

In the Tables, ``O'' denotes Openness, with values
\opennessSymbolOpenOpenWeight (\opennessTextOpenOpenWeight), 
\opennessSymbolOpenClosedWeight (\opennessTextOpenClosedWeight), 
\opennessSymbolClosedWithApi (\opennessTextClosedWithApi), and
\opennessSymbolClosed (\opennessTextClosed). ``T'' denotes Tooling, with values
\opennessSymbolOpenOpenWeight (\toolingTextStandard),
\opennessSymbolOpenClosedWeight (\toolingTextEquivalent), and 
\opennessSymbolClosed (\toolingTextCustom). The openness values apply to the agent (including the model used). ``$\pm$'' denote 95\% confidence intervals. \captionParetoNotice Our results reveal several noteworthy insights. %

\begin{table}[tp]
  \setlength{\tabcolsep}{2.1pt}
  \centering
  \scriptsize
  
  \caption{Overall results for agents that can solve all the tasks (additional results in \cref{apx:tab:results-overall}). 
  Reported values are macro averages over benchmark statistics; confidence intervals are omitted.
    \captionDaggerNotice
    \captionParetoNotice
    } 

  \rowcolors{8}{gray!10}{}
  \begin{tabular}{@{}l l L{2.1cm} L{2.4cm} C D C D C D C D C S[round-mode=places,round-precision=2,table-format=2.2{*}]@{}}
    \toprule
    O & T & Agent & Model & \multicolumn{2}{c}{Overall} & \multicolumn{2}{c}{\makecell{Literature\\Understanding}} & \multicolumn{2}{c}{\makecell{Code\\\& Execution}} & \multicolumn{2}{c}{\makecell{Data\\Analysis}} & \multicolumn{2}{c}{\makecell{End-to-End\\Discovery}} \\
    \cmidrule(lr){5-6} \cmidrule(lr){7-8} \cmidrule(lr){9-10} \cmidrule(lr){11-12} \cmidrule(lr){13-14}
    & & & & {Score} & {Cost} & {Score} & {Cost} & {Score} & {Cost} & {Score} & {Cost} & {Score} & {Cost} \\
    \midrule

\opennessSymbolOpenClosedWeight & \toolingSymbolCustom & \agentAsta & mixture & \B 52.9604 & \B 3.3950 & \B 62.2254 & \B 0.5792 & 47.5954 & 0.1896 & 33.1725 & 0.2463 & \B 68.8483 & \B 12.5650 \\ %
\opennessSymbolOpenClosedWeight & \toolingSymbolStandard & \agentReAct & \modelGPTFiveShort & \B 44.0311 & \B 0.3090 & \B 54.5726 & \B 0.2993 & \B 55.0173 & \B 0.3509 & 30.4808 & 0.0920 & 36.0537 & 0.4938 \\ %
\opennessSymbolOpenClosedWeight & \toolingSymbolStandard & \agentReAct & \modelOThreeShort & \B 39.4482 & \B 0.1635 & 46.8120 & 0.3529 & 49.3122 & 0.1917 & \B 33.6734 & \B 0.0392 & 27.9952 & 0.0702 \\ %
\opennessSymbolOpenClosedWeight & \toolingSymbolEquivalent & \agentSmolagents & \modelClaudeSonnetFourShort & 38.1363 & 1.0239 & 42.6663 & 0.7089 & 39.6001 & 1.9575 & 28.7968 & 0.2369 & 41.4821 & 1.1922 \\ %
\opennessSymbolOpenClosedWeight & \toolingSymbolEquivalent & \agentSmolagents & \modelGPTFiveShort & \B 37.5411 & \B 0.1265 & 45.9737 & 0.1151 & 30.9304 & 0.0958 & 26.7289 & 0.0770 & \B 46.5313 & \B 0.2182 \\ %
\opennessSymbolOpenClosedWeight & \toolingSymbolStandard & \agentReAct & \modelGPTFiveMiniShort & \B 31.6481 & \B 0.0366 & 36.5484 & 0.0489 & \B 50.5296 & \B 0.0517 & \B 26.9209 & \B 0.0111 & 12.5934 & 0.0347 \\ %
\opennessSymbolOpenClosedWeight & \toolingSymbolStandard & \agentReAct & \modelClaudeThreeFiveHaikuShort & \B 21.8866 & \B 0.0361 & 36.1903 & 0.0346 & 22.3951 & 0.0533 & 24.3174 & 0.0116 & 4.6436 & 0.0449 \\ %
\opennessSymbolOpenOpenWeight & \toolingSymbolEquivalent & \agentSmolagents & \modelLlamaFourScoutShort & 11.0526 & 0.1081 & 19.9890 & 0.0329 & 3.6049 & 0.1244 & \B 20.1634 & \B 0.0078 & 0.4531 & 0.2672 \\ %

    \bottomrule
  \end{tabular}
  \label{tab:results-overall}
\end{table}

\begin{table}[tp]
  \centering
  \scriptsize
  \renewcommand{\uncertaintysize}{\scriptsize}
  
  \caption{\catlit search benchmarks results (additional results in \cref{apx:tab:results-lit-search}). 
    \captionDaggerNotice
    \captionParetoNotice
  }
  \rowcolors{5}{}{gray!10}
  \begin{tabular}{@{}l l L{1.7cm} L{2.7cm} A B A B@{}}
    \toprule
    O & T & Agent & Model & \multicolumn{2}{c}{\evalpaperfinderNoSmall} & \multicolumn{2}{c}{\evallitqasearchftNoSmall} \\
    \cmidrule(lr){5-6} \cmidrule(lr){7-8}
    & & & & {Score} & {Cost} & {Score} & {Cost} \\
    \midrule

\opennessSymbolOpenClosedWeight & \toolingSymbolEquivalent & \agentPaperFinder & \modelGeminiTwoFlashShort, \modelGPTFourOShort & \B 39.7198 +- 3.0986 & \B 0.0626 +- 0.0052 & \B 90.6667 +- 6.6280 & \B 0.1116 +- 0.0071 \\ %
\opennessSymbolOpenClosedWeight & \toolingSymbolCustom & \agentAsta & mixture & 37.5748 +- 3.1429 & 0.0626 +- 0.0052 & 90.6667 +- 6.6280 & 0.1116 +- 0.0071 \\ %
\opennessSymbolOpenClosedWeight & \toolingSymbolStandard & \agentReAct & \modelGPTFiveShort & 26.3661 +- 3.8554 & 0.4279 +- 0.0484 & 82.6667 +- 8.6247 & 0.3886 +- 0.0552 \\ %
\opennessSymbolOpenClosedWeight & \toolingSymbolStandard & \agentReAct & \modelOThreeShort & 19.2781 +- 3.7267 & 0.5181 +- 0.0674 & 57.3333 +- 11.2691 & 0.7900 +- 0.1270 \\ %
\opennessSymbolOpenClosedWeight & \toolingSymbolEquivalent & \agentSmolagents & \modelGPTFourPointOneShort & 16.5279 +- 3.4743 & 0.0795 +- 0.0073 & 50.6667 +- 11.3913 & 0.0949 +- 0.0373 \\ %
\opennessSymbolOpenClosedWeight & \toolingSymbolEquivalent & \agentSmolagents & \modelClaudeSonnetFourShort & 22.1318 +- 3.4724 & 0.9752 +- 0.1385 & 52.0000 +- 11.3832 & 1.0995 +- 0.0965 \\ %
\opennessSymbolClosedWithApi & \toolingSymbolCustom & \agentYouComSearch & {--} & 7.2012 +- 1.9667 & {--} & 36.0000 +- 10.9366 & {--} \\ %

    \bottomrule
  \end{tabular}
  \label{tab:results-lit-search}
\end{table}

\begin{table}[tp]
  \setlength{\tabcolsep}{4pt}
  \centering
  \scriptsize
  \renewcommand{\uncertaintysize}{\scriptsize}

  \caption{\catlit QA benchmarks results (additional results in \cref{apx:tab:results-lit-qa}). 
  Agents without an API could not be evaluated on LitQA2-FT. 
    \captionDaggerNotice
    \captionParetoNotice
    }
  
  \rowcolors{5}{}{gray!10}
  \begin{tabular}{@{}l l L{2.8cm} L{2.8cm} A B A B@{}}
    \toprule
    O & T & Agent & Model & \multicolumn{2}{c}{\evalsqa} & \multicolumn{2}{c}{\evallitqaft} \\
    \cmidrule(lr){5-6} \cmidrule(lr){7-8}
    & & & & {Score} & {Cost} & {Score} & {Cost} \\
    \midrule

\opennessSymbolOpenClosedWeight & \toolingSymbolStandard & \agentReAct & \modelGPTFiveShort & 79.8396 +- 3.5004 & 0.3729 +- 0.0343 & \B 82.6667 +- 8.6247 & \B 0.2763 +- 0.1139 \\ %
\opennessSymbolOpenClosedWeight & \toolingSymbolCustom & \agentAsta & mixture & 87.7190 +- 1.4274 & 1.5286 +- 0.2911 & 70.6667 +- 10.3736 & 0.3061 +- 0.0931 \\ %
\opennessSymbolClosedWithApi & \toolingSymbolCustom & \agentFutureHouseCrow & \modelGPTFourPointOneMiniShort, \modelOThreeMiniShort, \modelGeminiTwoPointFiveFlashShort & 81.0570 +- 1.7413 & 0.1065 +- 0.0035 & 72.0000 +- 10.2302 & 0.0646 +- 0.0031 \\ %
\opennessSymbolClosedWithApi & \toolingSymbolCustom & \agentFutureHouseFalcon & \modelGPTFourPointOneMiniShort, \modelGeminiTwoPointFiveFlashShort, \modelOThreeMiniShort & 77.5945 +- 1.3240 & 0.4031 +- 0.0506 & 74.6667 +- 9.9095 & 0.2195 +- 0.0112 \\ %
\opennessSymbolOpenClosedWeight & \toolingSymbolStandard & \agentReAct & \modelOThreeShort & 66.3740 +- 2.9673 & 0.2753 +- 0.0385 & 80.0000 +- 9.1138 & 0.3465 +- 0.0833 \\ %
\opennessSymbolOpenClosedWeight & \toolingSymbolEquivalent & \agentSmolagents & \modelGPTFiveShort & 68.3871 +- 4.3921 & 0.1537 +- 0.0141 & 73.3333 +- 10.0757 & 0.1007 +- 0.0261 \\ %
\opennessSymbolClosedWithApi & \toolingSymbolCustom & \agentPerplexitySQA & \modelGeminiTwoPointFiveFlashShort, \modelPerplexitySonarDeepResearch & 67.2758 +- 1.1920 & 0.4163 +- 0.0192 & 73.3333 +- 10.0757 & 0.2185 +- 0.0158 \\ %
\opennessSymbolOpenClosedWeight & \toolingSymbolEquivalent & \agentSmolagents & \modelGPTFourPointOneShort & 73.6775 +- 2.0892 & 0.0795 +- 0.0162 & \B 65.3333 +- 10.8434 & \B 0.0348 +- 0.0047 \\ %
\opennessSymbolClosedWithApi & \toolingSymbolCustom & \agentYouComResearch & {--} & 55.0019 +- 2.1562 & {--} & 8.0000 +- 6.1813 & {--} \\ %

\midrule

\opennessSymbolOpenClosedWeight & \toolingSymbolEquivalent & \agentScholarQA & \modelClaudeSonnetFourShort & 87.9147 +- 1.2270 & 1.3137 +- 0.2812 & {--} & {--} \\ %
\opennessSymbolOpenClosedWeight & \toolingSymbolEquivalent & \agentScholarQANoTables & \modelGeminiTwoPointFiveFlashUnpinned & \B 87.6541 +- 1.4316 & \B 0.1263 +- 0.0095 & {--} & {--} \\ %
\opennessSymbolOpenClosedWeight & \toolingSymbolEquivalent & \agentScholarQANoTables & \modelClaudeSonnetFourShort & 86.2425 +- 1.4041 & 0.3931 +- 0.0303 & {--} & {--} \\ %
\opennessSymbolOpenClosedWeight & \toolingSymbolEquivalent & \agentScholarQANoTables & \modelGPTFiveUnpinned & 85.9128 +- 1.5605 & 1.0993 +- 0.0736 & {--} & {--} \\ %
\opennessSymbolClosed & \toolingSymbolCustom & \agentElicit & {--} & 85.5343 +- 1.6108 & {--} & {--} & {--} \\ %
\opennessSymbolClosed & \toolingSymbolCustom & \agentSciSpace       & \modelClaudeSonnetFourShort & 84.6 +- 1.3 & {--} & {--} & {--} \\
\opennessSymbolOpenClosedWeight & \toolingSymbolCustom & \agentSTORM & \modelGPTThreeFiveTurboShort, \modelGPTFourOShort & 78.3020 +- 2.3597 & 0.0941 +- 0.0022 & {--} & {--} \\ %
\opennessSymbolClosedWithApi & \toolingSymbolCustom & \agentOpenAIDeepResearch & \modelDRAmbiguous, \modelGeminiTwoPointFiveProShort & 79.3958 +- 1.4246 & 1.8033 +- 0.0386 & {--} & {--} \\ %
\opennessSymbolOpenOpenWeight & \toolingSymbolEquivalent & \agentOpenScholar & \modelOpenScholar & \B 57.9581 +- 2.5860 & \B 0.0040 +- 0.0001 & {--} & {--} \\ %

    \bottomrule
  \end{tabular}
  \label{tab:results-lit-qa}
\end{table}

\begin{table}[tp]
  \centering
  \small
  
  \caption{\catlit \evaltables task benchmark results (additional results in \cref{apx:tab:results-lit-table}). 
    \captionDaggerNotice
    \captionParetoNotice
  }

  \rowcolors{5}{gray!10}{}
  \begin{tabular}{@{}l l l l A B@{}}
    \toprule
    O & T & Agent & Model & \multicolumn{2}{c}{\evaltables} \\
    \cmidrule(lr){5-6}
    & & & & {Score} & {Cost} \\
    \midrule

\opennessSymbolOpenClosedWeight & \toolingSymbolCustom & \agentAsta & mixture & \B 42.9411 +- 3.7244 & \B 0.5167 +- 0.0561 \\ %
\opennessSymbolOpenClosedWeight & \toolingSymbolEquivalent & \agentAstaTableAgent & \modelGPTFiveUnpinned & 42.6085 +- 3.4968 & 1.2812 +- 0.1395 \\ %
\opennessSymbolOpenClosedWeight & \toolingSymbolEquivalent & \agentAstaTableAgent & \modelGPTFiveMiniUnpinned & \B 41.7115 +- 3.6991 & \B 0.1720 +- 0.0189 \\ %
\opennessSymbolOpenClosedWeight & \toolingSymbolStandard & \agentReAct & \modelOThreeShort & 32.9295 +- 3.3195 & 0.0501 +- 0.0040 \\ %
\opennessSymbolOpenClosedWeight & \toolingSymbolEquivalent & \agentSmolagents & \modelGPTFiveShort & 31.5124 +- 3.2385 & 0.0599 +- 0.0042 \\ %

    \bottomrule
  \end{tabular}
  \label{tab:results-lit-table}
\end{table}

\begin{table}[tp]
  \setlength{\tabcolsep}{1pt}
  \centering
  \scriptsize
  \renewcommand{\uncertaintysize}{\scriptsize}

  \caption{\catcoding category results (additional results in \cref{apx:tab:results-coding}). 
    \captionDaggerNotice
    \captionParetoNotice
  }

  \rowcolors{6}{gray!10}{}
  \begin{tabular}{@{}l l L{1.9cm} L{1.7cm} A B A B A U{table-format=2.3,round-mode=places,round-precision=3}{table-format=1.4{*},round-mode=places,round-precision=4}@{}}
    \toprule
    O & T & Agent & Model & \multicolumn{2}{c}{\evalsuper} & \multicolumn{2}{c}{\evalcorebench} & \multicolumn{2}{c}{\evaldatasci} \\
    \cmidrule(lr){5-6} \cmidrule(lr){7-8} \cmidrule(lr){9-10}
    & & & & {Score} & {Cost} & {Score} & {Cost} & {Score} & {Cost} \\
    \midrule

\opennessSymbolOpenClosedWeight & \toolingSymbolStandard & \agentReAct & \modelGPTFiveShort & \B 41.1058 +- 12.8578 & \B 0.5890 +- 0.1400 & 45.9459 +- 16.2796 & 0.4427 +- 0.1388 & \B 78.0000 +- 2.7079 & \B 0.0210 +- 0.0009 \\ %
\opennessSymbolOpenClosedWeight & \toolingSymbolStandard & \agentReAct & \modelGPTFiveMiniShort & \B 34.6429 +- 13.2074 & \B 0.1049 +- 0.0461 & \B 45.9459 +- 16.2796 & \B 0.0472 +- 0.0143 & \B 71.0000 +- 2.9662 & \B 0.0030 +- 0.0001 \\ %
\opennessSymbolOpenClosedWeight & \toolingSymbolStandard & \agentReAct & \modelOThreeShort & 16.2910 +- 9.5855 & 0.3690 +- 0.0969 & \B 56.7568 +- 16.1835 & \B 0.1957 +- 0.0763 & \B 74.8889 +- 2.8348 & \B 0.0102 +- 0.0007 \\ %
\opennessSymbolOpenClosedWeight & \toolingSymbolCustom & \agentAsta & mixture & 19.3598 +- 10.3855 & 0.3317 +- 0.0570 & 48.6486 +- 16.3274 & 0.2260 +- 0.0933 & 74.7778 +- 2.8389 & 0.0110 +- 0.0007 \\ %
\opennessSymbolOpenClosedWeight & \toolingSymbolEquivalent & \agentSmolagents & \modelClaudeSonnetFourShort & 11.7011 +- 8.0336 & 3.5592 +- 1.7661 & 32.4324 +- 15.2920 & 2.1994 +- 0.7799 & 74.6667 +- 2.8431 & 0.1140 +- 0.0079 \\ %
\opennessSymbolOpenClosedWeight & \toolingSymbolEquivalent & \agentSmolagents & \modelGPTFiveShort & 3.6111 +- 4.8391 & 0.0793 +- 0.0228 & 13.5135 +- 11.1677 & 0.1898 +- 0.1059 & 75.6667 +- 2.8050 & 0.0185 +- 0.0007 \\ %
\opennessSymbolOpenClosedWeight & \toolingSymbolEquivalent & \agentSmolagents & \modelClaudeThreeFiveHaikuShort & 16.7540 +- 9.6495 & 0.8120 +- 0.5808 & 0.0000 & 0.3323 +- 0.2095 & 9.8889 +- 1.9514 & 0.0237 +- 0.0103 \\ %

\midrule

\opennessSymbolOpenClosedWeight & \toolingSymbolEquivalent & \agentAstaCode & \modelGPTFourPointOneShort & 16.2540 +- 9.4464 & 0.2850 +- 0.0591 & {--} & {--} & {--} & {--} \\ %
\opennessSymbolOpenClosedWeight & \toolingSymbolEquivalent & \agentAstaCode & \modelGPTFiveShort & 13.5132 +- 9.3913 & 0.3715 +- 0.0717 & {--} & {--} & {--} & {--} \\ %

    \bottomrule
  \end{tabular}
  \label{tab:results-coding}
\end{table}

\begin{table}[tp]
  \centering
  \small

  \caption{\catdata \evaldiscoverybench results (additional results in \cref{apx:tab:results-data}). 
    \captionDaggerNotice
    \captionParetoNotice
  }

  \rowcolors{8}{gray!10}{}
  \begin{tabular}{@{}l l l l A B@{}}
    \toprule
    O & T & Agent & Model & \multicolumn{2}{c}{\evaldiscoverybench} \\
    \cmidrule(lr){5-6}
    & & & & {Score} & {Cost} \\
    \midrule

\opennessSymbolOpenClosedWeight & \toolingSymbolStandard & \agentReAct & \modelOThreeShort & \B 33.6734 +- 5.0963 & \B 0.0392 +- 0.0043 \\ %
\opennessSymbolOpenClosedWeight & \toolingSymbolCustom & \agentAsta & mixture & 33.1725 +- 5.0587 & 0.2463 +- 0.0710 \\ %
\opennessSymbolOpenClosedWeight & \toolingSymbolEquivalent & \agentDataVoyager & \modelOThreeUnpinned, \modelGPTFourOUnpinned & 31.1268 +- 5.0476 & 0.2335 +- 0.0614 \\ %
\opennessSymbolOpenClosedWeight & \toolingSymbolStandard & \agentReAct & \modelGPTFiveShort & 30.4808 +- 4.8361 & 0.0920 +- 0.0091 \\ %
\opennessSymbolOpenClosedWeight & \toolingSymbolEquivalent & \agentSmolagents & \modelClaudeSonnetFourShort & 28.7968 +- 4.7879 & 0.2369 +- 0.0190 \\ %

    \bottomrule
  \end{tabular}
  \label{tab:results-data}
\end{table}

\begin{table}[tp]
  \setlength{\tabcolsep}{3pt}
  \centering
  \small
  \renewcommand{\uncertaintysize}{\scriptsize}

  \caption{\catendtoend category results (additional results in \cref{apx:tab:results-endtoend}). 
    \captionDaggerNotice
    \captionParetoNotice
  }

  \rowcolors{5}{}{gray!10}
    \begin{tabular}{@{}l l L{2.7cm} L{2.2cm} A U{table-format=2.3,round-mode=places,round-precision=3}{table-format=1.3{*},round-mode=places,round-precision=3} A U{table-format=2.3,round-mode=places,round-precision=3}{table-format=1.3{*},round-mode=places,round-precision=3}@{}}
    \toprule
    O & T & Agent & Model & \multicolumn{2}{c}{\evalendtoend} & \multicolumn{2}{c}{\evalendtoendhard} \\
    \cmidrule(lr){5-6} \cmidrule(lr){7-8}
    & & & & {Score} & {Cost} & {Score} & {Cost} \\
    \midrule

\opennessSymbolOpenClosedWeight & \toolingSymbolCustom & \agentAutoAsta & \modelClaudeSonnetFourShort & \B 70.4709 +- 6.2304 & \B 10.6427 +- 0.7170 & \B 68.2114 +- 4.3677 & \B 14.4873 +- 1.0501 \\ %
\opennessSymbolOpenClosedWeight & \toolingSymbolCustom & \agentAsta & mixture & \B 70.4125 +- 6.2983 & \B 10.6427 +- 0.7170 & \B 67.2842 +- 5.2905 & \B 14.4873 +- 1.0501 \\ %
\opennessSymbolOpenClosedWeight & \toolingSymbolCustom & \agentCodeScientist & \modelClaudeSonnetThreeSevenShort & \B 65.3458 +- 7.1086 & \B 2.7598 +- 0.5100 & \B 64.4985 +- 5.5100 & \B 3.5487 +- 0.6921 \\ %
\opennessSymbolOpenClosedWeight & \toolingSymbolEquivalent & \agentSmolagents & \modelGPTFiveShort & \B 62.7545 +- 9.8341 & \B 0.2047 +- 0.0246 & 30.3081 +- 10.5246 & 0.2317 +- 0.0430 \\ %
\opennessSymbolOpenClosedWeight & \toolingSymbolStandard & \agentReAct & \modelClaudeSonnetFourShort & 52.5225 +- 6.7948 & 0.7493 +- 0.0720 & 38.8998 +- 6.8844 & 0.8355 +- 0.0574 \\ %
\opennessSymbolOpenClosedWeight & \toolingSymbolEquivalent & \agentSmolagents & \modelClaudeSonnetFourShort & 47.1630 +- 6.0675 & 0.8729 +- 0.1097 & 35.8012 +- 7.7854 & 1.5115 +- 0.3067 \\ %
\opennessSymbolOpenClosedWeight & \toolingSymbolStandard & \agentFaker & \modelGPTFourPointOneUnpinned & \B 39.1740 +- 6.9341 & \B 0.0263 +- 0.0013 & \B 25.3660 +- 4.5482 & \B 0.0288 +- 0.0011 \\ %
\opennessSymbolOpenClosedWeight & \toolingSymbolStandard & \agentReAct & \modelOThreeShort & 34.9483 +- 10.1077 & 0.0652 +- 0.0103 & 21.0421 +- 7.6100 & 0.0751 +- 0.0189 \\ %
\opennessSymbolOpenClosedWeight & \toolingSymbolStandard & \agentReAct & \modelGPTFiveShort & 30.0039 +- 11.9339 & 0.4034 +- 0.0526 & \B 42.1035 +- 11.4110 & \B 0.5842 +- 0.0721 \\ %

    \bottomrule
  \end{tabular}
  \label{tab:results-endtoend}
\end{table}

\section{Full Experimental Results}
\label{sec:appendix:full-results}

\cref{sec:experiments} presented results for the best agents (i.e., agents running with the best underlying model), plus a few additional important data points. Here we show the full set of results for all configurations of agents that were tested
(a superset of the results in \cref{sec:experiments}). We also show plots of scores vs. costs, including the Pareto frontier (showing the best agent for a given cost). 
In the Tables, ``O'' denotes Openness, with values
\opennessSymbolOpenOpenWeight (\opennessTextOpenOpenWeight), 
\opennessSymbolOpenClosedWeight (\opennessTextOpenClosedWeight), and
\opennessSymbolClosed (\opennessTextClosed). ``T'' denotes Tooling, with values
\opennessSymbolOpenOpenWeight (\toolingTextStandard),
\opennessSymbolOpenClosedWeight (\toolingTextEquivalent), and 
\opennessSymbolClosed (\toolingTextCustom). ``$\pm$'' denote 95\% confidence intervals.

\paragraph{Statistical Methodology}
All confidence intervals shown are 95\% CIs computed as $\pm 1.96 \times \text{SE}$, where SE is the standard error.
For individual benchmarks, standard errors are calculated from the variance across evaluation samples within each task.
For category-level aggregations, standard errors are propagated analytically using weighted averaging: $\text{SE}_{\text{category}} = \sqrt{\sum w_i^2 \cdot \text{SE}_i^2} / \sum w_i$, where $w_i$ are the task weights (uniform at 1.0 except for the two LitQA tasks which each have weight 0.5).
This propagation assumes independence between tasks, which could slightly underestimate uncertainty.

\begin{table}[tp]
  \setlength{\tabcolsep}{2.1pt}
  \centering
  \tiny
    \renewcommand{\uncertaintysize}{\tiny}

  \caption{Overall results for agents that can solve all the tasks. 
  Reported values are macro averages over benchmark statistics; cost confidence intervals are omitted for space.
    \captionDaggerNotice
    } 

  \rowcolors{8}{gray!10}{}
  \begin{tabular}{@{}l l L{1.4cm} L{1.4cm} A D A D A D A D A D}
    \toprule
    O & T & Agent & Model & \multicolumn{2}{c}{Overall} & \multicolumn{2}{c}{\makecell{Literature\\Understanding}} & \multicolumn{2}{c}{\makecell{Code\\\& Execution}} & \multicolumn{2}{c}{\makecell{Data\\Analysis}} & \multicolumn{2}{c}{\makecell{End-to-End\\Discovery}} \\
    \cmidrule(lr){5-6} \cmidrule(lr){7-8} \cmidrule(lr){9-10} \cmidrule(lr){11-12} \cmidrule(lr){13-14}
    & & & & {Score} & {Cost} & {Score} & {Cost} & {Score} & {Cost} & {Score} & {Cost} & {Score} & {Cost} \\
    \midrule
\opennessSymbolOpenClosedWeight & \toolingSymbolStandard & \agentReAct & \modelClaudeThreeFiveHaikuShort & \B 21.8866 +- 1.5947 & \B 0.0361 +- 0.0030 & 36.1903 +- 2.2521 & 0.0346 +- 0.0016 & 22.3951 +- 2.9698 & 0.0533 +- 0.0090 & 24.3174 +- 4.6993 & 0.0116 +- 0.0010 & 4.6436 +- 2.1718 & 0.0449 +- 0.0079 \\ %
\opennessSymbolOpenClosedWeight & \toolingSymbolStandard & \agentReAct & \modelClaudeSonnetFourShort & 40.1286 +- 2.3607 & 0.4026 +- 0.0155 & 45.3608 +- 2.3080 & 0.3554 +- 0.0094 & 46.2202 +- 6.5761 & 0.3301 +- 0.0397 & 23.2221 +- 4.1479 & 0.1323 +- 0.0085 & 45.7112 +- 4.8364 & 0.7924 +- 0.0460 \\ %
\opennessSymbolOpenClosedWeight & \toolingSymbolStandard & \agentReAct & \modelGPTFourPointOneShort & 31.5913 +- 2.2527 & 0.1985 +- 0.0224 & 46.3567 +- 2.2674 & 0.5395 +- 0.0834 & 32.3821 +- 5.0519 & 0.0942 +- 0.0259 & \B 30.5362 +- 5.0581 & \B 0.0248 +- 0.0030 & 17.0901 +- 4.9950 & 0.1356 +- 0.0205 \\ %
\opennessSymbolOpenClosedWeight & \toolingSymbolStandard & \agentReAct & \modelGPTFourOShort & 16.2210 +- 1.4327 & 0.1223 +- 0.0096 & 31.8013 +- 2.2316 & 0.1524 +- 0.0133 & 18.3327 +- 3.4871 & 0.1512 +- 0.0267 & 13.2332 +- 3.7464 & 0.0397 +- 0.0095 & 1.5170 +- 1.2904 & 0.1459 +- 0.0225 \\ %
\opennessSymbolOpenClosedWeight & \toolingSymbolStandard & \agentReAct & \modelGPTFiveMiniShort & \B 31.6481 +- 2.6534 & \B 0.0366 +- 0.0044 & 36.5484 +- 2.9167 & 0.0489 +- 0.0054 & \B 50.5296 +- 7.0574 & \B 0.0517 +- 0.0161 & \B 26.9209 +- 4.7829 & \B 0.0111 +- 0.0012 & 12.5934 +- 5.6087 & 0.0347 +- 0.0050 \\ %
\opennessSymbolOpenClosedWeight & \toolingSymbolStandard & \agentReAct & \modelGPTFiveShort & \B 44.0311 +- 3.0112 & \B 0.3090 +- 0.0207 & \B 54.5726 +- 2.2129 & \B 0.2993 +- 0.0217 & \B 55.0173 +- 6.9736 & \B 0.3509 +- 0.0657 & 30.4808 +- 4.8361 & 0.0920 +- 0.0091 & 36.0537 +- 8.2558 & 0.4938 +- 0.0446 \\ %
\opennessSymbolOpenClosedWeight & \toolingSymbolStandard & \agentReAct & \modelGeminiTwoPointFiveFlashShort & 15.3260 +- 1.3618 & 0.7107 +- 0.1702 & 32.7917 +- 2.9598 & 0.4559 +- 0.0755 & 26.0490 +- 4.1192 & 0.4544 +- 0.1216 & 1.9277 +- 1.6859 & 0.1006 +- 0.0065 & 0.5357 +- 1.0500 & 1.8319 +- 0.6655 \\ %
\opennessSymbolOpenOpenWeight & \toolingSymbolStandard & \agentReAct & \modelLlamaFourScoutShort & 7.9488 +- 1.0515 & 0.6784 +- 0.0426 & 19.7665 +- 2.4748 & 1.6030 +- 0.1355 & 4.7963 +- 1.8616 & 0.1039 +- 0.0228 & 5.8621 +- 2.5846 & 0.1917 +- 0.0208 & 1.3705 +- 1.1916 & 0.8152 +- 0.0985 \\ %
\opennessSymbolOpenClosedWeight & \toolingSymbolStandard & \agentReAct & \modelOThreeShort & \B 39.4482 +- 2.6408 & \B 0.1635 +- 0.0127 & 46.8120 +- 2.3214 & 0.3529 +- 0.0272 & 49.3122 +- 6.3406 & 0.1917 +- 0.0411 & \B 33.6734 +- 5.0963 & \B 0.0392 +- 0.0043 & 27.9952 +- 6.3261 & 0.0702 +- 0.0108 \\ %
\opennessSymbolOpenClosedWeight & \toolingSymbolEquivalent & \agentSmolagents & \modelClaudeThreeFiveHaikuShort & 12.6963 +- 1.4865 & 0.2974 +- 0.1099 & 20.8938 +- 1.9270 & 0.0512 +- 0.0082 & 8.8810 +- 3.2816 & 0.3893 +- 0.2058 & 16.5142 +- 4.1274 & 0.0237 +- 0.0071 & 4.4960 +- 1.9588 & 0.7253 +- 0.3885 \\ %
\opennessSymbolOpenClosedWeight & \toolingSymbolEquivalent & \agentSmolagents & \modelClaudeSonnetFourShort & 38.1363 +- 2.3308 & 1.0239 +- 0.1663 & 42.6663 +- 2.3641 & 0.7089 +- 0.0400 & 39.6001 +- 5.8354 & 1.9575 +- 0.6435 & 28.7968 +- 4.7879 & 0.2369 +- 0.0190 & 41.4821 +- 4.9352 & 1.1922 +- 0.1629 \\ %
\opennessSymbolOpenClosedWeight & \toolingSymbolEquivalent & \agentSmolagents & \modelGPTFourPointOneShort & 32.7869 +- 2.4011 & 0.3212 +- 0.2229 & \B 43.8569 +- 2.3438 & \B 0.0668 +- 0.0066 & 25.5529 +- 5.1582 & 0.1066 +- 0.0567 & 28.3999 +- 4.8659 & 0.0451 +- 0.0183 & 33.3379 +- 6.0389 & 1.0663 +- 0.8895 \\ %
\opennessSymbolOpenClosedWeight & \toolingSymbolEquivalent & \agentSmolagents & \modelGPTFourOShort & 13.6156 +- 1.5524 & 0.3588 +- 0.1248 & 22.6834 +- 2.1147 & 0.0762 +- 0.0049 & 8.7013 +- 3.0670 & 0.6356 +- 0.2756 & 17.8158 +- 4.2452 & 0.0540 +- 0.0041 & 5.2618 +- 2.5800 & 0.6693 +- 0.4161 \\ %
\opennessSymbolOpenClosedWeight & \toolingSymbolEquivalent & \agentSmolagents & \modelGPTFiveMiniShort & 29.0925 +- 2.2963 & 0.0648 +- 0.0280 & \B 38.4677 +- 2.6629 & \B 0.0221 +- 0.0033 & 28.2674 +- 3.9838 & 0.0902 +- 0.0691 & 27.6804 +- 4.9305 & 0.0711 +- 0.0407 & 21.9545 +- 6.0908 & 0.0759 +- 0.0784 \\ %
\opennessSymbolOpenClosedWeight & \toolingSymbolEquivalent & \agentSmolagents & \modelGPTFiveShort & \B 37.5411 +- 2.4745 & \B 0.1265 +- 0.0112 & 45.9737 +- 2.5276 & 0.1151 +- 0.0067 & 30.9304 +- 4.1634 & 0.0958 +- 0.0361 & 26.7289 +- 4.7303 & 0.0770 +- 0.0057 & \B 46.5313 +- 7.2020 & \B 0.2182 +- 0.0248 \\ %
\opennessSymbolOpenClosedWeight & \toolingSymbolEquivalent & \agentSmolagents & \modelGeminiTwoPointFiveFlashShort & 26.3847 +- 2.3483 & 0.7088 +- 0.2076 & 35.5625 +- 2.5259 & 0.0515 +- 0.0168 & 16.6341 +- 4.3321 & 0.5571 +- 0.3942 & 24.7251 +- 4.7004 & 0.0172 +- 0.0068 & 28.6170 +- 6.4023 & 2.2093 +- 0.7309 \\ %
\opennessSymbolOpenOpenWeight & \toolingSymbolEquivalent & \agentSmolagents & \modelLlamaFourScoutShort & 11.0526 +- 1.3895 & 0.1081 +- 0.0436 & 19.9890 +- 2.1834 & 0.0329 +- 0.0218 & 3.6049 +- 2.3574 & 0.1244 +- 0.1261 & \B 20.1634 +- 4.5180 & \B 0.0078 +- 0.0018 & 0.4531 +- 0.3918 & 0.2672 +- 0.1182 \\ %

\opennessSymbolOpenClosedWeight & \toolingSymbolCustom & \agentAsta & mixture & \B 52.9604 +- 2.3583 & \B 3.3950 +- 0.1613 & \B 62.2254 +- 1.9949 & \B 0.5792 +- 0.0750 & 47.5954 +- 6.5192 & 0.1896 +- 0.0365 & 33.1725 +- 5.0587 & 0.2463 +- 0.0710 & \B 68.8483 +- 4.1127 & \B 12.5650 +- 0.6358 \\ %
    \bottomrule
  \end{tabular}
  \label{apx:tab:results-overall}
\end{table}

\begin{table}[tp]
  \centering
  \scriptsize
  \renewcommand{\uncertaintysize}{\scriptsize}
  
  \caption{\catlit search benchmarks results. 
    \captionDaggerNotice
  }
  \rowcolors{5}{}{gray!10}
  \begin{tabular}{@{}l l L{2.0cm} L{2.5cm} A B A B@{}}
    \toprule
    O & T & Agent & Model & \multicolumn{2}{c}{\evalpaperfinderNoSmall} & \multicolumn{2}{c}{\evallitqasearchftNoSmall} \\
    \cmidrule(lr){5-6} \cmidrule(lr){7-8}
    & & & & {Score} & {Cost} & {Score} & {Cost} \\
    \midrule
\opennessSymbolOpenClosedWeight & \toolingSymbolStandard & \agentReAct & \modelClaudeThreeFiveHaikuShort & 10.6964 +- 2.6060 & 0.0612 +- 0.0047 & 60.0000 +- 11.1621 & 0.0693 +- 0.0070 \\ %
\opennessSymbolOpenClosedWeight & \toolingSymbolStandard & \agentReAct & \modelClaudeSonnetFourShort & 20.2560 +- 3.1804 & 0.5406 +- 0.0250 & 46.6667 +- 11.3669 & 0.6062 +- 0.0306 \\ %
\opennessSymbolOpenClosedWeight & \toolingSymbolStandard & \agentReAct & \modelGPTFourPointOneShort & 16.4740 +- 3.1986 & 0.8669 +- 0.1828 & 65.3333 +- 10.8434 & 0.8187 +- 0.2576 \\ %
\opennessSymbolOpenClosedWeight & \toolingSymbolStandard & \agentReAct & \modelGPTFourOShort & 12.9320 +- 3.3242 & 0.2671 +- 0.0315 & 66.6667 +- 10.7407 & 0.3283 +- 0.0805 \\ %
\opennessSymbolOpenClosedWeight & \toolingSymbolStandard & \agentReAct & \modelGPTFiveMiniShort & 22.0395 +- 3.7476 & 0.0595 +- 0.0087 & 56.0000 +- 11.3099 & 0.1180 +- 0.0261 \\ %
\opennessSymbolOpenClosedWeight & \toolingSymbolStandard & \agentReAct & \modelGPTFiveShort & 26.3661 +- 3.8554 & 0.4279 +- 0.0484 & 82.6667 +- 8.6247 & 0.3886 +- 0.0552 \\ %
\opennessSymbolOpenClosedWeight & \toolingSymbolStandard & \agentReAct & \modelGeminiTwoPointFiveFlashShort & 6.5049 +- 2.2606 & 1.1959 +- 0.2135 & 57.3333 +- 11.2691 & 0.6499 +- 0.4000 \\ %
\opennessSymbolOpenOpenWeight & \toolingSymbolStandard & \agentReAct & \modelLlamaFourScoutShort & 5.4086 +- 2.2142 & 2.8155 +- 0.3194 & 37.3333 +- 11.0206 & 4.3262 +- 0.7950 \\ %
\opennessSymbolOpenClosedWeight & \toolingSymbolStandard & \agentReAct & \modelOThreeShort & 19.2781 +- 3.7267 & 0.5181 +- 0.0674 & 57.3333 +- 11.2691 & 0.7900 +- 0.1270 \\ %
\opennessSymbolOpenClosedWeight & \toolingSymbolEquivalent & \agentSmolagents & \modelClaudeThreeFiveHaikuShort & 4.6424 +- 2.0854 & 0.0701 +- 0.0112 & 2.6667 +- 3.6708 & 0.0961 +- 0.0597 \\ %
\opennessSymbolOpenClosedWeight & \toolingSymbolEquivalent & \agentSmolagents & \modelClaudeSonnetFourShort & 22.1318 +- 3.4724 & 0.9752 +- 0.1385 & 52.0000 +- 11.3832 & 1.0995 +- 0.0965 \\ %
\opennessSymbolOpenClosedWeight & \toolingSymbolEquivalent & \agentSmolagents & \modelGPTFourPointOneShort & 16.5279 +- 3.4743 & 0.0795 +- 0.0073 & 50.6667 +- 11.3913 & 0.0949 +- 0.0373 \\ %
\opennessSymbolOpenClosedWeight & \toolingSymbolEquivalent & \agentSmolagents & \modelGPTFourOShort & 12.4984 +- 3.5218 & 0.0980 +- 0.0102 & 20.0000 +- 9.1138 & 0.1048 +- 0.0232 \\ %
\opennessSymbolOpenClosedWeight & \toolingSymbolEquivalent & \agentSmolagents & \modelGPTFiveMiniShort & 17.1772 +- 3.2367 & 0.0335 +- 0.0118 & \B 48.0000 +- 11.3832 & \B 0.0363 +- 0.0104 \\ %
\opennessSymbolOpenClosedWeight & \toolingSymbolEquivalent & \agentSmolagents & \modelGPTFiveShort & 19.9954 +- 3.8599 & 0.1205 +- 0.0116 & 54.6667 +- 11.3425 & 0.1519 +- 0.0288 \\ %
\opennessSymbolOpenClosedWeight & \toolingSymbolEquivalent & \agentSmolagents & \modelGeminiTwoPointFiveFlashShort & 14.7166 +- 3.4093 & 0.0443 +- 0.0055 & 36.0000 +- 10.9366 & 0.0888 +- 0.1000 \\ %
\opennessSymbolOpenOpenWeight & \toolingSymbolEquivalent & \agentSmolagents & \modelLlamaFourScoutShort & \B 7.0248 +- 2.8587 & \B 0.0128 +- 0.0025 & \B 6.6667 +- 5.6835 & \B 0.0098 +- 0.0013 \\ %

\opennessSymbolOpenClosedWeight & \toolingSymbolCustom & \agentAsta & mixture & 37.5748 +- 3.1429 & 0.0626 +- 0.0052 & 90.6667 +- 6.6280 & 0.1116 +- 0.0071 \\ %

    \addlinespace[0.5em]
\opennessSymbolOpenClosedWeight & \toolingSymbolEquivalent & \agentPaperFinder & \modelGeminiTwoFlashShort, \modelGPTFourOShort & \B 39.7198 +- 3.0986 & \B 0.0626 +- 0.0052 & \B 90.6667 +- 6.6280 & \B 0.1116 +- 0.0071 \\ %
\opennessSymbolClosedWithApi & \toolingSymbolCustom & \agentYouComSearch & \missing & 7.2012 +- 1.9667 & \missing & 36.0000 +- 10.9366 & \missing \\ %
    \bottomrule
  \end{tabular}
  \label{apx:tab:results-lit-search}
\end{table}

\begin{figure}[htbp]
  \centering
  \includegraphics[width=\textwidth]{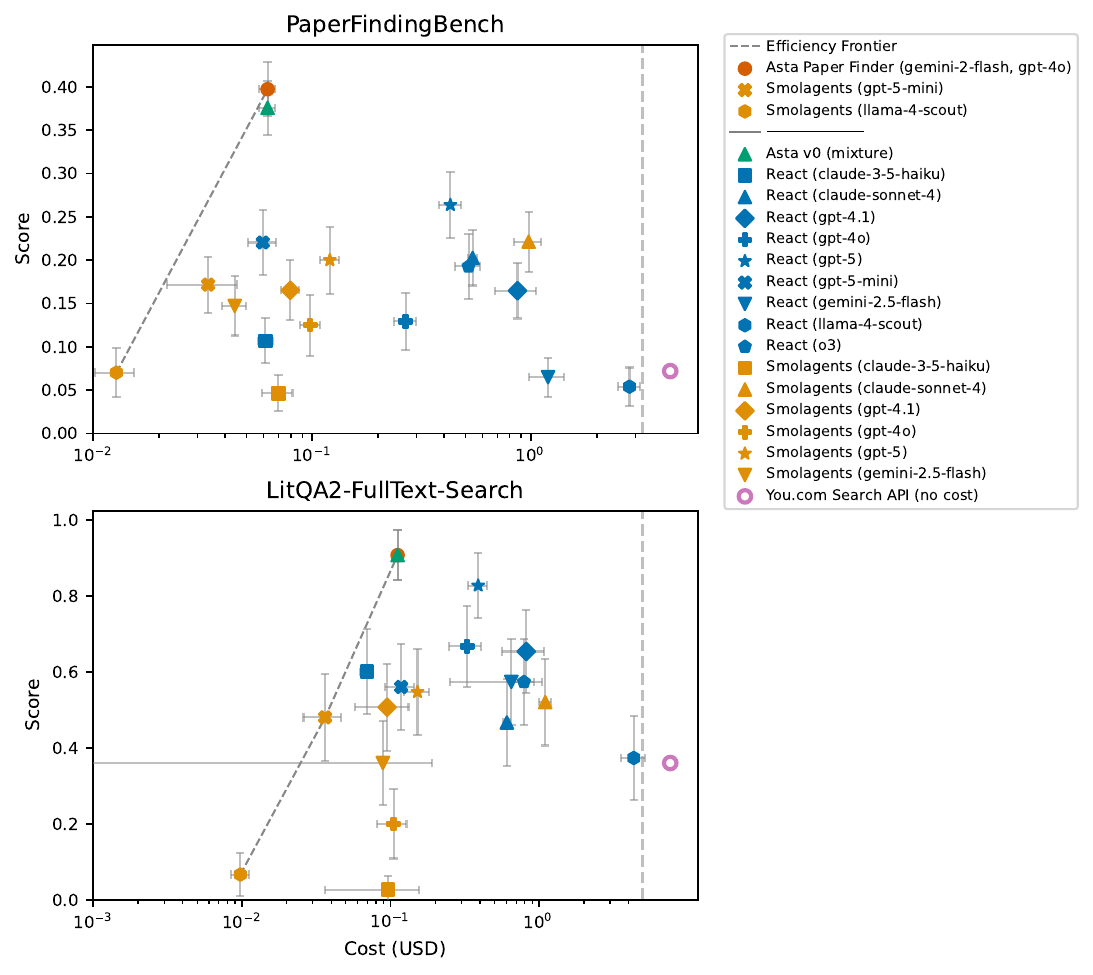}
  \caption{Score vs. cost analysis for \catlit{} search benchmarks (\cref{apx:tab:results-lit-search}). Points indicate means; error bars denote 95\% confidence intervals. Points on the Pareto frontier are connected with dotted lines, representing optimal quality-cost trade-offs for each eval (\evalpaperfinder, \evallitqasearchft). Note: the x-axis (cost) uses a log scale.}
  \label{apx:fig:results-lit-search}
\end{figure}

\begin{table}[tp]
  \setlength{\tabcolsep}{4pt}
  \centering
  \scriptsize
  \renewcommand{\uncertaintysize}{\scriptsize}

  \caption{\catlit QA benchmarks results. 
  Agents without an API could not be evaluated on LitQA2-FT. 
    Models in parentheses indicate self-reported models.
    \captionDaggerNotice
    }

  \rowcolors{5}{}{gray!10}
  \begin{tabular}{@{}l l L{3.1cm} L{2.4cm} A B A B@{}}
    \toprule
    O & T & Agent & Model & \multicolumn{2}{c}{\evalsqa} & \multicolumn{2}{c}{\evallitqaft} \\
    \cmidrule(lr){5-6} \cmidrule(lr){7-8}
    & & & & {Score} & {Cost} & {Score} & {Cost} \\
    \midrule
\opennessSymbolOpenClosedWeight & \toolingSymbolStandard & \agentReAct & \modelClaudeThreeFiveHaikuShort & 66.3264 +- 2.8299 & 0.0190 +- 0.0009 & 32.0000 +- 10.6284 & 0.0218 +- 0.0035 \\ %
\opennessSymbolOpenClosedWeight & \toolingSymbolStandard & \agentReAct & \modelClaudeSonnetFourShort & 78.3049 +- 2.1739 & 0.3902 +- 0.0187 & 68.0000 +- 10.6284 & 0.2383 +- 0.0257 \\ %
\opennessSymbolOpenClosedWeight & \toolingSymbolStandard & \agentReAct & \modelGPTFourPointOneShort & 70.1332 +- 3.1534 & 0.7326 +- 0.2428 & 77.3333 +- 9.5393 & 0.2219 +- 0.0970 \\ %
\opennessSymbolOpenClosedWeight & \toolingSymbolStandard & \agentReAct & \modelGPTFourOShort & 53.2846 +- 3.3591 & 0.1007 +- 0.0118 & 22.6667 +- 9.5393 & 0.0456 +- 0.0147 \\ %
\opennessSymbolOpenClosedWeight & \toolingSymbolStandard & \agentReAct & \modelGPTFiveMiniShort & 26.7262 +- 7.3963 & 0.0271 +- 0.0038 & \B 74.6667 +- 9.9095 & \B 0.0749 +- 0.0285 \\ %
\opennessSymbolOpenClosedWeight & \toolingSymbolStandard & \agentReAct & \modelGPTFiveShort & 79.8396 +- 3.5004 & 0.3729 +- 0.0343 & \B 82.6667 +- 8.6247 & \B 0.2763 +- 0.1139 \\ %
\opennessSymbolOpenClosedWeight & \toolingSymbolStandard & \agentReAct & \modelGeminiTwoPointFiveFlashShort & 52.8106 +- 7.9715 & 0.0630 +- 0.0256 & 36.0000 +- 10.9366 & 0.4355 +- 0.1396 \\ %
\opennessSymbolOpenOpenWeight & \toolingSymbolStandard & \agentReAct & \modelLlamaFourScoutShort & 24.8403 +- 5.0830 & 0.5882 +- 0.1436 & 41.3333 +- 11.2198 & 0.1696 +- 0.1036 \\ %
\opennessSymbolOpenClosedWeight & \toolingSymbolStandard & \agentReAct & \modelOThreeShort & 66.3740 +- 2.9673 & 0.2753 +- 0.0385 & 80.0000 +- 9.1138 & 0.3465 +- 0.0833 \\ %
\opennessSymbolOpenClosedWeight & \toolingSymbolEquivalent & \agentSmolagents & \modelClaudeThreeFiveHaikuShort & 49.9018 +- 4.3840 & 0.0422 +- 0.0040 & 25.3333 +- 9.9095 & 0.0556 +- 0.0101 \\ %
\opennessSymbolOpenClosedWeight & \toolingSymbolEquivalent & \agentSmolagents & \modelClaudeSonnetFourShort & 72.4064 +- 2.0645 & 0.7937 +- 0.0516 & 50.6667 +- 11.3913 & 0.6266 +- 0.0655 \\ %
\opennessSymbolOpenClosedWeight & \toolingSymbolEquivalent & \agentSmolagents & \modelGPTFourPointOneShort & 73.6775 +- 2.0892 & 0.0795 +- 0.0162 & \B 65.3333 +- 10.8434 & \B 0.0348 +- 0.0047 \\ %
\opennessSymbolOpenClosedWeight & \toolingSymbolEquivalent & \agentSmolagents & \modelGPTFourOShort & 46.2960 +- 4.0107 & 0.0784 +- 0.0083 & 14.6667 +- 8.0606 & 0.0498 +- 0.0100 \\ %
\opennessSymbolOpenClosedWeight & \toolingSymbolEquivalent & \agentSmolagents & \modelGPTFiveMiniShort & 57.3408 +- 5.2778 & 0.0204 +- 0.0015 & \B 50.6667 +- 11.3913 & \B 0.0154 +- 0.0047 \\ %
\opennessSymbolOpenClosedWeight & \toolingSymbolEquivalent & \agentSmolagents & \modelGPTFiveShort & 68.3871 +- 4.3921 & 0.1537 +- 0.0141 & 73.3333 +- 10.0757 & 0.1007 +- 0.0261 \\ %
\opennessSymbolOpenClosedWeight & \toolingSymbolEquivalent & \agentSmolagents & \modelGeminiTwoPointFiveFlashShort & 63.6577 +- 4.5823 & 0.0797 +- 0.0443 & 41.3333 +- 11.2198 & 0.0339 +- 0.0060 \\ %
\opennessSymbolOpenOpenWeight & \toolingSymbolEquivalent & \agentSmolagents & \modelLlamaFourScoutShort & 39.5772 +- 4.8440 & 0.0084 +- 0.0005 & \B 42.6667 +- 11.2691 & \B 0.0130 +- 0.0020 \\ %

\opennessSymbolOpenClosedWeight & \toolingSymbolCustom & \agentAsta & mixture & 87.7190 +- 1.4274 & 1.5286 +- 0.2911 & 70.6667 +- 10.3736 & 0.3061 +- 0.0931 \\ %

    \addlinespace[0.5em]
\opennessSymbolOpenClosedWeight & \toolingSymbolEquivalent & \agentScholarQA & \modelOThreeShort & \B 88.7411 +- 1.1760 & \B 2.9320 +- 0.4075 & {--} & {--} \\ %
\opennessSymbolOpenClosedWeight & \toolingSymbolEquivalent & \agentScholarQA & \modelClaudeSonnetFourShort & 87.9147 +- 1.2270 & 1.3137 +- 0.2812 & {--} & {--} \\ %
\opennessSymbolOpenClosedWeight & \toolingSymbolEquivalent & \agentScholarQANoTables & \modelClaudeSonnetFourShort & 86.2425 +- 1.4041 & 0.3931 +- 0.0303 & {--} & {--} \\ %

\opennessSymbolOpenClosedWeight & \toolingSymbolEquivalent & \agentScholarQANoTables & \modelGeminiTwoPointFiveFlashUnpinned & \B 87.6541 +- 1.4316 & \B 0.1263 +- 0.0095 & {--} & {--} \\ %

\opennessSymbolOpenClosedWeight & \toolingSymbolEquivalent & \agentScholarQANoTables & \modelGPTFourOMiniShort & \B 78.5016 +- 1.9356 & \B 0.0117 +- 0.0005 & {--} & {--} \\ %
    
\opennessSymbolOpenClosedWeight & \toolingSymbolEquivalent & \agentScholarQANoTables & \modelGPTFiveUnpinned & 85.9128 +- 1.5605 & 1.0993 +- 0.0736 & {--} & {--} \\ %
    
\opennessSymbolClosed & \toolingSymbolCustom & \agentElicit & {--} & 85.5343 +- 1.6108 & {--} & {--} & {--} \\ %
\opennessSymbolClosedWithApi & \toolingSymbolCustom & \agentPerplexitySQA & \modelGeminiTwoPointFiveFlashShort, \modelPerplexitySonarDeepResearch & 67.2758 +- 1.1920 & 0.4163 +- 0.0192 & 73.3333 +- 10.0757 & 0.2185 +- 0.0158 \\ %
\opennessSymbolClosedWithApi & \toolingSymbolCustom & \agentYouComResearch & {--} & 55.0019 +- 2.1562 & {--} & 8.0000 +- 6.1813 & {--} \\ %
    \opennessSymbolClosed & \toolingSymbolCustom & \agentSciSpace       & \modelClaudeSonnetFourShort & 84.6 +- 1.3 & {--} & {--} & {--} \\
\opennessSymbolOpenOpenWeight & \toolingSymbolEquivalent & \agentOpenScholar & \modelOpenScholar & \B 57.9581 +- 2.5860 & \B 0.0040 +- 0.0001 & {--} & {--} \\ %
\opennessSymbolClosedWithApi & \toolingSymbolCustom & \agentOpenAIDeepResearch & \modelDRAmbiguous, \modelGeminiTwoPointFiveProShort & 79.3958 +- 1.4246 & 1.8033 +- 0.0386 & {--} & {--} \\ %
\opennessSymbolClosedWithApi & \toolingSymbolCustom & \agentFutureHouseCrow & \modelGPTFourPointOneMiniShort, \modelOThreeMiniShort, \modelGeminiTwoPointFiveFlashShort & 81.0570 +- 1.7413 & 0.1065 +- 0.0035 & 72.0000 +- 10.2302 & 0.0646 +- 0.0031 \\ %
\opennessSymbolClosedWithApi & \toolingSymbolCustom & \agentFutureHouseFalcon & \modelGPTFourPointOneMiniShort, \modelGeminiTwoPointFiveFlashShort, \modelOThreeMiniShort & 77.5945 +- 1.3240 & 0.4031 +- 0.0506 & 74.6667 +- 9.9095 & 0.2195 +- 0.0112 \\ %
\opennessSymbolOpenClosedWeight & \toolingSymbolCustom & \agentSTORM & \modelGPTThreeFiveTurboShort, \modelGPTFourOShort & 78.3020 +- 2.3597 & 0.0941 +- 0.0022 & {--} & {--} \\ %

    \bottomrule
  \end{tabular}
  \label{apx:tab:results-lit-qa}
\end{table}

\begin{figure}[htbp]
  \centering
  \includegraphics[width=\textwidth]{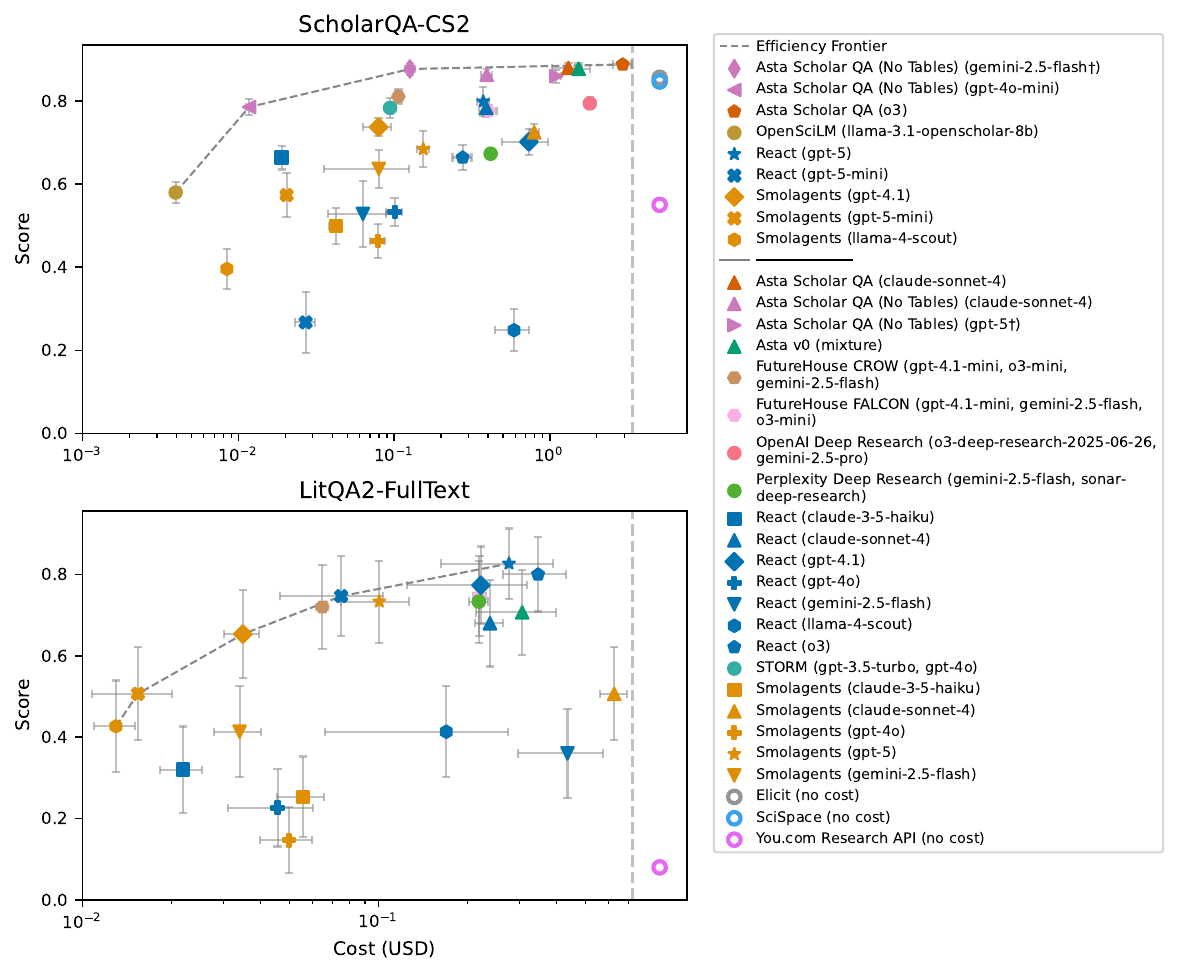}
  \caption{Score vs. cost analysis for \catlit{} QA benchmarks (\cref{apx:tab:results-lit-qa}). Points indicate means; error bars denote 95\% confidence intervals. Points on the Pareto frontier are connected with dotted lines, representing optimal quality-cost trade-offs for each eval (\evalsqa, \evallitqaft). Note: the x-axis (cost) uses a log scale.
    \captionDaggerNotice
    }
  \label{apx:fig:results-lit-qa}
\end{figure}

\begin{table}[tp]
  \centering
  \small
  
  \caption{\catlit \evaltables task benchmark results.
  }

  \rowcolors{5}{gray!10}{}
  \begin{tabular}{@{}l l l l A B@{}}
    \toprule
    O & T & Agent & Model & \multicolumn{2}{c}{\evaltables} \\
    \cmidrule(lr){5-6}
    & & & & {Score} & {Cost} \\
    \midrule
\opennessSymbolOpenClosedWeight & \toolingSymbolStandard & \agentReAct & \modelClaudeThreeFiveHaikuShort & 21.7385 +- 2.6388 & 0.0127 +- 0.0010 \\ %
\opennessSymbolOpenClosedWeight & \toolingSymbolStandard & \agentReAct & \modelClaudeSonnetFourShort & 25.5490 +- 3.1379 & 0.0685 +- 0.0054 \\ %
\opennessSymbolOpenClosedWeight & \toolingSymbolStandard & \agentReAct & \modelGPTFourPointOneShort & 27.4864 +- 3.1526 & 0.0383 +- 0.0037 \\ %
\opennessSymbolOpenClosedWeight & \toolingSymbolStandard & \agentReAct & \modelGPTFourOShort & 16.3218 +- 2.3998 & 0.0548 +- 0.0048 \\ %
\opennessSymbolOpenClosedWeight & \toolingSymbolStandard & \agentReAct & \modelGPTFiveMiniShort & \B 32.0946 +- 3.2919 & \B 0.0128 +- 0.0012 \\ %
\opennessSymbolOpenClosedWeight & \toolingSymbolStandard & \agentReAct & \modelGPTFiveShort & 29.4182 +- 3.7469 & 0.0640 +- 0.0051 \\ %
\opennessSymbolOpenClosedWeight & \toolingSymbolStandard & \agentReAct & \modelGeminiTwoPointFiveFlashShort & 25.1845 +- 3.1398 & 0.0219 +- 0.0021 \\ %
\opennessSymbolOpenOpenWeight & \toolingSymbolStandard & \agentReAct & \modelLlamaFourScoutShort & 9.4838 +- 2.3276 & 0.7604 +- 0.1015 \\ %
\opennessSymbolOpenClosedWeight & \toolingSymbolStandard & \agentReAct & \modelOThreeShort & 32.9295 +- 3.3195 & 0.0501 +- 0.0040 \\ %
\opennessSymbolOpenClosedWeight & \toolingSymbolEquivalent & \agentSmolagents & \modelClaudeThreeFiveHaikuShort & 15.0309 +- 2.8153 & 0.0168 +- 0.0025 \\ %
\opennessSymbolOpenClosedWeight & \toolingSymbolEquivalent & \agentSmolagents & \modelClaudeSonnetFourShort & 24.7938 +- 2.8750 & 0.2037 +- 0.0179 \\ %
\opennessSymbolOpenClosedWeight & \toolingSymbolEquivalent & \agentSmolagents & \modelGPTFourPointOneShort & 27.2224 +- 3.1022 & 0.0435 +- 0.0045 \\ %
\opennessSymbolOpenClosedWeight & \toolingSymbolEquivalent & \agentSmolagents & \modelGPTFourOShort & 14.6058 +- 2.4608 & 0.0511 +- 0.0072 \\ %
\opennessSymbolOpenClosedWeight & \toolingSymbolEquivalent & \agentSmolagents & \modelGPTFiveMiniShort & \B 30.0196 +- 3.2081 & \B 0.0088 +- 0.0006 \\ %
\opennessSymbolOpenClosedWeight & \toolingSymbolEquivalent & \agentSmolagents & \modelGPTFiveShort & 31.5124 +- 3.2385 & 0.0599 +- 0.0042 \\ %
\opennessSymbolOpenClosedWeight & \toolingSymbolEquivalent & \agentSmolagents & \modelGeminiTwoPointFiveFlashShort & 25.2091 +- 2.8444 & 0.0206 +- 0.0023 \\ %
\opennessSymbolOpenOpenWeight & \toolingSymbolEquivalent & \agentSmolagents & \modelLlamaFourScoutShort & 8.6871 +- 2.1944 & 0.0989 +- 0.0872 \\ %

\opennessSymbolOpenClosedWeight & \toolingSymbolCustom & \agentAsta & mixture & \B 42.9411 +- 3.7244 & \B 0.5167 +- 0.0561 \\ %

    \addlinespace[0.5em]
\opennessSymbolOpenClosedWeight & \toolingSymbolEquivalent & \agentAstaTableAgent & \modelGPTFourPointOneShort & 38.8056 +- 3.4606 & 0.3466 +- 0.0382 \\ %
\opennessSymbolOpenClosedWeight & \toolingSymbolEquivalent & \agentAstaTableAgent & \modelClaudeThreeFiveHaikuShort & 31.0558 +- 3.5885 & 0.1650 +- 0.0180 \\ %
\opennessSymbolOpenClosedWeight & \toolingSymbolEquivalent & \agentAstaTableAgent & \modelClaudeSonnetFourShort & 37.2138 +- 3.3313 & 0.6761 +- 0.0736 \\ %
\opennessSymbolOpenClosedWeight & \toolingSymbolEquivalent & \agentAstaTableAgent & \modelGeminiTwoPointFiveFlashShort & 34.3657 +- 3.2871 & 0.1328 +- 0.0147 \\ %
\opennessSymbolOpenClosedWeight & \toolingSymbolEquivalent & \agentAstaTableAgent & \modelOThreeShort & 41.6220 +- 3.4967 & 0.5166 +- 0.0557 \\ %
\opennessSymbolOpenClosedWeight & \toolingSymbolEquivalent & \agentAstaTableAgent & \modelGeminiTwoPointFiveProShort & 35.3817 +- 3.4744 & 0.9932 +- 0.1575 \\ %
\opennessSymbolOpenOpenWeight & \toolingSymbolEquivalent & \agentAstaTableAgent & \modelLlamaFourScoutShort & 26.3560 +- 3.3349 & 0.0253 +- 0.0032 \\ %
\opennessSymbolOpenClosedWeight & \toolingSymbolEquivalent & \agentAstaTableAgent & \modelGPTFiveUnpinned & 42.6085 +- 3.4968 & 1.2812 +- 0.1395 \\ %
\opennessSymbolOpenClosedWeight & \toolingSymbolEquivalent & \agentAstaTableAgent & \modelGPTFiveMiniUnpinned & \B 41.7115 +- 3.6991 & \B 0.1720 +- 0.0189 \\ %
    \bottomrule
  \end{tabular}
  \label{apx:tab:results-lit-table}
\end{table}

\begin{figure}[htbp]
  \centering
  \includegraphics[width=\textwidth]{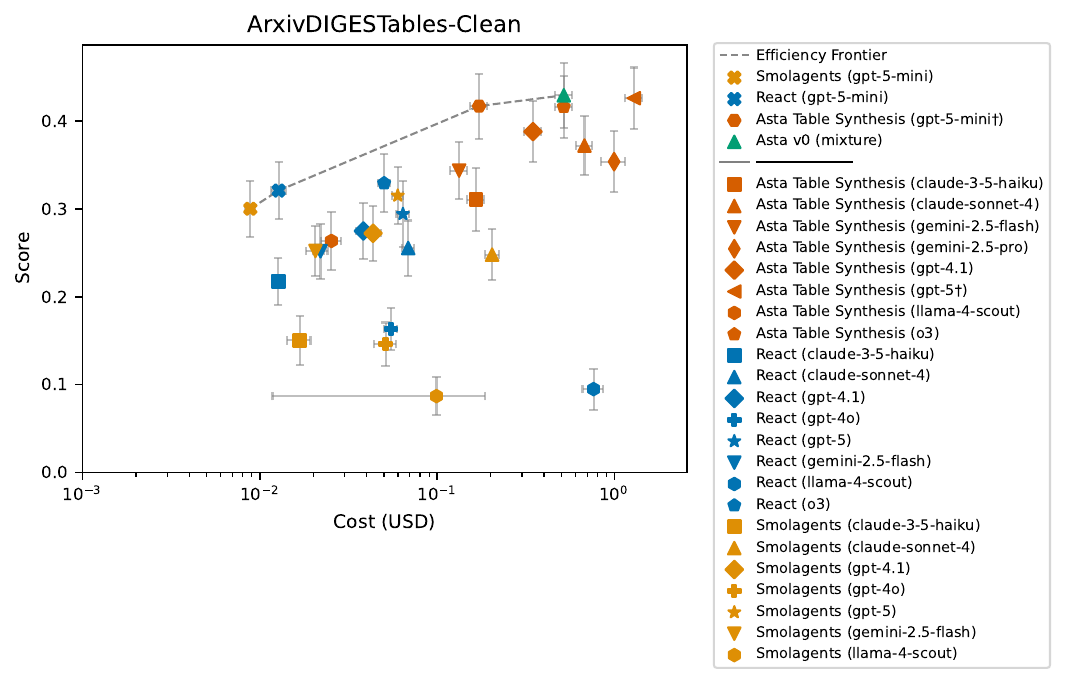}
  \caption{Score vs. cost analysis for the \catlit{} \evaltables benchmark (\cref{apx:tab:results-lit-table}). Points indicate means; error bars denote 95\% confidence intervals. Points on the Pareto frontier are connected with dotted lines, representing optimal quality-cost trade-offs for each eval. Note: the x-axis (cost) uses a log scale.
    \captionDaggerNotice
    }
  \label{apx:fig:results-lit-table}
\end{figure}

\begin{table}[tp]
  \setlength{\tabcolsep}{1pt}
  \centering
  \scriptsize
  \renewcommand{\uncertaintysize}{\scriptsize}

  \caption{\catcoding category results. 
  }

  \rowcolors{6}{gray!10}{}
  \begin{tabular}{@{}l l L{1.9cm} L{1.7cm} A B A B A U{table-format=2.3,round-mode=places,round-precision=3}{table-format=1.4{*},round-mode=places,round-precision=4}@{}}
    \toprule
    O & T & Agent & Model & \multicolumn{2}{c}{\evalsuper} & \multicolumn{2}{c}{\evalcorebench} & \multicolumn{2}{c}{\evaldatasci} \\
    \cmidrule(lr){5-6} \cmidrule(lr){7-8} \cmidrule(lr){9-10}
    & & & & {Score} & {Cost} & {Score} & {Cost} & {Score} & {Cost} \\
    \midrule
\opennessSymbolOpenClosedWeight & \toolingSymbolStandard & \agentReAct & \modelClaudeThreeFiveHaikuShort & 13.0741 +- 8.2927 & 0.0774 +- 0.0167 & 0.0000 & 0.0771 +- 0.0213 & 54.1111 +- 3.2574 & 0.0055 +- 0.0002 \\ %
\opennessSymbolOpenClosedWeight & \toolingSymbolStandard & \agentReAct & \modelClaudeSonnetFourShort & 22.5644 +- 11.1394 & 0.4479 +- 0.0871 & 40.5405 +- 16.0384 & 0.4986 +- 0.0810 & 75.5556 +- 2.8093 & 0.0439 +- 0.0020 \\ %
\opennessSymbolOpenClosedWeight & \toolingSymbolStandard & \agentReAct & \modelGPTFourPointOneShort & 11.2275 +- 7.5202 & 0.1564 +- 0.0694 & 18.9189 +- 12.7942 & 0.1186 +- 0.0350 & 67.0000 +- 3.0738 & 0.0075 +- 0.0003 \\ %
\opennessSymbolOpenClosedWeight & \toolingSymbolStandard & \agentReAct & \modelGPTFourOShort & 5.9259 +- 6.6604 & 0.3192 +- 0.0687 & 5.4054 +- 7.3867 & 0.1242 +- 0.0410 & 43.6667 +- 3.2422 & 0.0103 +- 0.0006 \\ %
\opennessSymbolOpenClosedWeight & \toolingSymbolStandard & \agentReAct & \modelGPTFiveMiniShort & \B 34.6429 +- 13.2074 & \B 0.1049 +- 0.0461 & \B 45.9459 +- 16.2796 & \B 0.0472 +- 0.0143 & \B 71.0000 +- 2.9662 & \B 0.0030 +- 0.0001 \\ %
\opennessSymbolOpenClosedWeight & \toolingSymbolStandard & \agentReAct & \modelGPTFiveShort & \B 41.1058 +- 12.8578 & \B 0.5890 +- 0.1400 & 45.9459 +- 16.2796 & 0.4427 +- 0.1388 & \B 78.0000 +- 2.7079 & \B 0.0210 +- 0.0009 \\ %
\opennessSymbolOpenClosedWeight & \toolingSymbolStandard & \agentReAct & \modelGeminiTwoPointFiveFlashShort & 20.0000 +- 10.6815 & 0.8751 +- 0.2953 & 2.7027 +- 5.2973 & 0.4695 +- 0.2140 & 55.4444 +- 3.2490 & 0.0186 +- 0.0032 \\ %
\opennessSymbolOpenOpenWeight & \toolingSymbolStandard & \agentReAct & \modelLlamaFourScoutShort & 4.7222 +- 5.2402 & 0.1745 +- 0.0655 & 0.0000 & 0.0271 +- 0.0181 & 9.6667 +- 1.9317 & 0.1100 +- 0.0077 \\ %
\opennessSymbolOpenClosedWeight & \toolingSymbolStandard & \agentReAct & \modelOThreeShort & 16.2910 +- 9.5855 & 0.3690 +- 0.0969 & \B 56.7568 +- 16.1835 & \B 0.1957 +- 0.0763 & \B 74.8889 +- 2.8348 & \B 0.0102 +- 0.0007 \\ %
\opennessSymbolOpenClosedWeight & \toolingSymbolEquivalent & \agentSmolagents & \modelClaudeThreeFiveHaikuShort & 16.7540 +- 9.6495 & 0.8120 +- 0.5808 & 0.0000 & 0.3323 +- 0.2095 & 9.8889 +- 1.9514 & 0.0237 +- 0.0103 \\ %
\opennessSymbolOpenClosedWeight & \toolingSymbolEquivalent & \agentSmolagents & \modelClaudeSonnetFourShort & 11.7011 +- 8.0336 & 3.5592 +- 1.7661 & 32.4324 +- 15.2920 & 2.1994 +- 0.7799 & 74.6667 +- 2.8431 & 0.1140 +- 0.0079 \\ %
\opennessSymbolOpenClosedWeight & \toolingSymbolEquivalent & \agentSmolagents & \modelGPTFourPointOneShort & 7.0370 +- 6.9249 & 0.1490 +- 0.1656 & 21.6216 +- 13.4477 & 0.0975 +- 0.0311 & 48.0000 +- 3.2659 & 0.0732 +- 0.0230 \\ %
\opennessSymbolOpenClosedWeight & \toolingSymbolEquivalent & \agentSmolagents & \modelGPTFourOShort & 3.9206 +- 4.9119 & 1.3509 +- 0.7154 & 5.4054 +- 7.3867 & 0.4187 +- 0.4095 & 16.7778 +- 2.4427 & 0.1372 +- 0.0642 \\ %
\opennessSymbolOpenClosedWeight & \toolingSymbolEquivalent & \agentSmolagents & \modelGPTFiveMiniShort & 14.1746 +- 8.8643 & 0.2403 +- 0.2072 & \B 5.4054 +- 7.3867 & \B 0.0143 +- 0.0037 & 65.2222 +- 3.1133 & 0.0162 +- 0.0046 \\ %
\opennessSymbolOpenClosedWeight & \toolingSymbolEquivalent & \agentSmolagents & \modelGPTFiveShort & 3.6111 +- 4.8391 & 0.0793 +- 0.0228 & 13.5135 +- 11.1677 & 0.1898 +- 0.1059 & 75.6667 +- 2.8050 & 0.0185 +- 0.0007 \\ %
\opennessSymbolOpenClosedWeight & \toolingSymbolEquivalent & \agentSmolagents & \modelGeminiTwoPointFiveFlashShort & 7.5000 +- 5.9505 & 0.7959 +- 0.9454 & 13.5135 +- 11.1677 & 0.8315 +- 0.7102 & 28.8889 +- 2.9629 & 0.0439 +- 0.0127 \\ %
\opennessSymbolOpenOpenWeight & \toolingSymbolEquivalent & \agentSmolagents & \modelLlamaFourScoutShort & 8.1481 +- 6.9933 & 0.3227 +- 0.3766 & 0.0000 & 0.0461 +- 0.0344 & 2.6667 +- 1.0532 & 0.0043 +- 0.0020 \\ %

\opennessSymbolOpenClosedWeight & \toolingSymbolCustom & \agentAsta & mixture & 19.3598 +- 10.3855 & 0.3317 +- 0.0570 & 48.6486 +- 16.3274 & 0.2260 +- 0.0933 & 74.7778 +- 2.8389 & 0.0110 +- 0.0007 \\ %

    \addlinespace[0.5em]
\opennessSymbolOpenClosedWeight & \toolingSymbolEquivalent & \agentAstaCode & \modelGPTFourPointOneShort & 16.2540 +- 9.4464 & 0.2850 +- 0.0591 & {--} & {--} & {--} & {--} \\ %
\opennessSymbolOpenClosedWeight & \toolingSymbolEquivalent & \agentAstaCode & \modelGPTFourOShort & 5.5556 +- 6.4000 & 0.4638 +- 0.1125 & {--} & {--} & {--} & {--} \\ %
\opennessSymbolOpenClosedWeight & \toolingSymbolEquivalent & \agentAstaCode & \modelGPTFiveShort & 13.5132 +- 9.3913 & 0.3715 +- 0.0717 & {--} & {--} & {--} & {--} \\ %
\opennessSymbolOpenClosedWeight & \toolingSymbolEquivalent & \agentAstaCode & \modelGPTFiveMiniShort & \B 12.8280 +- 9.0665 & \B 0.0674 +- 0.0139 & {--} & {--} & {--} & {--} \\ %
    \bottomrule
  \end{tabular}
  \label{apx:tab:results-coding}
\end{table}

\begin{figure}[htbp]
  \centering
  \includegraphics[width=\textwidth]{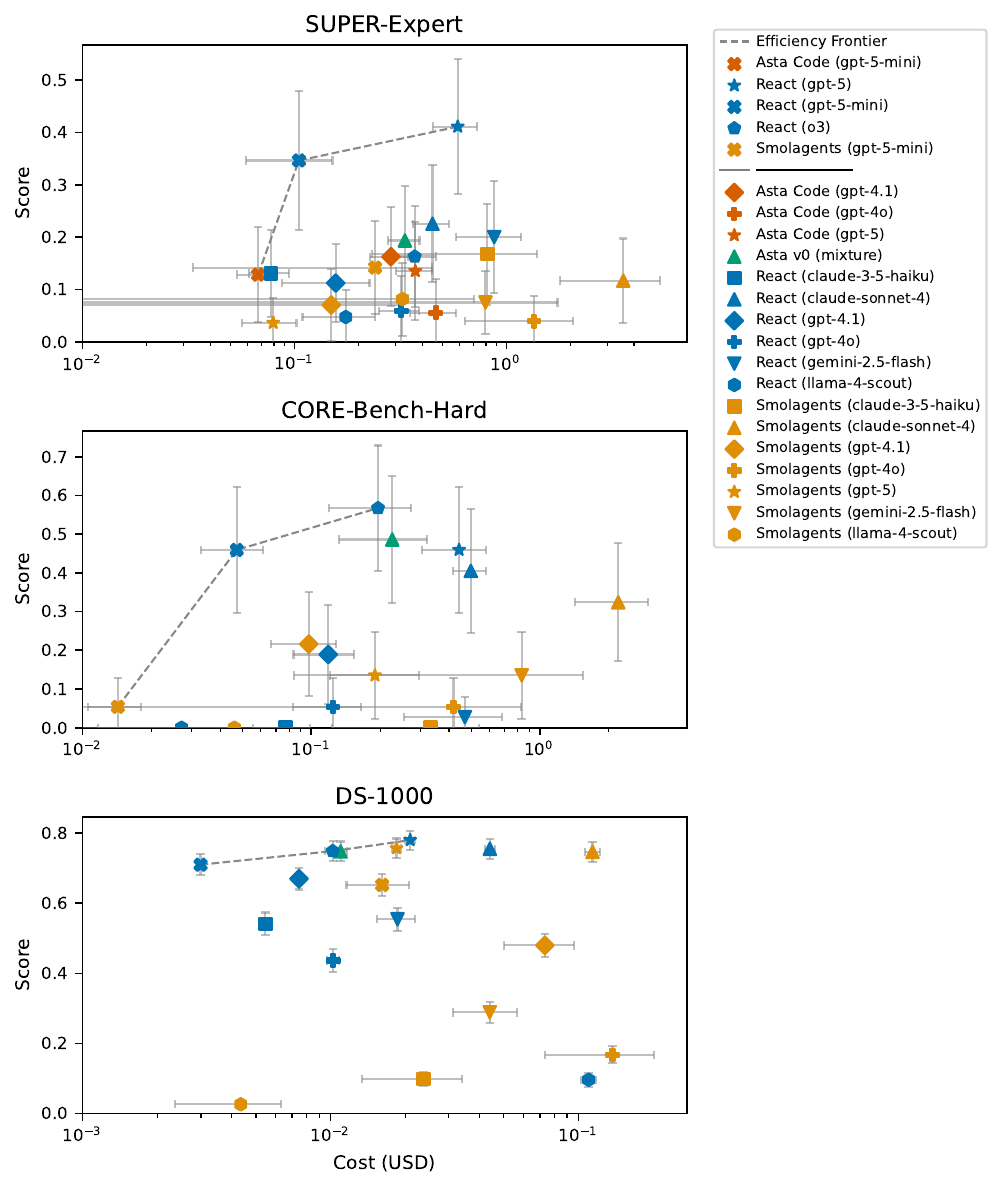}
  \caption{Score vs. cost analysis for \catcoding{} benchmarks (\cref{apx:tab:results-coding}). Points indicate means; error bars denote 95\% confidence intervals. Points on the Pareto frontier are connected with dotted lines, representing optimal quality-cost trade-offs for each eval (\evalcorebench, \evalsuper, \evaldatasci). Note: the x-axis (cost) uses a log scale.
    \captionDaggerNotice
    }
  \label{apx:fig:results-coding}
\end{figure}

\FloatBarrier   %

\begin{table}[!htp]
 \centering
 \small

  \caption{\catdata \evaldiscoverybench results.
  }

  \rowcolors{8}{gray!10}{}
    \begin{tabular}{@{}l l l L{4cm} A B@{}}
    \toprule
    O & T & Agent & Model & \multicolumn{2}{c}{\evaldiscoverybench} \\
    \cmidrule(lr){5-6}
    & & & & {Score} & {Cost} \\
    \midrule
\opennessSymbolOpenClosedWeight & \toolingSymbolStandard & \agentReAct & \modelClaudeThreeFiveHaikuShort & 24.3174 +- 4.6993 & 0.0116 +- 0.0010 \\ %
\opennessSymbolOpenClosedWeight & \toolingSymbolStandard & \agentReAct & \modelClaudeSonnetFourShort & 23.2221 +- 4.1479 & 0.1323 +- 0.0085 \\ %
\opennessSymbolOpenClosedWeight & \toolingSymbolStandard & \agentReAct & \modelGPTFourPointOneShort & \B 30.5362 +- 5.0581 & \B 0.0248 +- 0.0030 \\ %
\opennessSymbolOpenClosedWeight & \toolingSymbolStandard & \agentReAct & \modelGPTFourOShort & 13.2332 +- 3.7464 & 0.0397 +- 0.0095 \\ %
\opennessSymbolOpenClosedWeight & \toolingSymbolStandard & \agentReAct & \modelGPTFiveMiniShort & \B 26.9209 +- 4.7829 & \B 0.0111 +- 0.0012 \\ %
\opennessSymbolOpenClosedWeight & \toolingSymbolStandard & \agentReAct & \modelGPTFiveShort & 30.4808 +- 4.8361 & 0.0920 +- 0.0091 \\ %
\opennessSymbolOpenClosedWeight & \toolingSymbolStandard & \agentReAct & \modelGeminiTwoPointFiveFlashShort & 1.9277 +- 1.6859 & 0.1006 +- 0.0065 \\ %
\opennessSymbolOpenOpenWeight & \toolingSymbolStandard & \agentReAct & \modelLlamaFourScoutShort & 5.8621 +- 2.5846 & 0.1917 +- 0.0208 \\ %
\opennessSymbolOpenClosedWeight & \toolingSymbolStandard & \agentReAct & \modelOThreeShort & \B 33.6734 +- 5.0963 & \B 0.0392 +- 0.0043 \\ %
\opennessSymbolOpenClosedWeight & \toolingSymbolEquivalent & \agentSmolagents & \modelClaudeThreeFiveHaikuShort & 16.5142 +- 4.1274 & 0.0237 +- 0.0071 \\ %
\opennessSymbolOpenClosedWeight & \toolingSymbolEquivalent & \agentSmolagents & \modelClaudeSonnetFourShort & 28.7968 +- 4.7879 & 0.2369 +- 0.0190 \\ %
\opennessSymbolOpenClosedWeight & \toolingSymbolEquivalent & \agentSmolagents & \modelGPTFourPointOneShort & 28.3999 +- 4.8659 & 0.0451 +- 0.0183 \\ %
\opennessSymbolOpenClosedWeight & \toolingSymbolEquivalent & \agentSmolagents & \modelGPTFourOShort & 17.8158 +- 4.2452 & 0.0540 +- 0.0041 \\ %
\opennessSymbolOpenClosedWeight & \toolingSymbolEquivalent & \agentSmolagents & \modelGPTFiveMiniShort & 27.6804 +- 4.9305 & 0.0711 +- 0.0407 \\ %
\opennessSymbolOpenClosedWeight & \toolingSymbolEquivalent & \agentSmolagents & \modelGPTFiveShort & 26.7289 +- 4.7303 & 0.0770 +- 0.0057 \\ %
\opennessSymbolOpenClosedWeight & \toolingSymbolEquivalent & \agentSmolagents & \modelGeminiTwoPointFiveFlashShort & 24.7251 +- 4.7004 & 0.0172 +- 0.0068 \\ %
\opennessSymbolOpenOpenWeight & \toolingSymbolEquivalent & \agentSmolagents & \modelLlamaFourScoutShort & \B 20.1634 +- 4.5180 & \B 0.0078 +- 0.0018 \\ %

\opennessSymbolOpenClosedWeight & \toolingSymbolCustom & \agentAsta & mixture & 33.1725 +- 5.0587 & 0.2463 +- 0.0710 \\ %
    \addlinespace[0.5em]
\opennessSymbolOpenClosedWeight & \toolingSymbolEquivalent & \agentDataVoyager & \modelGPTFourPointOneUnpinned, \modelGPTFourOUnpinned & 29.8843 +- 5.0479 & 0.1472 +- 0.0199 \\ %
\opennessSymbolOpenClosedWeight & \toolingSymbolEquivalent & \agentDataVoyager & \modelClaudeSonnetFourShort, \modelGPTFourOUnpinned & 25.7241 +- 4.5812 & 0.5234 +- 0.0502 \\ %
\opennessSymbolOpenClosedWeight & \toolingSymbolEquivalent & \agentDataVoyager & \modelOThreeUnpinned, \modelGPTFourOUnpinned & 31.1268 +- 5.0476 & 0.2335 +- 0.0614 \\ %
\opennessSymbolOpenClosedWeight & \toolingSymbolEquivalent & \agentDataVoyager & \modelGPTFiveUnpinnedMinimal, \modelGPTFourOUnpinned & 27.0413 +- 4.6706 & 0.2152 +- 0.0290 \\ %
\opennessSymbolOpenClosedWeight & \toolingSymbolEquivalent & \agentDataVoyager & \modelGPTFiveUnpinned, \modelGPTFourOUnpinned & 29.5626 +- 4.8773 & 0.3541 +- 0.0753 \\ %
    \bottomrule
  \end{tabular}
  \label{apx:tab:results-data}
\end{table}

\begin{figure}[htbp]
  \centering
  \includegraphics[width=\textwidth]{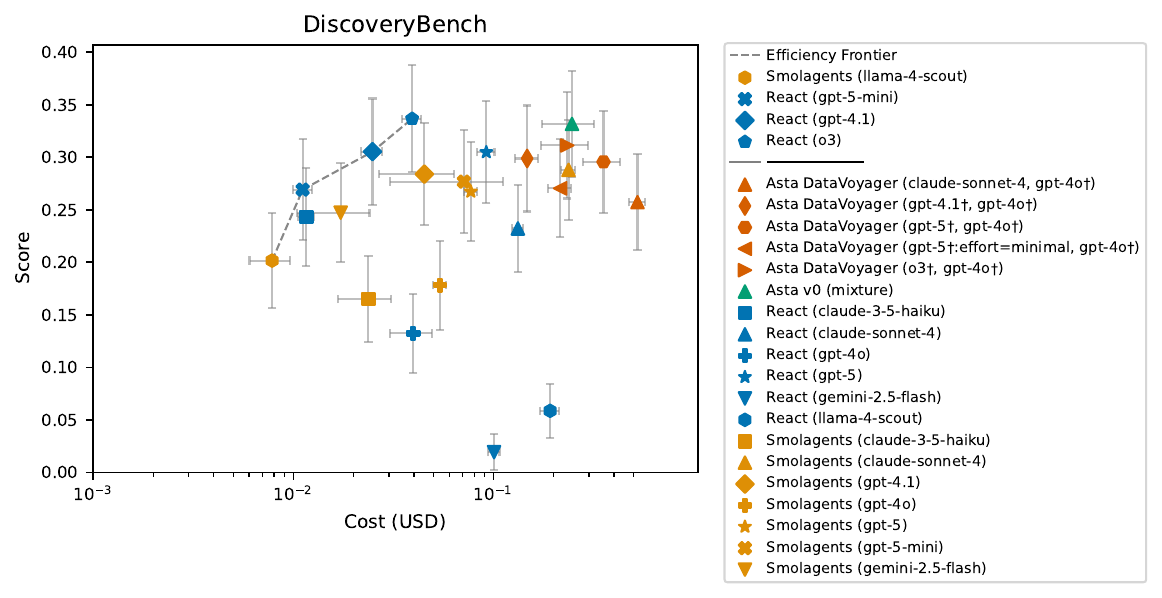}
  \caption{Score vs. cost analysis for \catdata{} sub-benchmarks. Points indicate means; error bars denote 95\% confidence intervals. Points on the Pareto frontier are denoted with red triangle markers, representing optimal quality-cost trade-offs for each eval (\evaldiscoverybench).
    \captionDaggerNotice
    }
  \label{fig:results-data}
\end{figure}

\begin{table}[tp]
  \setlength{\tabcolsep}{3pt}
  \centering
  \small

  \caption{\catendtoend category results. 
  }

  \rowcolors{5}{}{gray!10}
    \begin{tabular}{@{}l l L{2.5cm} L{2.0cm} A U{table-format=2.3,round-mode=places,round-precision=3}{table-format=1.3{*},round-mode=places,round-precision=3} A U{table-format=2.3,round-mode=places,round-precision=3}{table-format=1.3{*},round-mode=places,round-precision=3}@{}}
    \toprule
    O & T & Agent & Model & \multicolumn{2}{c}{\evalendtoend} & \multicolumn{2}{c}{\evalendtoendhard} \\
    \cmidrule(lr){5-6} \cmidrule(lr){7-8}
    & & & & {Score} & {Cost} & {Score} & {Cost} \\
    \midrule
\opennessSymbolOpenClosedWeight & \toolingSymbolStandard & \agentReAct & \modelClaudeThreeFiveHaikuShort & 4.4544 +- 2.7512 & 0.0419 +- 0.0113 & 4.8328 +- 3.3611 & 0.0478 +- 0.0110 \\ %
\opennessSymbolOpenClosedWeight & \toolingSymbolStandard & \agentReAct & \modelClaudeSonnetFourShort & 52.5225 +- 6.7948 & 0.7493 +- 0.0720 & 38.8998 +- 6.8844 & 0.8355 +- 0.0574 \\ %
\opennessSymbolOpenClosedWeight & \toolingSymbolStandard & \agentReAct & \modelGPTFourPointOneShort & 19.3493 +- 7.3425 & 0.1319 +- 0.0237 & 14.8309 +- 6.7740 & 0.1393 +- 0.0335 \\ %
\opennessSymbolOpenClosedWeight & \toolingSymbolStandard & \agentReAct & \modelGPTFourOShort & 1.6162 +- 1.6948 & 0.1566 +- 0.0353 & 1.4179 +- 1.9463 & 0.1353 +- 0.0278 \\ %
\opennessSymbolOpenClosedWeight & \toolingSymbolStandard & \agentReAct & \modelGPTFiveMiniShort & 9.4823 +- 7.5827 & 0.0297 +- 0.0058 & 15.7045 +- 8.2664 & 0.0396 +- 0.0080 \\ %
\opennessSymbolOpenClosedWeight & \toolingSymbolStandard & \agentReAct & \modelGPTFiveShort & 30.0039 +- 11.9339 & 0.4034 +- 0.0526 & \B 42.1035 +- 11.4110 & \B 0.5842 +- 0.0721 \\ %
\opennessSymbolOpenClosedWeight & \toolingSymbolStandard & \agentReAct & \modelGeminiTwoPointFiveFlashShort & 0.0000 & 2.4006 +- 1.1489 & 1.0714 +- 2.1000 & 1.2633 +- 0.6720 \\ %
\opennessSymbolOpenOpenWeight & \toolingSymbolStandard & \agentReAct & \modelLlamaFourScoutShort & 1.8819 +- 2.1243 & 0.8177 +- 0.1345 & 0.8590 +- 1.0805 & 0.8128 +- 0.1440 \\ %
\opennessSymbolOpenClosedWeight & \toolingSymbolStandard & \agentReAct & \modelOThreeShort & 34.9483 +- 10.1077 & 0.0652 +- 0.0103 & 21.0421 +- 7.6100 & 0.0751 +- 0.0189 \\ %
\opennessSymbolOpenClosedWeight & \toolingSymbolEquivalent & \agentSmolagents & \modelClaudeThreeFiveHaikuShort & 5.2560 +- 3.1325 & 0.9457 +- 0.5601 & 3.7360 +- 2.3527 & 0.5049 +- 0.5384 \\ %
\opennessSymbolOpenClosedWeight & \toolingSymbolEquivalent & \agentSmolagents & \modelClaudeSonnetFourShort & 47.1630 +- 6.0675 & 0.8729 +- 0.1097 & 35.8012 +- 7.7854 & 1.5115 +- 0.3067 \\ %
\opennessSymbolOpenClosedWeight & \toolingSymbolEquivalent & \agentSmolagents & \modelGPTFourPointOneShort & 36.6486 +- 9.3248 & 0.1777 +- 0.1461 & 30.0272 +- 7.6760 & 1.9548 +- 1.7729 \\ %
\opennessSymbolOpenClosedWeight & \toolingSymbolEquivalent & \agentSmolagents & \modelGPTFourOShort & 5.3869 +- 3.9407 & 0.4730 +- 0.3468 & 5.1368 +- 3.3310 & 0.8656 +- 0.7566 \\ %
\opennessSymbolOpenClosedWeight & \toolingSymbolEquivalent & \agentSmolagents & \modelGPTFiveMiniShort & 22.3234 +- 9.6360 & 0.0757 +- 0.1136 & 21.5856 +- 7.4524 & 0.0761 +- 0.1079 \\ %
\opennessSymbolOpenClosedWeight & \toolingSymbolEquivalent & \agentSmolagents & \modelGPTFiveShort & \B 62.7545 +- 9.8341 & \B 0.2047 +- 0.0246 & 30.3081 +- 10.5246 & 0.2317 +- 0.0430 \\ %
\opennessSymbolOpenClosedWeight & \toolingSymbolEquivalent & \agentSmolagents & \modelGeminiTwoPointFiveFlashShort & 34.0400 +- 10.1766 & 1.8772 +- 0.8301 & 23.1940 +- 7.7716 & 2.5414 +- 1.2032 \\ %
\opennessSymbolOpenOpenWeight & \toolingSymbolEquivalent & \agentSmolagents & \modelLlamaFourScoutShort & 0.1667 +- 0.3267 & 0.2834 +- 0.1524 & 0.7396 +- 0.7124 & 0.2510 +- 0.1809 \\ %

\opennessSymbolOpenClosedWeight & \toolingSymbolCustom & \agentAsta & mixture & \B 70.4125 +- 6.2983 & \B 10.6427 +- 0.7170 & \B 67.2842 +- 5.2905 & \B 14.4873 +- 1.0501 \\ %

    \addlinespace[0.5em]
\opennessSymbolOpenClosedWeight & \toolingSymbolStandard & \agentFaker & \modelGPTFourPointOneUnpinned & \B 39.1740 +- 6.9341 & \B 0.0263 +- 0.0013 & \B 25.3660 +- 4.5482 & \B 0.0288 +- 0.0011 \\ %
\opennessSymbolOpenClosedWeight & \toolingSymbolCustom & \agentAutoAsta & \modelGPTFourPointOneUnpinned & 36.6236 +- 7.7198 & 7.6103 +- 1.6495 & 39.2917 +- 7.0397 & 9.3188 +- 1.2431 \\ %
\opennessSymbolOpenClosedWeight & \toolingSymbolCustom & \agentAutoAsta & \modelClaudeSonnetFourShort & \B 70.4709 +- 6.2304 & \B 10.6427 +- 0.7170 & \B 68.2114 +- 4.3677 & \B 14.4873 +- 1.0501 \\ %
\opennessSymbolOpenClosedWeight & \toolingSymbolCustom & \agentCodeScientist & \modelClaudeSonnetThreeSevenShort & \B 65.3458 +- 7.1086 & \B 2.7598 +- 0.5100 & \B 64.4985 +- 5.5100 & \B 3.5487 +- 0.6921 \\ %
    \bottomrule
  \end{tabular}
  \label{apx:tab:results-endtoend}
\end{table}

\begin{figure}[htbp]
  \centering
  \includegraphics[width=\textwidth]{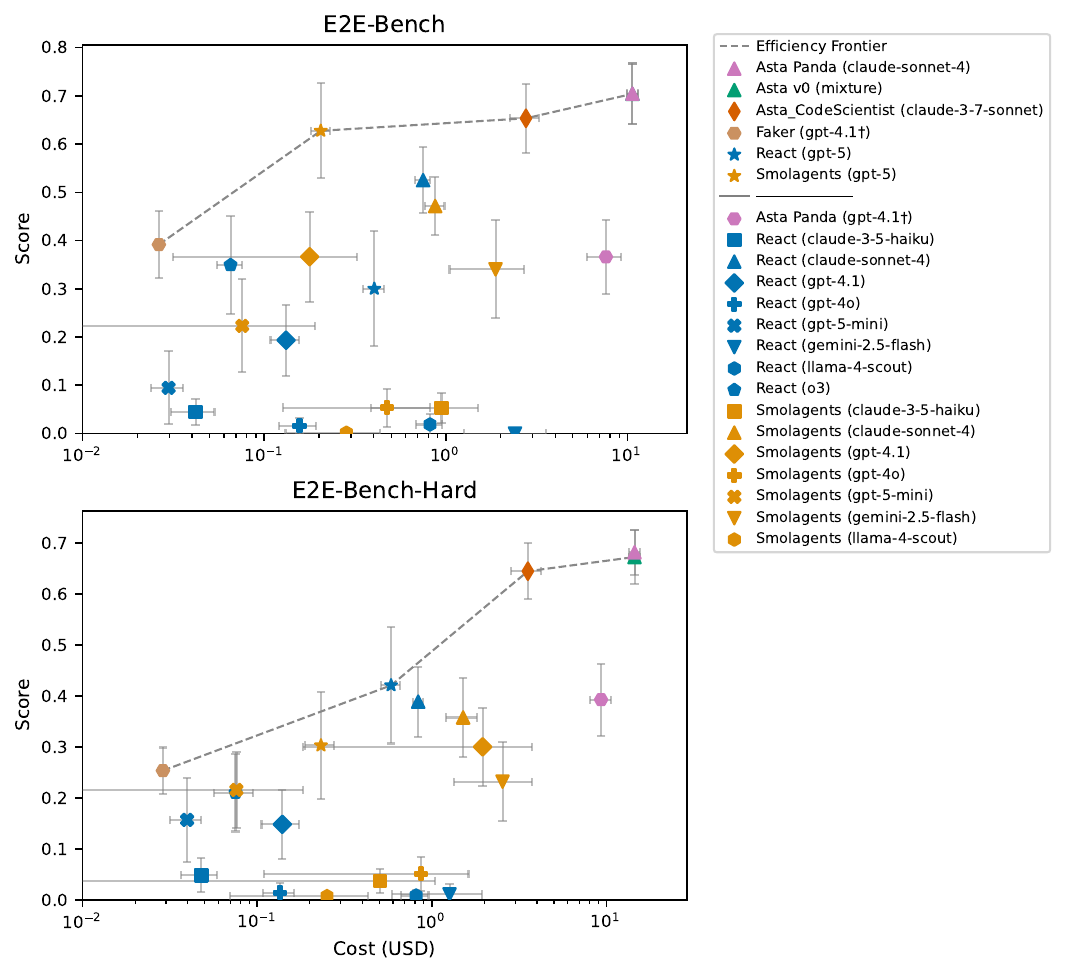}
  \caption{Score vs. cost analysis for \catendtoend{} benchmarks (\cref{apx:tab:results-endtoend}). Points indicate means; error bars denote 95\% confidence intervals. Points on the Pareto frontier are connected with dotted lines, representing optimal quality-cost trade-offs for each eval (\evalendtoend, \evalendtoendhard). Note: the x-axis (cost) uses a log scale.
    \captionDaggerNotice
    }
  \label{apx:fig:results-endtoend}
\end{figure}

\section{Evaluations}
\label{sec:appendix-evals}

\subsection{Short descriptions}

\paragraph{\evalpaperfinder}

\evalpaperfinder tests an agent's ability to handle challenging scientific search queries.  Given a textual query string, the task is to return a ranked list of papers that satisfy the query.  This new benchmark is a subset of our own internal evaluation for our literature-search agent (\agentPaperFinder).  Unlike existing paper-finding benchmarks, which are restricted to semantic search queries, our dataset includes metadata and navigational queries along with a diverse mix of semantic queries.  The queries are sourced from PaperFinder\footnote{\url{https://paperfinder.allen.ai/chat}} and OpenSciLM\footnote{\url{https://openscilm.allen.ai/}} user logs and the LitSearch \citep{ajith2024litsearchretrievalbenchmarkscientific} and PaSa \citep{he-etal-2025-pasa} datasets. Evaluating retrieval tasks is challenging, and our chosen evaluation metrics along with other benchmark details are discussed in \cref{sec:eval-paperfinder}. Briefly, navigational and metadata queries are evaluated in terms of F1 over the result set, and semantic queries use the harmonic mean of \emph{estimated} recall and nDCG.  The final evaluation metric is an average of per-query scores.

\paragraph{\evallitqaft/\evallitqasearchft}
These two benchmarks measure an agent's ability to answer questions and retrieve papers within the biomedical domain.  They are based on the LitQA2 dataset \citep{Skarlinski2024LanguageAgents}, which contains 199 multiple-choice questions, each associated with a target paper whose full-text can potentially answer the question. To enable fair comparison for agents using our standard retrieval tools, we filter the original dataset to a subset of 85 questions where the associated relevant paper is available in our \toollitapi snippet search index within the specified cutoff date (see \cref{tab:evals}). Following  \citet{Skarlinski2024LanguageAgents}, \evallitqaft evaluates in terms of {\em accuracy}, the fraction of questions with a correct answer.  \evallitqasearchft isolates the retrieval task aimed at finding $K$ papers such that one of them is the target paper for the question, and evaluates on recall@30 (as used in \citet{Skarlinski2024LanguageAgents}).  To avoid double-counting this benchmark when computing aggregate macro-averaged \catlit scores (compared to other benchmarks in that category), we weight each of these two evals by 0.5 in the macro-average.  For additional details and comparisons, see \cref{sec:litqa2-search-results}.

\paragraph{\evalsqa}
The \evalsqa benchmark tests an agent's ability to answer long-form scientific questions.  Given a complex scientific question like ``How is diversity typically evaluated in recommendation systems?'' the task is to identify relevant prior work and compose a long-form answer report that appropriately cites sources.  \evalsqa is a new benchmark that builds upon the recent ScholarQA-CS \citep{Asai2024OpenScholarSS} by incorporating real scientific queries and introducing four facets for coverage and precision evaluation of both answers and their attributions, using LLM-as-judge.  The average of these four facet scores is the final evaluation metric.  For more detail, see \cref{sec:eval-sqa}.

\paragraph{\evaltables}
The \evaltables benchmark tests an agent's ability to create a literature review table---one whose rows are publications and whose columns consist of aspects used to compare and contrast a set of papers. Given a set of related papers and a caption describing the table's intent (e.g., ``Overview of LLM pretraining benchmarks''), the task is to automatically output a complete literature review table.  We release a new benchmark that builds on \texttt{ArxivDIGESTables}, the first high-quality dataset for literature review table generation created by \citet{newman-etal-2024-arxivdigestables} by extracting review tables from ArXiv papers. Our evaluation includes two key improvements: (i) we curate a small clean subset of instances from the original test set, and (ii) we introduce an end-to-end evaluation methodology for the task. Tables are scored by prompting an LLM to ``unroll'' them into statements.  The evaluation metric is the proportion of ground truth statements from the reference table that are entailed (according to an LLM judge) by the unrolled generated table. For more detail, see \cref{sec:eval-tables}.

\paragraph{\evalsuper}
The \evalsuper benchmark \citep{bogin-etal-2024-super-emnlp} (\underline{S}etting \underline{UP} and \underline{E}xecuting tasks from \underline{R}esearch repositories) tests the ability of code agents to set up and execute Python machine learning experiments reported in ML and NLP papers. It targets the common yet often non-trivial and time-consuming task of setting up and running code from sparsely documented repositories accompanying published papers.
Given a natural language instruction along with a GitHub repository pointer (e.g., asking to train a model following a paper's code at a given URL), the task is to clone the repository, install any needed dependencies, configure, run the requested training/evaluation, and report the outcome (e.g., model accuracy).
In contrast to other repository-centered code execution tasks, the particular focus here is on \emph{low-resource} research repositories on GitHub---like those researchers often encounter when validating and expanding upon prior published work. 
For more detail, see \cref{sec:eval-super}.

\paragraph{\evalcorebench}
The \evalcorebench benchmark~\citep{siegel-etal-2025-corebench} tests an agent's ability to reproduce experiments and analyses from papers.  The input is a "capsule" from CodeOcean.com containing code and data released alongside a published paper, as well as a set of instructions indicating specific analyses to perform with the capsule (full example in \cref{sec:corebench-example}). The task is to perform these analyses and write answers in a \texttt{report.json} file. The capsules in {\evalcorebench} are chosen to be highly reproducible and span a variety of domains, including computer science, social science, and medicine, and use Python and R programming languages. For more detail, see \cref{sec:eval-corebench}.

\paragraph{\evaldatasci}
The \evaldatasci benchmark~\citep{lai2023ds} tests the ability of code models on routine data science tasks encountered in everyday research. The input is a coding question and an incomplete code snippet that the agent must fill in to answer the question (see example in \cref{sec:datasci-example}). The output code snippet is graded by running it against a (problem-specific) test case. This benchmark contains 1000 problems involving 7 Python libraries that were originally collected from StackOverflow and perturbed to avoid training leakage. We use the task implementation provided in Inspect evals \citep{inspect_evals_2025} and report the accuracy of the proposed code passing the target test cases. For more detail, see \cref{sec:eval-datasci}.

\paragraph{\evaldiscoverybench}
The \evaldiscoverybench \citep{majumder-etal-2025-discoverybench} benchmark aims to test whether the agent can automatically find and verify hypotheses from given dataset(s), performing data-driven analysis. The input to the task is a discovery goal and a collection of datasets and their respective metadata, and the output is a hypothesis addressing the goal with the highest specificity for the context, variables, and relationship supported by the dataset(s). Optionally, a workflow for deriving a hypothesis can be output to augment information already present in the hypothesis. This is the \emph{first} comprehensive benchmark to test agents' or language models' ability to perform data analysis---including data preparation, basic statistical analysis, complex data transformation, and modeling---on datasets from 6 diverse domains, such as sociology and engineering. We collect task datasets from open public repositories made available by already published works from the 6 domains. The discovery goals are extracted from the associated papers to the datasets, or human-annotated, where each gold output (i.e., the hypothesis) is rigorously verified by data analysis experts. The performance on the benchmark is measured as the alignment of the predicted and gold hypotheses. The final metric, Hypothesis Matching Score, is a product of three LLM-as-judge scores that measure the alignment of the predicted and the gold hypotheses in the dimensions of their context, associated variables, and the relationship among them. For more detail, see \cref{sec:eval-discoverybench}.

\paragraph{\evalendtoend} The \evalendtoend task aims to test whether agents can perform the full research pipeline of ideation, planning, (software) experiment design, implementation, execution, analysis, and producing a final report, i.e., a complete research cycle. The input to the task is a research question in the domain of AI/NLP and a detailed description of the steps to investigate it, and the output is a technical report, a trace of the agent's reasoning, and any code or artifacts (e.g., datasets) generated. This is a new release and forms the first agent-neutral benchmark (i.e., a benchmark that isn't designed to highlight the strengths and scope of a particular agent) designed to compare automatic scientific discovery (ASD) agents.  It fills a gap in the current research landscape where there are many such agents, e.g., AI Scientist \citep{Lu2024TheAS}, AgentLab \citep{agentlab}, and CodeScientist \citep{codescientist}, but no systematic way to compare them. In practice, to allow
more controlled
system-to-system comparisons, the problems are specified in considerable detail and hence only weakly test the ideation and planning steps. At the same time, these problems are not as prescriptive as typical ML coding problems, e.g., in MLAgentBench \citep{mlagentbench}. The problems are created via a mixture of machine generation and human review, and include a detailed task description and a problem-specific evaluation rubric.  The final score is an overall LLM-as-judge assessment based on three LLM-as-judge scores obtained by evaluating each relevant agent output (report, code, and artifacts) against the rubric. For more detail, see \cref{sec:eval-endtoend}.

\paragraph{\evalendtoendhard}
This task is similar to \evalendtoend{}, except the problems are generally harder. It follows the same task definition, evaluation, baselines, and environment as \evalendtoend{}, however the data collection method is different. For more detail, see \cref{sec:eval-endtoendhard}.

\subsection[\evalpaperfinderPlain]{\evalpaperfinderNoSmall}
\label{sec:eval-paperfinder}

In the rise of LLM-based agentic workflows, the ability to answer \textbf{challenging} scientific search queries, across a wide range of searching criteria, have become possible. However, current paper finding benchmarks largely confine themselves to a small subset of search query kinds (e.g. LitSearch \citep{ajith2024litsearchretrievalbenchmarkscientific}, PaSa \citep{he-etal-2025-pasa} and LitQA2 dataset \citep{Skarlinski2024LanguageAgents}). They focus on purely semantic criteria, not covering metadata or navigational queries, and they are missing a methodological process to cover the different within-semantic challenging types. 

\evalpaperfinder is a subset of our own internal evaluation for our literature-search agent (\agentPaperFinder), which focuses on challenging queries (the internal evaluation also mixes in a bunch of easier queries, to ensure stability as a product and avoid regressions). \evalpaperfinder is designed to be \emph{challenging} (including things that our system currently does not perform well on) and \emph{realistic} (based to the extent possible on real-world queries and information needs). It also aims to be \emph{broad and diverse} in two axes: first, it covers a broader set of information needs. Unlike existing datasets that focus on semantic queries that search for a set of unknown-to-the-user papers based on description of their content, our benchmark includes also “navigational” queries that seek a single known-to-the-user paper based on a short reference (“the alpha-geometry paper”), and queries that define paper sets based on a wide set of metadata criteria (“acl 2024 papers that cite the transformers paper”). The second axis of diversity is within the semantic-search category, in which we seek to include different types of query challenges. The dataset mixes the different categories, and doesn’t clearly indicate which query belongs to which category (even though a human will very easily tell). This is following our belief that a literature-search agent should be able to handle all these query types, even if by merely routing them to different sub-agents. 

PaperFindingBench includes 48 navigational queries, 43 metadata queries, and 242 semantic queries. Some of the metadata queries contain (easy) navigational queries as part of their criteria, but there is currently a strict separation between metadata and semantic queries (metadata queries do not involve a semantic component and vice-versa), which may change in future versions.

\paragraph{Dataset Creation} 

\emph{The Navigational queries} are based on PaperFinder\footnote{\url{https://paperfinder.allen.ai/chat}} usage logs, to include queries that, at least at some point in time, paper-finder failed on.

\emph{The semantic queries} are curated from a mix of sources: PaperFinder usage logs, OpenSciLM\footnote{\url{https://openscilm.allen.ai/}} usage logs, and existing literature-search datasets: LitSearch \citep{ajith2024litsearchretrievalbenchmarkscientific} and PaSa \citep{he-etal-2025-pasa}.  We first identified a subset of queries that were challenging for the PaperFinder system, by looking for queries that returned few or no results identified by the system as “perfectly relevant”, and for which we assessed (for query logs) or know (for the annotated dataset) that relevant papers exist. We then manually inspected a collection of such queries to identify challenge types.\footnote{These include, for example, multiple criteria, complex relations between criteria, use of uncommon terms, use of incorrect jargon, seeking details that are not part of the main claim of the paper, query providing unnecessary or even distracting background information.}  Finally, we created a set in which all challenge types are represented, while prioritizing queries for which running PaperFinder in an ablation mode with any of its components resulted in fewer perfectly-relevant papers for the ones that we do find. The set contains a mix of queries for which we assume there are many relevant results,  and queries for which we assume only a handful of results exist. For numerous queries, assessing the relevance of the paper cannot be done solely based on title and abstract, but requires evidence from the paper’s full text.

\emph{Metadata queries} These were hand-crafted to achieve broad coverage of semantic-scholar API usage, as well as interaction between APIs, as well as challenges that are solvable but not directly supported by the APIs, such as negation (“not citing the transformers paper”). 
The queries include nesting and recursion of properties, and are inspired by the most complex queries we saw in the dataset, and taken up a notch or two. We emphasized queries that require combining multiple APIs.

\paragraph{Evaluation}
Evaluating retrieval is challenging, as it ideally requires a gold-set of all relevant documents in the corpus, which is often not known. Such a gold-set \emph{is} available for the navigational and the metadata queries (each metadata query is internally associated with python code that uses the APIs to solve it completely, and whose results we use as the gold set). For the semantic queries, the full-coverage gold-set does not exist, and we resort to a combination of partial annotation and LLM-based judgement. Each query is associated with a (potentially empty) small-set of known-to-be-good matches, as well as with a weighted set of relevance criteria that should be individually verified by the LLM against evidence from the paper for the paper to be considered a good match. The individual relevance criteria were automatically generated by an LLM based on a (potentially expanded version of) the original query. For a fifth of the queries, the relevance criteria were manually verified and corrected or tweaked. As the tweaks and corrections turned out to be mostly minimal, and as the LLM-based relevance criteria were proved to be highly effective for the queries for which manual annotation for some papers is available, we consider all the relevance criteria as reliable, though they may be further improved in future versions. As we aim to assess retrieval and not the judging-LLM’s ability to handle long-contexts, we don’t provide the paper’s full-text for relevance judgement but rather require each result item to be associated with extracted evidence text (either from the paper itself or from papers citing it), which is then fed to the LLM for relevance judgement.

\paragraph{Scoring Metrics} We use two different scoring metrics.

\noindent\emph{For the navigational and metadata queries}, for which the gold-set is known, we use F1 over the result-set to score individual queries. 

\noindent\emph{For the semantic queries}, which are based on LLM judgement, we can compute precision, but not recall. One potential metric would be simply the number of returned documents that are LLM-judged to be relevant, however, this number is unbounded and harder to integrate with other scores in AstaBench. We thus opted to compute recall over an \emph{estimated} set size for each query (that is, we divide by an estimated set size and not a definitive one), to bound the numbers between 0 and 1. The estimated set size is determined by running multiple variations of PaperFinder with very lenient threshold, taking the union of the resulting set, and then multiplying it by a factor that ranges from 2 to 10 to estimate an upper bound and allow room for additional papers (smaller initial sets are less reliable and are multiplied by a larger number). Note that in extreme cases, this may result in a recall number larger than 1. We bound this by considering the retrieval-adjusted metric of $recall@k$ where we set $k$ to be the estimated set size (this corresponds to the established $recall@R$ metric, but we compute $estimated-recall@estimated$). Computing recall@k fulfills two purposes: it bounds the score in 1, and also discourages submission of “junk” results.

We balance recall@k not by precision, but by nDCG, as it provides a more relevant signal (favoring ranking relevant documents over irrelevant ones). The combination of nDCG and recall@estimated makes precision mostly redundant. To provide a single score for each individual query, we combine the recall and nDCG numbers using an harmonic mean (F1 over estimated-recall and nDCG).

To provide a single unified score for the entire dataset, we average the individual query scores, overall queries regardless of their type.

\paragraph{Tools Cutoff Date}
We encourage participants to use the keyword and snippet search functionalities provided in \toollitapi. In any case we expect submissions to follow the same cutoff date as the corpus cutoff date for both these tools which is set to June 1$^{st}$ 2025.

\paragraph{Example Input}
An example input can be found in \cref{sec:paperfinder-example}.

\subsection[\evalsqaPlain]{\evalsqaNoSmall}
\label{sec:eval-sqa}
Scientific literature reviews are a longstanding component of scientific workflows, and today are increasingly automated by commercial and open long-form QA services, such as OpenAI Deep Research, ScholarQA \citep{Singh2025Ai2SQ}, Elicit, Perplexity, Paper QA \citep{Skarlinski2024LanguageAgents}, and many others. Evaluating long-form answers to literature review questions is a challenging problem in natural language processing.  Many acceptable long-form answers exist for any given question, and even with a dataset of ``gold'' answers, it is difficult to define how to score a given answer across the relevant dimensions of quality (coverage, correctness, attribution, etc.).  The task is especially challenging in the scientific domain, where assessing an answer requires deep subject-matter expertise and can change over time. \citet{Asai2024OpenScholarSS} introduced ScholarQABench, which consists of multiple datasets to evaluate scientific QA systems over several dimensions. Only one of its datasets---ScholarQA-CS, which we build on in our work---evaluates answer coverage based on a set of target key {\em ingredients} (necessary points to cover in a comprehensive answer, manually annotated in that work) for each question.  The authors of ScholarQA-CS identify several limitations of their dataset, including that the annotated key ingredients could be subject to ``gaming'' because they reflect specific preferences of the two annotators, and that the full evaluation relies on heuristically set weight terms.  In our new dataset, we instead collect a diverse set of key ingredients from a variety of candidate system responses, and also develop new LLM-as-judge approaches for answer relevance and improved citation evaluation.  %

\paragraph{Evaluation}
Our \evalsqa evaluation takes in an answer to a question and outputs a score which is an average of four constituent measures of answer quality: {\em citation recall} (whether each claim in the answer is fully supported by its citations), {\em citation precision} (whether each citation in the answer supports its associated claim, at least partially), {\em answer relevance} (whether each paragraph of the answer addresses the question) and {\em answer coverage} (the fraction of necessary points covered in the answer).

All four evaluations rely on an LLM as judge, and the prompts are given in \cref{sec:sqa-eval-prompts}.  To enable accurate assessment of citation recall and citation precision, we leverage a feature of many evaluated systems: they provide quotes from each cited article intended to support the associated claim.  For each claim, if the LLM judge assesses that the claim is fully supported by any combination of its citations and they include at least one supporting quote, that claim receives a citation recall score of 1.0.  If the LLM judge assesses support based on the cited paper's title but there are no supporting quotes (this can happen because the system lacks the quote feature or because the particular sources' texts are unavailable to the system e.g. for copyright reasons), the claim receives a score of 0.5.  Otherwise, the claim receives a score of 0. Our final citation recall measure is an average over claims.  To compute citation precision, we use the LLM judge assessments of whether a citation provides at least partial support for its associated claim.  If yes, the citation receives a score of 1 (or 0.5 if it lacks a quote), otherwise it gets a score of 0. Our final citation precision is the average of these scores macro-averaged by claim.
Note that these citation metrics do not verify that citations refer to real papers or that quoted snippets actually appear in the cited sources. Accordingly, in our evaluation we discarded snippets for systems that we evaluated which do not have access to real literature and therefore would be likely to hallucinate and receive inflated scores.
For answer relevance, we instruct the LLM judge to evaluate the answer, one paragraph at a time, and instruct it to return a list of paragraphs that are not directly relevant for answering the query. Our final answer relevance score is the proportion of relevant paragraphs.

The fourth measure, answer coverage, is more challenging to assess because it requires not only evaluating the answer itself, but also identifying the key elements that a correct answer to the question must include.   Inspired by the approach taken in TREC information retrieval competitions \citep{Craswell2021OverviewOTA}, for each question we gather a pool of candidate ingredients from the systems we are evaluating,\footnote{Specifically, we source from the eight ``QA-long'' systems listed in \cref{tab:agents} plus two baseline LLMs without retrieval---Claude Sonnet 4.0 without thinking and Google's Gemini 2.5 Pro. All reports sourced were obtained before the cutoff date of June 24, 2025.} and assess the ingredients using an LLM judge. Specifically, for each evaluation question, we ask the LLM judge to extract key ingredients from each system's answer, identify specific details associated with each ingredient, and classify each ingredient's importance as "answer critical" (must-haves for answering the question) or "valuable" (nice to have, but not critical). We then cluster the extracted ingredients by instructing the LLM judge to group semantically similar ingredients together while retaining the importance label. This process results in question-specific rubrics of ingredient clusters. The ingredient extraction prompts are given in \cref{sec:sqa-eval-ingredient-extraction}.

The rubric ingredients are used at answer evaluation-time to measure coverage. For each ingredient cluster, the LLM judge gives a score of 0 (does not meet the criterion described in the rubric ingredient), 1 (somewhat meets the criterion) or 2 (perfectly meets the criterion). The final answer coverage score is a weighted average of the individual ingredient scores, with ingredient importance  determining the weight (with ``answer critical'' ingredients counting twice as much as the ``valuable'' ingredients). The answer coverage prompt is shown in \cref{sec:sqa-eval-prompts}, with a sample rubric in \cref{sec:sqa-rubric}.  

\paragraph{Data Collection}
As our test set, we gather 100 user questions issued to \agentOpenScholar \citep{Asai2024OpenScholarSS}, filtered for language, quality and topic (we select questions from the computer science domain). The details of the selection process are given in \cref{sec:sqa-query-selection}. As a development set, we retain the previously published ScholarQA-CS dataset \citep{Asai2024OpenScholarSS} of 100 questions and update its ingredient lists using the same methodology described above.

\paragraph{Choice of LLM Judge}

Since our evaluation is based upon LLM as a judge, we selected an LLM that can handle long input contexts for processing long-form answers and also follow the various constraints described in our prompts. We choose to use gemini-2.5 models. We correlated the performance of {\tt gemini-2.5-flash} and {\tt gemini-2.5-pro} as the judge on the task optimized systems \cref{sec:agents} evaluated on \evalsqa, and found that the Pearson correlation was 0.995. We therefore use {\tt gemini-2.5-flash} as the official evaluator given its lower usage cost.

\paragraph{Validation of \evalsqaPlain} 

We empirically validate our evaluation by measuring how well its ranked scores correlate with expert annotator judgments.  Specifically, the annotators are presented with a query and answers from three models and are asked to rank them based on answer quality, taking into account the quality of citations, the relevance of the text, as well as, other more subjective preferences like the flow, organization, and structure. We conduct this three-way comparison over all eval test questions, selecting at random answers from a pool of six agents---\agentScholarQA, \agentOpenAIDeepResearch, \agentElicit, \agentPerplexitySQA, \agentSTORM, and a Qwen3-8B model finetuned on QA pairs collected from production Asta Scholar QA, and calculate win rates. At the system level, we find moderate human–model agreement (Kendall $\tau$ = 0.467), which rises substantially to 0.800 when excluding Elicit outputs for which experts show systematic dispreference. At instance-level, we observe an overall agreement of 68.1\% ($\tau$ of 0.369) for instances with a clear winner (i.e., human agreement). This agreement is higher than the agreement with individual metrics (38.7\%-63.5\%), which suggests that the metrics may be working in concert to more accurately capture human judgment—complementing one another in ways that counterbalance their individual weaknesses and collectively achieving more than any single metric can on its own.

For rubric validation, we additionally investigate the concern of bias arising out of sourcing our candidate ingredients from the systems we evaluate. In particular, we examine how the answer coverage scores change for systems when they are held out of the ingredient extraction stage. Specifically, we select systems with competitive answer coverage scores (\agentScholarQA, \agentOpenAIDeepResearch, \agentElicit, \agentPerplexitySQA, \agentSciSpace) and create five different sets of rubrics for our test questions, where each set holds out one of the five systems and sources ingredients from the remaining nine systems. We recalculate the answer coverage scores using the held-out rubrics and compare them against the answer coverage scores from our full (10-system) reported results. The results show that the effect of being held out varies across systems, with three (\agentScholarQA, \agentElicit, and \agentSciSpace) experiencing significant drops in performance in the held-out condition (average 2.5 point drop; p <= 0.01) and two (\agentOpenAIDeepResearch and \agentPerplexitySQA) with insignificant drops (< 1 point drop; p > 0.16). The degree of held-in bias decreases as we add more systems: separate hold-out experiments with a 4-system rubric shows a 5 point average drop across all evaluated systems.  This suggests that including more systems for rubric creation is helpful for mitigating bias. However, more investigation is necessary to determine the reasons for bias and how to most fairly evaluate systems that we not used for rubric creation.

\paragraph{Tools Cutoff Date}
Our long-form QA task relies on access to the keyword and snippet search functionalities provided in \toollitapi.  The corpus cutoff date for both these tools is set to May 1$^{st}$ 2025 for this task.

\paragraph{Example Input}
An example input can be found in \cref{sec:sqa-example}.

\subsubsection{Query Selection}
\label{sec:sqa-query-selection}
Here we outline the procedure for collecting 100 test set queries. We obtained from OpenScholar on Feb 21, 2025 8K random input queries with three words or more, and used an LLM (Claude Sonnet 3.5) to annotate them over five dimensions: language, field of study, clarity, completeness, and query type.\footnote{For query type, we instruct the model to distinguish between queries that contain an identifiable request, queries that resemble search terms, and  queries that seek to test the capability of the agent (e.g., ``can u write ?" or ``can i speak chinese?''[sic]).} Based on the generated annotations, we down select to English, Computer Science queries that express clear research request, for a total of 3.5K queries. We then random sample 200 instances, which are then manually examined by four of our authors for question clarity, quality, and answerability to obtain our final 100 test queries. For detailed prompts, see \cref{sec:sqa-query-selection-prompts}.

\subsection[\evaltablesPlain]{\evaltablesNoSmall}
\label{sec:eval-tables}
\paragraph{Data Collection} \citet{padmakumar2025setting} identify that instances in \texttt{ArxivDIGESTables} sometimes contain one of the following issues:
\begin{itemize}
    \item \emph{Generic} columns (e.g., year of publication, research focus etc.)
    \item \emph{Unrecoverable} columns containing information that cannot be obtained from full-texts of papers in the table (e.g., dataset instances)
\end{itemize}
Generic columns are trivially easy to generate (over-optimistic performance estimates), while unrecoverable columns are impossible to generate (under-optimistic estimates). Therefore, evaluating on a subset free from these issues ensures that we obtain a realistic estimate of model performance. Since filtering such instances automatically is non-trivial, \citet{padmakumar2025setting} manually curate \evaltables, a subset of $170$ instances free of these issues. We use this subset, randomly sampling $100$ instances to create the test set and using the remaining as a validation set. Table~\ref{apx:tab:arxivdigest_fos} presents the distribution of fields of study in \evaltables.

\begin{table}[t]
\centering
\begin{tabular}{l r}
\toprule
\textbf{Field of Study} & \textbf{\# Papers} \\
\midrule
Computer Science & 94.3\% \\
Mathematics & 21.3\% \\
Engineering & 9.1\% \\
Medicine & 5.8\% \\
Physics & 4.2\% \\
Biology & 1.3\% \\
Other &  0.8\%\\
\bottomrule
\end{tabular}
\caption{Distribution of fields of study (FoS) among papers in the \evaltables validation and test sets. Note that a paper can have multiple FoS tags. Tags with fewer than five papers are grouped into the ``Other'' category, which includes Geology, Sociology, Materials Science, History, Political Science, Environmental Science and Chemistry.}
\label{apx:tab:arxivdigest_fos}
\end{table}

\paragraph{Evaluation} \citet{newman-etal-2024-arxivdigestables} originally proposed a reference-based automated evaluation procedure for the task of literature review table generation. Their procedure consists of two components: evaluating the schema (columns) and values (cells) for a generated table. However, this decomposed evaluation has two disadvantages. First, it requires agents evaluated on this task to expose the same set of components (column generation and cell value generation), instead of allowing flexibility in agent design. Second, cell value evaluation is conducted by providing agents with the set of ``gold'' columns from the reference table and assessing how well generated cell values match the cell values in the reference table. Therefore, this evaluation component effectively just measures the ability of agents to perform question answering over a single paper. To address these disadvantages, we develop an end-to-end evaluation methodology inspired by \textsc{TabEval} \citep{ramu-etal-2024-bad}. The \textsc{TabEval} protocol first represents a generated table's semantics by breaking it down into a list of natural language atomic statements, a process referred to as \emph{table unrolling}. Then, it compares these statements against ground truth statements produced from a reference table using entailment-based measures. We adopt the same approach, prompting GPT-4o to perform unrolling on generated tables, and then reporting the proportion of ground truth statements from the reference table that are entailed by the unrolled generated table (judged by GPT-4o) as recall. The prompts for table unrolling and assessing entailment are provided in \cref{sec:unroll-prompt} and \cref{sec:eval-prompt}.

\paragraph{Example Input}
An example input can be found in \cref{sec:tables-example}.

\subsection[\evalsuperPlain]{\evalsuperNoSmall}
\label{sec:eval-super}

\paragraph{Task}
Each input in \evalsuper consists of (a) a question specifying a particular research task to execute within a code repository (see example in \cref{sec:super-example}), (b) a specification of a particular output result to produce, and (c) and details of the corresponding GitHub repository. The goal then is for the agent to download the target repository, and perform all of the necessary setup and configuration needed for running the repository code, modify specific details in the code as needed for the task (e.g., dataset name or location), execute the target task, and finally report the result in the desired format.

\paragraph{Annotation}
What makes \evalsuper challenging is that such repositories are not well-documented, each repository has its own set of issues, and while it's sometimes possible to make a high-level solution plan, it is very difficult to predict what specific error will one encounter during the setup and execution process. Gold solution annotations for these tasks were therefore obtained using high skilled annotators familiar with running ML and NLP experiments, hired through Upwork.\footnote{\url{https://www.upwork.com}}. They produced solutions in the form of Jupyter notebooks,\footnote{\url{https://jupyter.org}} which are also available as part of the benchmark.

\paragraph{Evaluation}
AstaBench includes two of the original splits from \citet{bogin-etal-2024-super-emnlp}: the \emph{Expert} split containing 45 end-to-end problems as our test set and
the \emph{Auto} split containing 50 auto-generated problems (generated based on the README file of respositories that pass a certain filter)
as our development set.
Scoring for the Expert split is done by computing the exact match metric between the produced solution and the annotated gold solution (often a \texttt{JSON} dictionary containing output experiment metrics such as loss values).

\paragraph{Example Input}  An example input can be found in \cref{sec:super-example}.

\subsection[\evalcorebenchPlain]{\evalcorebenchNoSmall}
\label{sec:eval-corebench}
The version of {\evalcorebench} that we include in {\astabench} is adapted in a few ways:
\begin{itemize}
    \item The original task comes with three difficulty levels (Easy, Medium, and Hard).  We use the Hard version, which makes the task more challenging by removing several files from the capsule (such as the \texttt{run} script and the pre-computed result files), so the agent has to figure out how to install and run the code before it can do its analyses.
    \item We remove instances that would require a GPU to run, to keep the resource requirements in line with the rest of the tasks.  This reduces the dataset to 37 samples instead of the original 45.
    \item Though not mentioned in the paper, the original benchmark code includes a standard prompt\footnote{\url{https://github.com/siegelz/core-bench/blob/db8a3d00c25fc30cf091f6310203b7c715268084/benchmark/benchmark_prompts.json}} that describes the general task requirements and expected format of the output report.  We always include these instructions in the task input to ensure that the task is self-contained.
    \item We use the train split of the original dataset as the validation split in {\astabench}.
\end{itemize}

The field of study distribution in the test set is 14 Social Sciences problems (37.8\%), 12 Medical Sciences (32.4\%), and 11 CS (29.7\%).

\paragraph{Example Input}  An example input can be found in \cref{sec:corebench-example}.

\subsection[\evaldatasciPlain]{\evaldatasciNoSmall}
\label{sec:eval-datasci}

We use the original version of \evaldatasci from \cite{lai2023ds} and the task implementation from Inspect evals \citep{inspect_evals_2025}. In contrast to the original test set, we reserve 100 examples from the original set for validation and system development. 

\paragraph{Example Input}  An example input can be found in \cref{sec:datasci-example}.

\subsection[\evaldiscoverybenchPlain]{\evaldiscoverybenchNoSmall}
\label{sec:eval-discoverybench}

\citep{majumder2024data} provide initial evidence for the automated scientific discovery paradigm within the setting of \emph{data-driven discovery}, where both search and verification of hypotheses may be carried out using a dataset alone (i.e., 
after physical experiments and data collection,
but the extent of this ability remains unclear. 
We, therefore, aim to systematically evaluate the following question:     \emph{How capable are current state-of-the-art LLMs at automated data-driven discovery?}.

Answering this question is hard, as data-driven discovery in the wild (real-world) is
diverse across domains and subject areas, which in turn makes it difficult to build
a robust evaluation framework to measure progress.
We address this using a pragmatic formalization of data-driven discovery, namely
the search for a {\it relationship} that may hold between {\it variables} in a {\it context},
where (importantly) the description of those facets may not be in the language of the
dataset. A data-driven discovery task then has one of these components missing, e.g., \textit{``How
did urban land use affect the invasion of introduced plants in Catalonia?''}.
Importantly, this formalization allows for systematic,
reproducible evaluation over a wide variety of real-world problems, by leveraging these facets.

\paragraph{Task} \evaldiscoverybench~\citep{majumder-etal-2025-discoverybench} is a novel benchmark for discovering data-driven hypotheses. In this benchmark, a \emph{data-driven discovery task} is defined as follows: Given one or more task dataset(s) and a discovery goal, derive a hypothesis addressing the goal with the highest specificity for the context, variables, and relationship supported by the dataset(s). Optionally, a workflow for deriving a hypothesis can be output to augment information already present in the hypothesis. Each hypotheses have to be verified programmatically (e.g., using Python) through a data analysis workflow.  

\paragraph{Data Collection} Our goal is to replicate the scientific process undertaken by researchers to search for and validate a hypothesis from one or more datasets. We focus on six scientific domains where data-driven research is the cornerstone of scientific progress: sociology, biology, humanities, economics, engineering, and meta-science. Our gold trajectories to solve a discovery task carefully follow the published papers' workflows in respective domains. As most of the papers are highly cited, peer-reviewed, and from top venues in the domains, it is reasonable to assume the published workflows are scientifically valid.

The domain distribution in \evaldiscoverybench is shown in Table~\ref{apx:tab:discoverybench_domains}.

\begin{table}[t]
\centering
\begin{tabular}{l r}
\toprule
\textbf{Domain} & \textbf{Percentage} \\
\midrule
Meta-science & 41.8\% \\
Sociology & 24.3\% \\
Humanities & 15.9\% \\
Biology & 6.7\% \\
Engineering & 6.3\% \\
Economics & 5.0\% \\
\bottomrule
\end{tabular}
\caption{Distribution of domains in \evaldiscoverybench.}
\label{apx:tab:discoverybench_domains}
\end{table}

\paragraph{Evaluation} We evaluate task performance by measuring the alignment of the predicted and gold hypotheses in natural language.
We designed a model-based evaluation strategy using \texttt{gpt-4-preview-0125} as the \emph{evaluator}, conditioned on our structured formalism of data-driven hypotheses, i.e., a hypothesis is composed of a context, variables, and a relationship between interacting variables. Critically, the evaluator assesses entailments/equivalences between linguistic elements of a predicted and gold hypothesis pair, following several LM-based language entailment as automatic tools for scientific claim verification.

\paragraph{Example Input}  An example input can be found in \cref{sec:discoverybench-example}.

\subsection[\evalendtoendPlain]{\evalendtoendNoSmall}
\label{sec:eval-endtoend}
\paragraph{Data and Data Collection} Each example is a research task in the domain of AI/NLP, for example: 
\begin{quote}
{\it ``Test whether effective prompts discovered for large language models can directly improve smaller models' performance on classification tasks.''}
\end{quote}
followed by a detailed description of the steps to perform this test. Tasks were created using a mixture of machine generation (using \agentCodeScientist's ideator tool) and human review and editing as follows: First, we collected all *ACL conference papers from 2021 or later with at least 100 citations and available on arXiv (288 papers). The ideator tool then picks two at random and uses these to LLM-generate up to five research ideas from the combination, repeated until we have $\sim$400 ideas, which are then automatically simplified, filtered, and ranked. Finally human expert raters reviewed the top ideas, discarding infeasible/impossible ideas or making small edits to repair them (if possible). The top 50 were used for the final dataset.

\paragraph{Evaluation} During idea generation, an example-specific scoring rubric is also auto-generated, asking whether all the necessary stages of research were conducted. Each rubric item is scored using LLM-as-judge against three facets of the ASD outputs separately (report, code, artifacts), to provide an overall score. More details a given in ~\cref{sec:e2e-validation}.

While we primarily report the average research-step completion rate (\cref{tab:results-endtoend}), \cref{tab:e2e-overall-completion-scores} shows the {\it overall} task completion scores (when {\it all} required rubric items are met). These overall scores are near zero, due to compounding, reflecting the continuing challenge of full end-to-end research.

\begin{table}
  \centering
  \small
  \begin{tabular}{llcc}
    \hline
    Agent	& Model	& \evalendtoend & \evalendtoendhard \\ \hline
    \agentFaker          & \modelGPTFourPointOneUnpinned     & 0.00 & 0.00 \\
    \agentCodeScientist  & \modelClaudeSonnetThreeSevenShort & 0.05 & 0.03 \\
    \agentAutoAsta       & \modelClaudeSonnetFourShort       & 0.00 & 0.03 \\ \hline
    \end{tabular}
    \caption{Overall end-to-end task completion rates ({\it all} required steps completed successfully).
      While individual step completion accuracy is reasonable (up to $\sim$70\%, \cref{tab:results-endtoend}),
      the likelihood of completing {\it all} (typically 10-15) steps remains near zero due to compounding.}
    \label{tab:e2e-overall-completion-scores}
\end{table}

\paragraph{Environment} Given the complexity and time/dollar cost of ASD agents, ASTABench supports cache-based agents where (a) answers to all examples are precomputed offline, then (b) a runtime cache-based agent simply retrieves cached answers to each question, allowing scoring in the ASTABench environment.

\paragraph{Example Input} An example input can be found in \cref{sec:endtoend-example}.

\subsection[\evalendtoendhardPlain]{\evalendtoendhardNoSmall}
\label{sec:eval-endtoendhard}

\paragraph{Data Collection} Rather than using \agentCodeScientist's ideator, we instead use the HypER hypothesis generation system \citep{vasu2025hyperliteraturegroundedhypothesisgeneration}. HypER first identifies a research trend starting from each of the highly cited ACL papers from the above collection. For each research trend it then generates an initial idea, which is then refined further based on relevant paper excerpts to propose novel, underexplored tasks. Unlike \evalendtoend, we do not apply a task simplification step, but keep the initial proposals unchanged. Next, the proposed tasks are automatically ranked and manually reviewed by human expert raters, who discard or fix infeasible tasks. Finally the top 50 tasks were used for the final dataset.

\paragraph{Example Input} An example input can be found in \cref{sec:endtoendhard-example}.

\section{Agents}
\label{sec:appendix-agents}

We describe the evaluated agents in two parts: (1) the Asta agents that we optimized for scientific research tasks, and (2) numerous baseline agents---both general and science-specific---that we provide access to through the suite.

\subsection{Asta Agents}
We release \numastaagentclasses scientific research-optimized agent classes, including \agentAsta, an orchestrator agent that automatically detects the type of task and dispatches to an appropriate task-specific sub-agent:

\paragraph{\agentPaperFinder} is our paper-seeking agent, which is intended to assist in locating sets of papers according to content-based and metadata criteria. It is implemented as a pipeline of manual-coded components which involve LLM decisions in several key-points, as well as LLM-based relevance judgments of retrieved abstracts and snippets. At a high-level, a query is analyzed and transformed into a structured object which is then fed to an execution planner that routes the analyzed query to one of several workflows, each covering a particular paper-seeking intent. Each workflow may involve multiple steps, and returns a relevance-judged set of papers, which is then ranked while weighting content relevance together with other criteria which may appear in the query (e.g., "early works on", "influential" etc).
This agent is a frozen-in-time and simplified version of our live paper-finding agent available to use in Asta, which is restricted to single-turn interactions, does not ask for clarifications nor refuses queries, and which is using only the tools exposed in the AstaBench public APIs. It is described in more details in \cref{sec:paper-finder-agent}.

\paragraph{\agentScholarQANoTables} is a previously published scientific long-form question answering system. It is composed of three components: retrieval to identify relevant passages from two Semantic Scholar corpora; a re-ranker to select the most relevant of the retrieved passages; and a multi-step LLM pipeline to create the final comprehensive report, including in-line citations.
We experiment with several LLMs (including \modelGPTFiveUnpinned) as part of the pipeline and report the best results with \modelClaudeSonnetFour. We further report results with \modelGPTFourOMiniShort, and \modelGeminiTwoPointFiveFlash to compare the performance and cost against a smaller LLM. See \citet{Singh2025Ai2SQ} for complete details on the system.

\paragraph{\agentScholarQA} is a variant of \agentScholarQANoTables that includes literature review tables. The Scholar QA system generates answers with sections each of which is either a long form paragraph or a list of items and their descriptions. In the latter case, the corresponding section also includes a literature review table comparing the cited papers across multiple dimensions relevant to the query. The creation of tables leads to more LLM calls resulting in higher costs as well. We report our best results with this variant with \modelClaudeSonnetFour as the backbone LLM. %

\paragraph{\agentAstaTableAgent} is a previously published literature review table generation system. It follows a two-step prompting workflow. Step 1 retrieves titles and abstracts of all input papers from the Semantic Scholar database and provides this information alongside the table's caption to an LLM to generate suggestions for columns/aspects along which papers can be compared. Step 2 rephrases each column as a natural language query and prompts an LLM to generate cell values per paper conditioned on snippets relevant to the column retrieved from the paper full-text. We report results with the following backbone LLMs in this two-step workflow: \modelGPTFourPointOneShort, \modelOThreeShort, \modelGPTFiveMiniUnpinned, \modelGPTFiveUnpinned, \modelClaudeThreeFiveHaikuShort, \modelClaudeSonnetFourShort, \modelGeminiTwoPointFiveFlash \modelGeminiTwoPointFivePro, and \modelLlamaFourScoutShort. See \citet{Singh2025Ai2SQ} for complete details.

\paragraph{\agentAstaCode} is an implementation of the React-style code agent in \cite{bogin-etal-2024-super-emnlp} that was originally designed for the \evalsuper evaluation. In addition to implementing a standard ReACT think-act-observe-submit loop, it also has a built-in tool for file editing and a custom trajectory representation that facilitates fine grained trajectory evaluation. This includes evaluating whether certain landmarks (i.e., expected points in the trajectory trace) have been reached by the agent to measure partial success, as well as the ability to run code agents with partially filled-in gold trajectories. While these evaluation features are currently limited to \evalsuper, this solver allows for other code tasks to be extended to facilitate this kind of intermediate evaluation, and has an abstract structure that allows for the implementation of other agent workflows beyond ReACT. 

\paragraph{\agentDataVoyager} is a role-based multi-agent system powered by a large generative model from \citep{majumder2024data}.
\agentDataVoyager can semantically understand a dataset, programmatically explore verifiable hypotheses using the available data, run basic statistical tests (e.g., correlation and regression analyses) by invoking pre-defined functions or generating code snippets, and finally analyze the output with detailed analyses.
The core components of the system consist of specialized agents that are designed to manage different aspects of the data-driven discovery process---planning, programming and code execution, and data analysis.
Additionally, to interpret plots generated during analyses, upon generation, we run a multi-modal generative model (here, \texttt{gpt-4o}) to produce a natural language summary of such figures so that other subagents can access that information as additional context. We employ the AutoGen framework\footnote{\url{https://microsoft.github.io/autogen/}} that allows agents to communicate in arbitrary order, dependent on the context, which is maintained by an Orchestrator agent. See \cite{majumder2024data} for complete details.

\paragraph{\agentAutoAsta} performs research via a LLM-based plan-and-act (hence "Panda") cycle. Given a research task, it first generates a natural language plan, then systematically performs each plan step in turn, then writes a report on the outcome. Each plan step is performed using a ReAct/CodeAct-style loop of (a) write Python code (b) execute it (c) reflect, and either recode (if step failed/incomplete) or move to the next plan step depending on the outcome. If there are too many failures the system replans from the failed step. Since the \agentAutoAsta~source code\footnote{\url{https://github.com/allenai/panda}}
 has not yet been integrated, we grade the cached results.

\paragraph{\agentCodeScientist}  is an autonomous scientific discovery system for domains comprising computational experiments (e.g., machine learning or NLP)~\citep{codescientist}.  \agentCodeScientist~implements idea creation and experiment construction through a joint genetic search over combinations of research articles and pre-specified codeblocks, which define common actions in the investigative domain (e.g., prompting a language model). 
Since the \agentCodeScientist~source code\footnote{\url{https://github.com/allenai/codescientist}} has not yet been integrated, we grade the cached results.

\paragraph{\agentAsta} is an orchestrator agent that automatically detects the type of task and dispatches to an appropriate task-specific sub-agent.
It uses a simple but effective text similarity approach, that achieves 100\% routing accuracy on the validation set. Once the task type is identified, {\agentAsta} hands off control to a specialized solver for that task category, chosen for best expected performance based on our preliminary experiments.  The full routing table can be found in \cref{sec:agents-asta-prompt}.

\subsection{Baseline Agents}

For the set of baseline agents, we provide two general agent classes and 11 scientific research-optimized agent classes:

\paragraph{\agentReAct} is a minimum-viable baseline solver that serves to measure the capabilities of LLMs without adding a sophsticated agentic architecture or task-optimized prompt.
It is a simple ReAct loop: a chat-LLM is given a message history (initially just containing its system prompt (see \cref{sec:agents-react-prompt}) and the task instance input) and provided tools, it generates an output message with some reasoning and attached tool calls, then the results of the tool calls are appended to the message history and the LLM is called again.
This continues until the \texttt{submit(answer)} tool is called, which breaks the loop and returns the final answer.  To guard against runaway loops, we also cap the agent at a maximum of 100 steps per task; in practice agents submit well before this limit (most models average under 12 steps).

The tool calls and responses are written with the native tool-calling format of the LLM (i.e., tool-call JSON objects attached to LLM output messages and special \texttt{tool} message types for responses).\footnote{E.g. for OpenAI models: \url{https://platform.openai.com/docs/guides/function-calling}}  The agent truncates tool call outputs to at most 16,384 bytes to prevent long outputs from causing errors in the LLM.

\paragraph{\agentSmolagents} is the reference \texttt{CodeAgent} from the \texttt{smolagents} library~\citep{smolagents}.  It is a ReAct agent, and as with the {\agentReAct} agent, the input at each step is a message history; however, the actions for {\agentSmolagents} are represented as code rather than via the native tool-calling format of the LLM.
Previous work has found that code-based tool calling can outperform other formats in practice \citep{codeact_wang2024}, and it has the theoretical advantages of being able to manipulate values by reference and represent logic structures such as loops in a single step, as opposed to the LLM having to simulate these structures over a long sequence of calls.
{\agentSmolagents} is instructed to produce a Python code block to take actions (see \cref{sec:agents-smolagents-prompt} for prompt details); the code block is executed in the stateful Python environment (\cref{sec:jupyter-tool}), and all of the agent's tools are made available as callable Python functions.  In addition, the agent can call a \texttt{final\_answer} function to submit its final answer.
The agent's next input includes both the return value of the final statement in the code block as well as any printed output, up to a maximum of 20,000 characters.

\paragraph{\agentYouComSearch} is a commercial Web and News Search API, which we accessed to obtain their responses.

\paragraph{\agentElicit}is a commercial AI research platform for finding, summarizing, and extracting insights from scientific papers, such as in systematic reviews. Elicit searches the Semantic Scholar database and draws on all major large language model providers to provide AI screening, extraction, and deep research reports with in-line citations. Elicit elected to make a submission to \evalsqa on \texttt{04-03-2025}, which we processed using an offline cached solver.

\paragraph{\agentFutureHouseCrow} is a general-purpose agent built on PaperQA2 that can search the literature and provide concise answers to questions \citep{Skarlinski2024LanguageAgents}. It uses a combination of OpenAI's \modelGPTFourPointOneMiniShort and \modelOThreeMiniShort as the backbone LLMs. Although PaperQA2 is open source, it does not include retrieval. As such, we accessed FutureHouse's API to obtain Crow responses.

\paragraph{\agentFutureHouseFalcon} is a closed-source agent for deep literature reviews and hypothesis evaluation, designed for long-form question answering\footnote{https://futurehouse.gitbook.io/futurehouse-cookbook/futurehouse-client}. Falcon also uses OpenAI's \modelGPTFourPointOneMiniShort and \modelOThreeMiniShort as the backbone LLM. We accessed FutureHouse's API to obtain Falcon responses.

\paragraph{\agentOpenAIDeepResearch} is a commercial deep research system that uses Web search and OpenAI's language models to answer scientific questions.  We obtained their reports by querying the \modelOThreeDRShort model via the OpenAI API for each question.

\paragraph{\agentOpenScholar} is a previously published question answering system based on fine-tuned open models \citep{Asai2024OpenScholarSS}.  It uses a custom wrapper to the snippet and keywords search functionalities of \toollitapi for retrieval and a custom reranker.  The \agentOpenScholar paper evaluated multiple variants of its RAG pipeline, here we evaluate the publicly available demo system which uses an open 8B-parameter Llama-3.1 backbone fine-tuned on synthetic data.

\paragraph{\agentPerplexitySQA} is a commercial deep research system that runs on Perplexity's proprietary search and closed LLM (Sonar).  We accessed \modelPerplexitySonarDeepResearch via Perplexity's API to obtain their responses.

\paragraph{\agentSciSpace} is a commercial system that searches Semantic Scholar, AMiner and OpenAlex, using multiple models across subtasks. Some models are fine-tuned for task-specific needs (e.g., reranking for relevance). SciSpace elected to make a submission to \evalsqa on \texttt{06-13-2025}, which we processed using a cached solver. In their submission, the LLM was identified as \modelClaudeSonnetFour which we report in \autoref{tab:results-lit-qa}.

\paragraph{\agentSTORM} is an open-source system from Stanford that uses You.com search and 
synthesizes comprehensive, Wikipedia-like articles on given topics or questions \citep{shao2024assistingwritingwikipedialikearticles}. STORM uses OpenAI's \verb|GPT-4o| and \verb|GPT-3.5| as LLM backbones in various parts of its pipeline. 

\paragraph{\agentYouComResearch} is a commercial deep research system that runs on You.com's search and unknown LLM.  We accessed You.com's API to obtain their responses.

\paragraph{\agentFaker} is a baseline agent used to validate the scoring metrics for the \catendtoend tasks. Faker simply prompts a LM to make up the report, code, and artifacts as best it can, to simulate a successful piece of research, without actually doing the work.

\subsection[\agentPaperFinderPlain]{\agentPaperFinderNoSmall}
\label{sec:paper-finder-agent}

The Asta Paper Finder agent (PaperFinder) is a frozen-in-time subset of the PaperFinder sub-component of the Asta project ("the PaperFinder Product"). AstaBench PaperFinder follows the overall paper-finding procedure of the product, but differs from it in the indices and APIs it can use, and the set of papers available to it. It also differs in some configuration options, and does not improve over time. Finally, unlike the product, it does not support multi-turn continuations, and is restricted to a single-turn scenario where the input is a complete query and the response is a ranked set of matching documents, and the evidence for each one.

PaperFinder is a system designed to locate scientific papers in a large corpus of scientific literature, while integrating several indices, APIs, search strategies and LLM-based judgments in an intelligent and effective manner. It handles three kinds of queries: navigational queries, that aim to find a specific paper known to the user, semantic queries that locates a set of papers based on semantic description of their content, and metadata queries, that aim to find papers based on metadata criteria. The types are not fully isolated, and metadata criteria may intersect with navigational or semantic criteria. It also supports modifiers like "central", "recent" or "early", which influence the ranking of the results based on metadata information.

The PaperFinder agent works as a pipeline of manual coded steps which involve LLM decisions in several key-points.\footnote{We found the manual-coding approach to be more efficient (in terms of number of LLM calls, number of tokens, and in terms of the ability to parallelize) and more reliable than a more dynamic process that grants more autonomy to the LLM, allowing it to write code and significantly influence the computation flow and search process. We do plan to switch at least some component to more dynamic workflows in later versions.} At a high level, a query enters the \emph{query analyzer} which transforms the query into a structured object reflecting the structure and semantics of the query. The analyzed query (which includes a \emph{semantic relevance criteria}) is then sent to an \emph{execution planner} which looks at the analyzer output and routes it to one of several sub-workflows, each of them dedicated to a particular kind of search (navigational, looking for a set of papers based on semantic criteria and potential additional metadata, queries that involve complex metadata criteria, and author-based queries). The result of each of these workflows is a set of papers and relevance judgments about each of them. These are then moved to a \emph{ranker} component that orders the papers in an order which is consistent with the user's request, weighing the relevance scores together with other criteria such as publication time and number of citations for each work, in particular if this is supported by the query (i.e., explicit requests for "recent", "early", "classic", "central", "well known", "little known" etc). The ranked results are then returned.

The PaperFinder agent uses the search APIs available in AstaBench.

\subsubsection{Query Analysis}
The query analyzer is LLM based and extracts a set of predefined properties of the query. The set of extracted properties is based on manual analysis of user-issued queries, and evolves over time. It covers primarily properties that are of use to the downstream components (search sub-flows and final ranker), but also includes some information that is not currently handled but that we would like to be aware of, for allowing to inform the user that a given query criteria is not supported (for example, author affiliations).

The query analyzer is implemented as several short prompts running in parallel, each targeting a different small subset of properties (ranging from 1 to 3). We do not claim this is the optimal way of structuring such a component, but we found it to be effective and have lower latency compared to a longer prompt that extracts all of the information pieces.

The query analyzer extracts the following properties:

\paragraph{Broad vs Navigational} Does the query target a specific paper (e.g., a paper's title, "the olmo paper", "the vaswani 2017 paper") or a set of papers that matches some criteria? This is similar to the navigational-vs-information-seeking distinction in traditional search queries.

\paragraph{Semantic Criteria} Semantic criteria is a constraint or a request about the content or title of the paper (papers about X, papers that do Y). Papers in academic scientific-literature retrieval benchmarks focus almost exclusively on this criteria. However, real-world queries may include additional details such as metadata constraints or other properties, as discussed below. A major role of the query analyzer is to separate the semantic criteria from the other properties, and populate it in its own dedicated string. Note that the semantic criteria may be complex and include many sub-criteria (``papers about X, Y and Z that do not do W''). The query analyzer treats these as a single criteria and extracts them as a single field. The analysis to sub-criteria happens down the line.

\paragraph{Relevance Criteria}
A main component of the paper-finder is judging the relevance of each individual candidate result. The query analyzer also breaks the semantic query into multiple sub-criteria (based on an LLM call), coupled with an importance score and a short description of each one. These criteria will be used for assessing the relevance of the individual results.

\paragraph{Metadata Constraints} Simple metadata fields (year, year-range, authors, venues, citation counts) are extracted at fields. For complex metadata constraints (nested, negated, refer to other papers, etc), if they exist, are translated into a complex data-structure which is beyond the scope of this paper.

\paragraph{Explicitly non-supported metadata constraints} These are based on metadata requests that appear frequently enough in our logs, but for which we do not currently have metadata support in the APIs and indices. Currently these includes author affiliation information ("papers from AI2 about language modeling").

\paragraph{Recency and centrality modifiers}. Common requests that correlate with metadata information, e.g. "central paper", "classic paper", "highly cited", "recent paper", "early works" etc.\footnote{Adjectives that do not correlate with metadata information, e.g., "good paper", "high quality paper", "interesting paper", "a good summary of" are currently ignored, though some of them ("a good summary of") may make their way into the semantic criteria in some cases.}

\subsubsection{Navigational Queries}

Specific paper requests are handled using a combination of three strategies that run in parallel:
\begin{enumerate}
    \item The semantic-scholar title API.
    \item Asking an LLM and then using the semantic-scholar title API to ground the answers to specific corpus-ids.
    \item Extracting key terms from the query, searching for sentences containing these terms, looking for citations within these sentences, and returning the top-cited items as candidates.
\end{enumerate}

Each of these strategies return zero or more results, which are then merged and returned.

\subsubsection{Semantic Queries}

\begin{figure}
    \centering
    \includegraphics[width=0.5\textwidth]{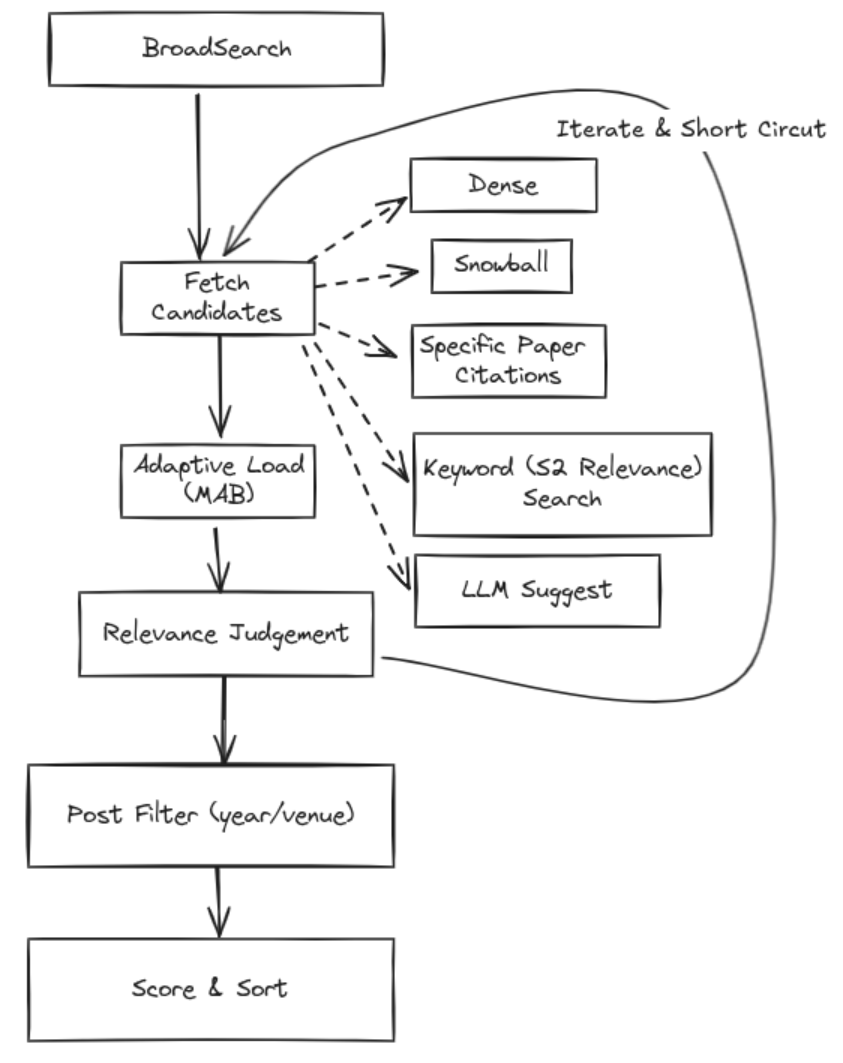}
    \caption{PaperFinder semantic query workflow}
\end{figure}

On a high-level, the process works by performing a series of retrieval steps, where each of them is followed by an LLM-based relevance filtering step. Each retrieval step broadens the scope of the previous ones, and is informed based on the relevant documents identified in the preceding steps.

\paragraph{Initial-search.}
The input to the first retrieval step is the semantic criteria from the user-query, as extracted by the query analyzer. Based on this criteria, an LLM generates $k$ rephrasing of it, and the $k+1$ queries (rephrasing and initial query) are sent to the semantic search API.

We now move from snippet-level to paper-level by aggregating the returned snippets according to the papers they come from. All snippets from the same paper are consolidated into a single item representing the paper, in which the snippets are ordered by their order of appearance in the paper's text. This aggregation is performed across queries: all the snippets in all the $k+1$ result sets participate in the aggregation, so that each \emph{paper} item potentially contains matches from multiple sources.

\noindent\emph{Cited papers.} For some queries, a non-negligible number of matching snippets refer to other papers (``Doe et al 2023 show that...''). We extract the set of papers mentioned in each snippet, and associate the snippet also to papers from this set. Thus, each snippet may participate in several paper items: both the paper it came from, and the papers it cites. Some paper items contain only evidence mentioned within them, other paper items contain only evidence from citing papers, and some contain a mix.

We now have a set of potential papers matching the query, each containing evidence snippets from multiple sources. To each of these we add also the title and abstract of the paper.

The following step is \emph{relevance judgment}, in which we filter the candidate paper set using LLM judgment (see below), resulting in a subset containing relevant papers with their relevance judgments. We keep the $m$ most promising papers for the query.
The order in which we go over the results matters for efficiency. We model this as a multi-armed bandits problem over the different sources (each query is a source).

\paragraph{Citation Tracking.}
The relevance-judgment groups the papers to categorical tiers, with \emph{highly-relevant} being the perfect matches.

This stage takes the top two categories (highly-relevant and somewhat-relevant), and performs forward and backward citation searches (a procedure known in the literature as \emph{snowballing}). In forward snowballing we look for papers that cite the papers in the set, while in backward snowballing we look for papers cited by the papers in the set. These will then also go through relevance judgment.

\paragraph{Followup queries}
We now formulate new queries based on the returned results. This is done by considering a subset of papers that were judged as relevant to the query, whose distance from the query in the embedding space was the largest. Intuitively, these are relevant results which are at the boundaries of the current search queries. An LLM reformulates a query based on the papers' titles, abstracts and returned snippets, as well as the original query. 
These are then handled like in the \emph{initial search} step: issuing queries to the vector-based API, adding cited papers, aggregating the results per paper, filtering papers that are already known from previous steps, sending to relevance judgment, and returning a result set, which is then combined with the existing result set.

\paragraph{Short-circuiting}
This process proceeds with iterations of citation tracking and followup queries for up to a predetermined number of rounds. During the process we keep track of the number of papers that were sent to relevance judgment, and the number of papers that passed it. The process stops if the number of found highly-relevant papers is sufficiently high, or if the number of relevance-judgment grows over a predetermined limit.

\paragraph{Relevance Judgment}
The relevance judgment component is applied separately to each of the found papers, and judges its relevance based on its information (title, abstract, extracted snippets, and referring snippets from other papers). The relevance judgment prompt considers each of the sub-criteria identified in query analysis, as well as the original query. Each sub-criteria is ranked as perfectly-relevant, somewhat-relevant or not-relevant. These are then combined to return a categorical relevance judgment (perfectly relevant, highly relevant, somewhat relevant, not-relevant).

\subsubsection{Metadata Queries} Simple metadata filters (venue, year) on top of semantic queries are handled as post-filters on the result set, or as ranking criteria (recent, highly cited). Queries that involve only metadata, or queries that involve a semantic criteria and a complex metadata criteria, are first sent to a dedicated metadata retrieval component, and then filtered for semantic match using the relevance judgment component. The metadata component uses LLM calls to analyze the metadata into a structured work-plan, which is then passed to a manually-coded executor which translates it to a series of API calls.

\subsubsection{Final Ranking}
Finally, we combine the relevance judgements with other criteria, based on the query analysis, using a heuristic that takes into account number of citations, publication date, and the preferences expressed in the query if they exist.

\subsection{Agent Source Code References}
\label{sec:appendix-agent-source-code}

\begin{itemize}
\item \agentPaperFinder\footnote{\agentpaperfinderurl}

\item \agentAstaTableAgent\footnote{\agentcodeurl{agent_baselines/solvers/arxivdigestables/asta_table_agent.py}{tables_solver}}

\item \agentScholarQANoTables\footnote{\agentcodeurl{agent_baselines/solvers/sqa/sqa.py}{sqa_solver}}

\item \agentAstaCode\footnote{\agentcodeurl{agent_baselines/solvers/code_agent/agent.py}{code_agent}}

\item \agentDataVoyager\footnote{\agentcodeurl{agent_baselines/solvers/datavoyager/agent.py}{datavoyager_solver}}

\item \agentAutoAsta (cached)\footnote{\agentcodeurl{agent_baselines/solvers/e2e_discovery/autoasta/autoasta_cached.py}{autoasta_cached_solver}}

\item \agentCodeScientist (cached)\footnote{\agentcodeurl{agent_baselines/solvers/e2e_discovery/codescientist/codescientist_cached.py}{codescientist_cached_solver}}

\item \agentAsta\footnote{\agentcodeurl{agent_baselines/solvers/asta/v0/asta.py}{fewshot_textsim_router}}

\item \agentReAct\footnote{\agentcodeurl{agent_baselines/solvers/react/basic_agent.py}{instantiated_basic_agent}}

\item \agentSmolagents\footnote{\agentcodeurl{agent_baselines/solvers/smolagents/agent.py}{smolagents_coder}}

\item \agentElicit (cached)\footnote{\agentcodeurl{agent_baselines/solvers/sqa/elicit/memorized_solver.py}{elicit_solver}}

\item \agentPerplexitySQA\footnote{\agentcodeurl{agent_baselines/solvers/sqa/formatted_perplexity.py}{formatted_solver}}

\item \agentSciSpace (cached)\footnote{\agentcodeurl{agent_baselines/solvers/sqa/scispace/scispace.py}{formatted_solver}}

\item \agentOpenScholar (cached)\footnote{\agentcodeurl{agent_baselines/solvers/sqa/openscholar/memorized_solver.py}{openscholar_solver}}

\item \agentOpenAIDeepResearch (cached)\footnote{\agentcodeurl{agent_baselines/solvers/sqa/general_memorized/memorized_solver.py}{formatted_solver}}

\item \agentFutureHouseCrow\footnote{\agentcodeurl{agent_baselines/solvers/futurehouse/futurehouse_solver.py}{futurehouse_solver}}

\item \agentFutureHouseFalcon\footnote{\agentcodeurl{agent_baselines/solvers/futurehouse/futurehouse_solver.py}{futurehouse_solver}}

\item \agentSTORM\footnote{\agentcodeurl{agent_baselines/solvers/sqa/storm_solver.py}{storm_solver}}

\item \agentYouComResearch\footnote{\agentcodeurl{agent_baselines/solvers/sqa/formatted_youcom.py}{formatted_solver}}

\item \agentYouComSearch\footnote{\agentcodeurl{agent_baselines/solvers/search/youcom_search.py}{youcom_solver}}

\item \agentFaker\footnote{\agentcodeurl{agent_baselines/solvers/e2e_discovery/faker/faker.py}{faker_solver}}
\end{itemize}

\subsection[\agentReActPlain prompt]{\agentReActNoSmall prompt}
\label{sec:agents-react-prompt}
The {\agentReAct} agent uses the system prompt from the InspectAI library's basic agent, constructed without knowledge of \astabench.

\begin{Verbatim}[breaklines=true,fontsize=\scriptsize]
You are a helpful assistant attempting to submit the correct answer. You have
several functions available to help with finding the answer. Each message may
may perform one function call. You will see the result of the function right
after sending the message. If you need to perform multiple actions, you can
always send more messages with subsequent function calls. Do some reasoning
before your actions, describing what function calls you are going to use and
how they fit into your plan.

When you have completed the task and have an answer, call the submit()
function to report it.
\end{Verbatim}

\subsection[\agentSmolagentsPlain prompt]{\agentSmolagentsNoSmall prompt}
\label{sec:agents-smolagents-prompt}

We use the default smolagents v1.17.0 system prompt, and additionally add tool definitions in the input user message when describing the task (note placeholders for \texttt{tool\_descriptions} and \texttt{task\_prompt}):

\begin{Verbatim}[breaklines=true,,fontsize=\small]
You have access to astabench tools in a sandbox environment. You can use these tools in your Python code:
{tool_descriptions}

Remember that you have a `final_answer(answer: str)` function that you must use to return your final answer and mark the task as completed.  The answer passed to the `final_answer` function should be a string formatted according to the task instructions; depending on the task, the string might need to contain structured outputs like JSON or code, and there may be other steps (such as writing files) that you need to perform in addition to calling `final_answer`.

{task_prompt}
\end{Verbatim}

The \texttt{task\_prompt} is simply the input from the task itself.  Each available tool is represented in \texttt{tool\_descriptions} as a function signature with the tool description and parameters.  For example, for \texttt{get\_paper} from \toollitapi, we have:
\begin{Verbatim}[breaklines=true,,fontsize=\small]
get_paper(paper_id: str, 
          fields: str = 'title,abstract,corpusId,authors,year,venue,
                         citationCount,referenceCount,influentialCitationCount')

Get details about a paper by its id.

Args:
    paper_id: The id of the paper to get. The following types of IDs are supported:
        <sha> - a Semantic Scholar ID, e.g. 649def34f8be52c8b66281af98ae884c09aef38b
        CorpusId:<id> - a Semantic Scholar numerical ID, e.g. CorpusId:215416146
        DOI:<doi> - a Digital Object Identifier, e.g. DOI:10.18653/v1/N18-3011
        ARXIV:<id> - arXiv.rg, e.g. ARXIV:2106.15928
        MAG:<id> - Microsoft Academic Graph, e.g. MAG:112218234
        ACL:<id> - Association for Computational Linguistics, e.g. ACL:W12-3903
        PMID:<id> - PubMed/Medline, e.g. PMID:19872477
        PMCID:<id> - PubMed Central, e.g. PMCID:2323736
        URL:<url> - URL from one of the sites listed below, e.g. URL:https://arxiv.org/abs/2106.15928v1

    fields: String of comma-separated fields to include in the response. E.g "url,year,authors".
    Default is "title". Available fields are: abstract, authors, citations, fieldsOfStudy, isOpenAccess,
    journal, publicationDate, references, tldr, url, venue, year.

Returns:
    The paper object.
\end{Verbatim}

\subsection{\agentAsta routing table}
\label{sec:agents-asta-prompt}

\agentAsta's routing approach starts by predicting task type based on the (character-level) lexical overlap of the input against a set of examples from the validation set.  This approach sometimes confuses highly similar tasks that have the same answer format (e.g. \evalpaperfinder and \evallitqasearchft), but as we want to route such tasks to the same sub-agent anyway, it achieves 100\% routing accuracy on the validation set.

Once the task type is identified, {\agentAsta} hands off control to a specialized solver for that task category, chosen for best expected performance based on our preliminary experiments:\footnote{Our \agentAsta experiments were started prior to the release of \modelGPTFiveShort, and due to time and the relatively poor performance of GPT-5 on many specialized solvers, we did not evaluate a \modelGPTFiveShort version for this work.  We also note that \agentAstaCode was chosen based on very early experiments with relatively old models, despite the final results showing better \evalsuper performance from \agentReAct with \modelOThreeShort.}
\begin{itemize}
\item \textbf{Paper search tasks} (\evalpaperfinder, \evallitqasearchft) → {\agentPaperFinder}
\item \textbf{Long-form QA} (\evalsqa) → {\agentScholarQA} with {\modelClaudeSonnetFourShort}
\item \textbf{Table generation} (\evaltables) → {\agentAstaTableAgent} with {\modelOThreeShort}
\item \textbf{Data analysis} (\evaldiscoverybench) → {\agentDataVoyager} with {\modelOThreeShort} configuration
\item \textbf{Code repository replication} (\evalsuper) → {\agentAstaCode} with {\modelGPTFourPointOneShort}
\item \textbf{End-to-end discovery} (\evalendtoend, \evalendtoendhard) → {\agentAutoAsta} with {\modelClaudeSonnetFourShort}
\item \textbf{Other tasks} (\evaldatasci, \evalcorebench, \evallitqaft) → {\agentReAct} with {\modelOThreeShort}
\end{itemize}
The orchestrator implements a fallback mechanism to enable sub-agents to opt out: if the predicted task-type's sub-agent doesn't produce an output, \agentAsta retries with the next most similar task type (up to 3 attempts).

\subsection{Validation of \catlit agents}
Some scientific QA agents are not capable of outputting structured data that conforms to a given schema. Accordingly, we take the plain text output of these QA agents and pass them through a "formatting" step. This formatting step uses an LLM (\modelGeminiTwoPointFiveFlashShort) to split the plain text report into sections, identifying the inline citations and returns a structured output that conforms to our \textit{SQAResponse} schema. There are also some agents that proport to have structured output capabilities but whose output quality drops dramatically when it is enabled. We also use the formatting step for these agents. The list of agents for which we use a formatting step are: You.com, Perplexity DR, OpenAI DR, and FuturHouse Crow and Falcon. 

For \agentPaperFinder, an expanded and continuously developed version of the agent—including a user interface and additional infrastructure—is actively used by a growing number of users. Throughout the extended period of development and real-world usage, we have validated the agent repeatedly using an internal eval set (which is a superset of the benchmark we now release including some additional simpler regression-testing queries). Although this internal set is not an established benchmark it has been proven useful to monitor retrieval quality and detect any regressions in recall or ranking performance. The increasing adoption among users serves as additional corroboration of both the effectiveness of the agent and the correctness of our internal evaluation methodology.

For \evallitqaft specifically, since it's a multiple-choice QA task, we evaluate the FutureHouse (creators of the original LitQA dataset) agents, and You.com and Perplexity DR because of api availability and their suitability to the task of short-form QA. The system can respond with only the correct choice or a short description with the correct choice as a json to be considered valid. For a handful of samples, we ensure the baseline systems can respond in the required format by issuing the same input prompt to their UI chat interfaces. Since \evallitqaft is a subset of the original, direct comparison with results in \citep{Skarlinski2024LanguageAgents} is difficult. Further, at the time, \texttt{PaperQA2} used \modelGPTFourTurboShort as the backbone LLM, while \agentFutureHouseCrow , which is based on PaperQA2 uses \modelGPTFourPointOneMiniShort. For sanity, we look at the difference between the average accuracy result reported for PaperQA2 (66.0) and \agentFutureHouseCrow (72.0) and conclude that evaluating on fewer questions and with better SOTA models explains it.

For \agentAstaTableAgent, we expect scores on our new end-to-end evaluation metric to generally be in the same range as the results reported by \cite{newman-etal-2024-arxivdigestables}.

For \agentPerplexitySQA, we set ``reasoning\_effort=high'' and ``search\_context\_size=high'', maximizing the model's compute and offering it the best possible performance on our datasets. The Perplexity API also provides a ``search\_mode'' parameter which can be set to ``academic'' to only retrieve academic sources. However, at the time of running the system (August 3rd--7th, 2025), this disabled web search entirely, so we did not set this parameter. Finally, while we found it may be possible to prompt \agentPerplexitySQA to extract quotes in each of its cited sources, the API does not explicitly return these snippets; thus, we evaluate the model as if it only cites the title and URL of each page.

\subsection{Validation of \catendtoend agents \label{sec:e2e-validation}}

\begin{figure}[t]
\centering
\includegraphics[width=\textwidth]{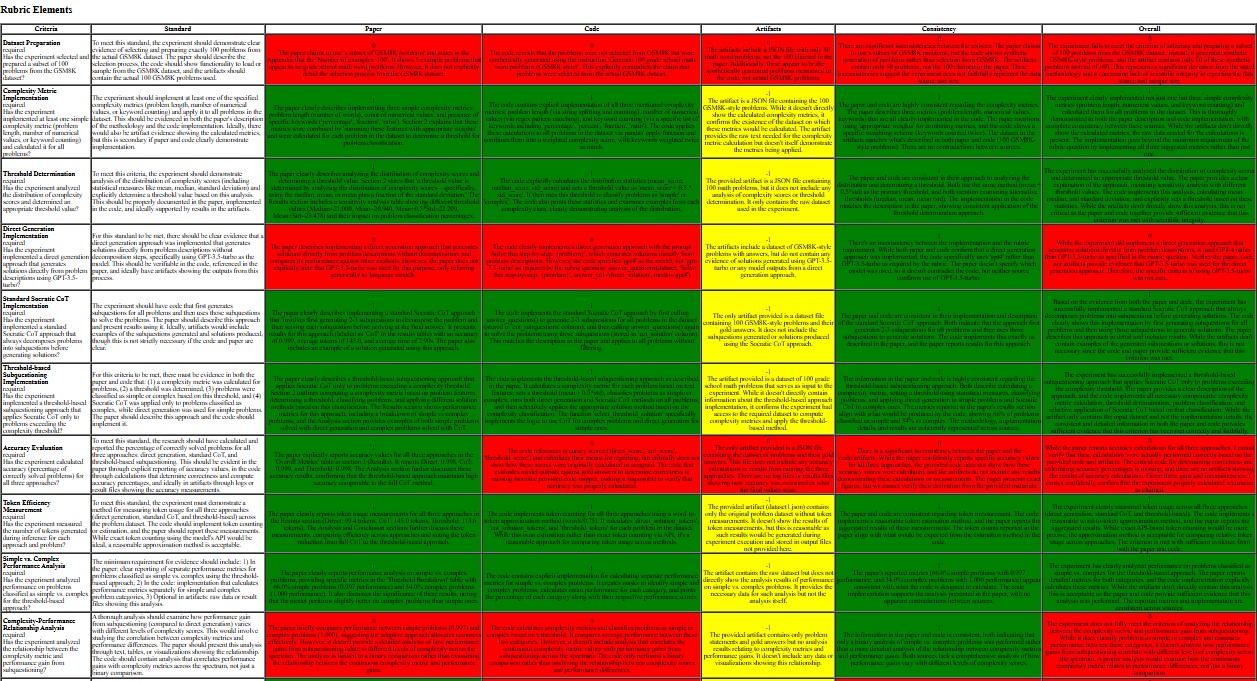}
\caption{Graphical presentation of 
scoring a single end-to-end answer. Each row is a different rubric item, columns 3, 4, and 5 show whether that rubric item was met (green), not met (red), or unknown (yellow) by the generated paper, code, and artifacts respectively. Column 6 indicates whether 3-5 are consistent (green) or not (red), with the overall verdict in the last column 7. The overall score is the average of the final column cells (green = 1, red = 0).}
\label{fig:e2e-scoring}
\end{figure}

To score and validate agents on end-to-end tasks, the E2E scorer uses a task-specific scoring rubric for each task, listing the key required facets of a valid result (e.g., downloads the right dataset, selects the right baseline, etc.). The rubrics were checked manually (and updated where needed) by human annotators. To apply these, the scorer uses LLM-as-judge to score each rubric item on each of three classes of artifact generated by the agent, namely: the generated report, the generated code, and the produced artifacts (e.g., datasets).
Scores are easily viewed in a generated HTML page (~\cref{fig:e2e-scoring}). Each facet is scored for ``meets criterion'' (green), ``fails criterion'' (red), ``no evidence either way'' (yellow). Only if all three facets are consistent and include a ``met'' is the overall criterion considered ``met''. This three-facet approach adds substantial robustness to scoring, in particular helping avoid false positives (FP), e.g., the report states an experiment was run, but the code shows otherwise, and recover from false negatives (FN), e.g., paper doesn't mention a criterion, but code shows it was indeed implemented, see \cref{tab:e2e-scoring}. The rubric scores were validated using spot-check sampling and verification by a human (judged 92\% correct on a dev set sample of 50 rubric items). Failures include occasional over-optimistic scoring (e.g., the paper only vaguely mentions a rubric item, but is still scored 1), or failures in the details (e.g., the required code has been implemented, scoring 1, but the implementation misses an important conceptual nuance of the experiment).

\begin{table}
  \small
  \centering
  \begin{tabular}{llcc}
  \hline
  & \multicolumn{3}{c}{Facets} \\
  & Paper & Code & Artifacts \\ \hline
TP: Meets criterion, and met elsewhere (overall score: 1)           & 0.44     & 0.32     & 0.48 \\
FP*: Meets criterion, but failed elsewhere (overall score: 0)        & 0.16     &  0.03     &  0.03 \\
FN*: No evidence either way, but met elsewhere (overall score: 1)    &  0.03     & 0.16     &  0.00 \\
TN: No evidence either way, and failed elsewhere (overall score: 0) &  0.02     & 0.24     &  0.01 \\
TN: Fails criterion (overall score 0)                               & 0.35     & 0.26     & 0.48 \\
\hline
\end{tabular}
\caption{Different ways that the three facets combine (fractions) for scoring 
an end-to-end rubric criterion, in particular how items that would have
been false positives (FP*) or false negatives (FN*) based on a single facet are corrected. 
For example, for 16\% of the answers, the produced paper suggested
  a rubric criterion was met, but the code and/or artifacts showed it was actually not, (desirably) resulting in an overall score of 0
  for that criterion, correcting what would have otherwise been a false positive based on the paper alone. \label{tab:e2e-scoring}}
\end{table}

\section{Additional Experimental Details and Results}
\label{sec:appendix:exp}

\subsection{Experimental Design}
\cref{tab:models} provides a list of models run in our experiments.

\begin{table}[tp]
  \renewcommand{\tabcolsep}{4pt}
  \centering
  \small

  \caption{Models run in our study. Model names are mapped to the model identifiers used during API calls, with $\ddagger$ used to disambiguate models that were called without their date identifiers for full transparency.}

  \rowcolors{2}{gray!10}{}
  \begin{tabular}{@{}L{3.5cm} L{3.4cm} L{1.5cm} l l@{}}
    \toprule
    Name & Model ID & Organization & Open-Weight & Inference Provider \\
    \midrule
    \modelGPTThreeFiveTurboShort & \modelGPTThreeFiveTurbo & OpenAI & \no & OpenAI \\
    \modelGPTFourOMiniShort & \modelGPTFourOMiniShort & OpenAI & \no & OpenAI \\
    \modelGPTFourOShort & \modelGPTFourO & OpenAI & \no & OpenAI \\
    \modelGPTFourOUnpinned & \modelGPTFourOUnpinnedID & OpenAI & \no & OpenAI \\
    \modelGPTFourPointOneShort & \modelGPTFourPointOne & OpenAI & \no & OpenAI \\
    \modelGPTFourPointOneUnpinned & \modelGPTFourPointOneUnpinnedID & OpenAI & \no & OpenAI \\
    \modelGPTFourPointOneMiniShort & \modelGPTFourPointOneMiniShort & OpenAI & \no & OpenAI \\
    \modelGPTFiveMiniShort & \modelGPTFiveMini & OpenAI & \no & OpenAI \\
    \modelGPTFiveMiniUnpinned & \modelGPTFiveMiniUnpinnedID & OpenAI & \no & OpenAI \\
    \modelGPTFiveShort & \modelGPTFive & OpenAI & \no & OpenAI \\
    \modelGPTFiveUnpinned & \modelGPTFiveUnpinnedID & OpenAI & \no & OpenAI \\
    \modelOThreeMiniShort & \modelOThreeMiniShort & OpenAI & \no & OpenAI \\
    \modelOThreeShort & \modelOThree & OpenAI & \no & OpenAI \\
    \modelOThreeUnpinned & \modelOThreeUnpinnedID & OpenAI & \no & OpenAI \\
    \modelClaudeThreeFiveHaikuShort & \modelClaudeThreeFiveHaiku & Anthropic & \no & Anthropic \\
    \modelClaudeSonnetThreeSevenShort & \modelClaudeSonnetThreeSeven & Anthropic & \no & Anthropic \\
    \modelClaudeSonnetFourShort & \modelClaudeSonnetFour & Anthropic & \no & Anthropic \\
    \modelGeminiTwoFlashShort & \modelGeminiTwoFlash & Google & \no & Google Vertex AI \\
    \modelGeminiTwoPointFiveFlashShort & \modelGeminiTwoPointFiveFlash & Google & \no & Google Vertex AI \\
    \modelGeminiTwoPointFiveFlashUnpinned & \modelGeminiTwoPointFiveFlashUnpinnedID & Google & \no & Google Vertex AI \\
    \modelGeminiTwoPointFiveProShort & \modelGeminiTwoPointFiveProShort & Google & \no & Google Vertex AI \\
    \modelPerplexitySonarDeepResearchShort & \modelPerplexitySonarDeepResearch & Perplexity & \no & Perplexity \\
    \modelLlamaFourScoutShort & \modelLlamaFourScout & Meta & \yes & Together AI \\
    \modelOpenScholar & \modelOpenScholar & Meta / Allen~AI & \yes & Self-hosted \\
    \bottomrule
  \end{tabular}
  \label{tab:models}
\end{table}

\subsection{Evaluation on full set of LitQA2 dataset}
\label{sec:litqa2-search-results}

This section presents additional details on evaluating on the LitQA2 dataset. When evaluating on our own literature search agent (PaperFinder), we provide it with the question text as is, without including the multiple choices and without attempting to translate the question into a paper-finding query-form.
We did not do any task-specific modifications or tuning of PaperFinder for this task.

As LitQA2 was designed as a full-text search benchmark, our main results are on the \evallitqasearchft subset, for which our corpus contains full-text to all papers. Here we report results also on the original LitQA2 dataset of \citet{Skarlinski2024LanguageAgents}, in which 114 out of the 199 queries have only their abstracts, and not full text, represented in our search index. The results in \cref{tab:litqa2-search} show that PaperFinder agent obtains very similar results to the agent of \cite{Skarlinski2024LanguageAgents} despite having access to only abstracts for over half the papers, and scores significantly higher on the subsets where full text is available.

\begin{table}[thp]
  \centering
  \small
  
  \caption{Retrieval scores on full set of LitQA2 dataset}
    \begin{tabular}{@{}l L{3.0cm} L{1.7cm} r R{1.0cm}@{}}
    \toprule
    Name & original-set portion & full-text percentage & recall & recall @30 \\
    \midrule
    PaperQA2 (\citet{Skarlinski2024LanguageAgents}) & full (199) & 100\% & 69.9 & 62.8 \\
    PaperFinder (ours) & full (199) & <50\% & 70.3 & 64.3 \\
    PaperFinder (ours) & \evallitqasearchft Test (75) & 100\% & \textbf{93.3} & \textbf{90.7} \\
    PaperFinder (ours) & \evallitqasearchft Val (10) & 100\% & \textbf{80} & \textbf{80} \\
    \bottomrule
  \end{tabular}
  \label{tab:litqa2-search}
\end{table}

\section{Evaluation Task Samples and Prompts}
\label{sec:appendix:evals-samples}

This section provides a higher level of detail for evaluation tasks through example problems and rubrics, plus detailed prompts.

\subsection[\evalpaperfinderPlain]{\evalpaperfinderNoSmall}
\subsubsection{Example Problem}
\label{sec:paperfinder-example}
\inputblockfile{samples/paper_finder_validation_sample.txt}

\subsection[\evallitqasearchftPlain]{\evallitqasearchftNoSmall}
\label{appendix:eval:litqasearch}
\subsubsection{Example Problem}
\label{sec:litqasearch-example}
\inputblockfile{samples/paper_finder_litqa2_validation_sample.txt}

\subsection[\evalsqaPlain]{\evalsqaNoSmall}
\subsubsection{Example Problem}
\label{sec:sqa-example}
\inputblockfile{samples/sqa_dev_sample.txt}

\subsubsection{Example Rubric}
\label{sec:sqa-rubric}
\inputblockfile{sqa_docs/sqa-sample-rubric.txt}

\subsubsection{Evaluation Prompts}
\label{sec:sqa-eval-prompts}

\paragraph{Citation Precision and Recall}
\inputblockfile{sqa_docs/prompt_eval_citation_recall.txt}

\paragraph{Answer Relevance}
\inputblockfile{sqa_docs/prompt_eval_answer_relevance.txt}

\paragraph{Answer Coverage}
\inputblockfile{sqa_docs/prompt_eval_answer_coverage.txt}

\subsubsection{Query Selection}
\label{sec:sqa-query-selection-prompts}
\paragraph{Query Annotation Prompt}
\inputblockfile{sqa_docs/prompt_query_selection.txt}

\subsubsection{Key Ingredient Extraction and Clustering Prompts}
\label{sec:sqa-eval-ingredient-extraction}

\paragraph{Ingredient Extraction}
\inputblockfile{sqa_docs/prompt_rubric_ingredient_extraction.txt}

\paragraph{Ingredient Clustering}
\inputblockfile{sqa_docs/prompt_rubric_ingredient_clustering.txt}

\subsection[\evallitqaftPlain]{\evallitqaftNoSmall}
\subsubsection{Example Problem}
\label{sec:litqa-example}
\inputblockfile{samples/litqa2_validation_sample.txt}

\subsection[\evaltablesPlain]{\evaltablesNoSmall}
\subsubsection{Example Problem}
\label{sec:tables-example}
\inputblockfile{samples/arxivdigestables_validation_sample.txt}

\subsubsection{Table Unrolling Prompt}
\label{sec:unroll-prompt}
\inputblockfile{samples/arxivdigestables_unroll_prompt.txt}

\subsubsection{Evaluation Prompt}
\label{sec:eval-prompt}
\inputblockfile{samples/arxivdigestables_eval_prompt.txt}

\subsection[\evalsuperPlain]{\evalsuperNoSmall}
\subsubsection{Example Problem}
\label{sec:super-example}
\inputblockfile{samples/super_validation_sample.txt}

\subsection[\evalcorebenchPlain]{\evalcorebenchNoSmall}
\subsubsection{Example Problem}
\label{sec:corebench-example}

The task input for the agent:
\inputblockfile{samples/core_bench_validation.txt}

The top-level contents of the corresponding capsule (with \textcolor{red}{red} items being omitted in the Hard version we use):

\dirtree{%
.1 ./.
.2 REPRODUCING.md.
.2 code/.
.3 LICENSE.
.3 README.md.
.3 config.json.
.3 lib.py.
.3 lib2.py.
.3 lib2noDTW.py.
.3 librun.py.
.3 preprocess.py.
.3 \textcolor{red}{run}.
.3 run.ipynb.
.2 data/.
.3 LICENSE.
.3 testPreprocessed.pickle.
.3 testRemoveBeginLast.
.3 testRemoveBeginLast\_10\_15.
.3 testRemoveBeginLast\_15\_20.
.3 testRemoveBeginLast\_20\_25.
.3 testRemoveBeginLast\_25\_30.
.3 testRemoveBeginLast\_5.
.3 testRemoveBeginLast\_5\_10.
.3 test\_quicktest.
.3 train.
.3 trainTrajModel.pickle.
.3 train\_quicktest.
.2 \textcolor{red}{environment/}.
.3 \textcolor{red}{Dockerfile}.
.2 metadata/.
.3 metadata.yml.
.2 \textcolor{red}{results/}.
.3 \textcolor{red}{expResult.pickle}.
.3 \textcolor{red}{expResult\_noDTW.pickle}.
.3 \textcolor{red}{output}.
.3 \textcolor{red}{output.txt}.
.3 \textcolor{red}{output\_noDTW.txt}.
.3 \textcolor{red}{run.html}.
}

And the (abridged) content of the \texttt{README.md} file:
\inputblockfile{samples/core_bench_validation_readme.txt}

\subsection[\evaldatasciPlain]{\evaldatasciNoSmall}
\subsubsection{Example Problem}
\label{sec:datasci-example}
\inputblockfile{samples/ds1000_validation_sample.txt}

\subsection[\evaldiscoverybenchPlain]{\evaldiscoverybenchNoSmall}
\subsubsection{Example Problem}
\label{sec:discoverybench-example}
\inputblockfile{samples/discoverybench_validation_sample.txt}

\subsection[\evalendtoendPlain]{\evalendtoendNoSmall}
\subsubsection{Example Problem}
\label{sec:endtoend-example}
\inputblockfile{samples/e2e_discovery_validation_sample.txt}

\subsection[\evalendtoendhardPlain]{\evalendtoendhardNoSmall}
\subsubsection{Example Problem}
\label{sec:endtoendhard-example}
\inputblockfile{samples/e2e_discovery_hard_validation_sample.txt}

\end{document}